\documentclass[aps,prx,reprint]{revtex4-2}
\usepackage{graphicx} 
\usepackage{amsmath}
\usepackage{booktabs}
\usepackage{placeins}
\usepackage{markdown}
\usepackage{makecell}
\usepackage{xurl}
\usepackage{longtable}
\usepackage{xcolor}
\definecolor{codepurple}{rgb}{0.58,0,0.82}
\definecolor{red}{HTML}{E63946}
\definecolor{blue}{HTML}{457B9D}
\definecolor{grey}{HTML}{F0F0F0}
\definecolor{yel}{HTML}{fca311}
\definecolor{purp}{HTML}{4E148C}

\usepackage{listings}
\lstdefinestyle{mystyle}{
    backgroundcolor=\color{grey}, 
    commentstyle=\color{purp},
    keywordstyle=\color{yel},
    stringstyle=\color{purp},
    basicstyle=\ttfamily\footnotesize,
    breakatwhitespace=false,         
    breaklines=true,                 
    captionpos=b,                    
    keepspaces=true,                                   
    showspaces=false,                
    showstringspaces=false,
    showtabs=false,                  
    tabsize=2,
    deletekeywords={abs},
    deletekeywords=[2]{abs},
}
\lstdefinestyle{supstyle}{
    backgroundcolor=\color{grey}, 
    commentstyle=\color{black},
    keywordstyle=\color{black},
    stringstyle=\color{black},
    basicstyle=\ttfamily\footnotesize,
    breakatwhitespace=false,         
    breaklines=true,                 
    captionpos=b,                    
    keepspaces=true,                               
    showspaces=false,                
    showstringspaces=false,
    showtabs=false,                  
    tabsize=2
}
\usepackage[most]{tcolorbox}
\tcbuselibrary{breakable,skins}

\usepackage{hyperref}
\hypersetup{
    colorlinks,
    linkcolor={black!50!black},
    citecolor={black!50!black},
    urlcolor={black!80!black},
}
\usepackage{orcidlink}
\usepackage{cleveref}
\crefname{section}{supplement}{supplements}

\renewcommand{\doi}[1]{DOI: #1} 

\begin{document}

\title[Agentic Exploration of Physics Models]{Agentic Exploration of Physics Models}
\author{Maximilian Nägele\,\orcidlink{0000-0001-6382-2077}}
\email{maximilian.naegele@mpl.mpg.de}
\affiliation{Max Planck Institute for the Science of Light, Staudtstra{\ss}e 2, 91058 Erlangen, Germany}
\affiliation{Department of Physics, Friedrich-Alexander Universit\"{a}t Erlangen-N\"{u}rnberg, Staudtstra{\ss}e 5, 91058 Erlangen, Germany}
\author{Florian Marquardt\,\orcidlink{0000-0003-4566-1753}}
\affiliation{Max Planck Institute for the Science of Light, Staudtstra{\ss}e 2, 91058 Erlangen, Germany}
\affiliation{Department of Physics, Friedrich-Alexander Universit\"{a}t Erlangen-N\"{u}rnberg, Staudtstra{\ss}e 5, 91058 Erlangen, Germany}

\begin{abstract}
The process of scientific discovery relies on an interplay of observations, analysis, and hypothesis generation. Machine learning is increasingly being adopted to address individual aspects of this process. However, it remains an open challenge to fully automate the heuristic, iterative loop required to discover the laws of an unknown system by exploring it through experiments and analysis, without tailoring the approach to the specifics of a given task. Here, we introduce SciExplorer, an agent that leverages large language model tool-use capabilities to enable exploration of systems without any domain-specific blueprints, and apply it to physical systems that are initially unknown to the agent. We test SciExplorer on a broad set of models spanning mechanical dynamical systems, wave evolution, and quantum many-body physics. Despite using a minimal set of tools, primarily based on code execution, we observe impressive performance on tasks such as recovering equations of motion from observed dynamics and inferring Hamiltonians from expectation values. The demonstrated effectiveness of this setup opens the door toward similar scientific exploration in other domains, without the need for fine-tuning or task-specific instructions.
\end{abstract}

\maketitle
\begin{figure*}[t]
    \centering
    \includegraphics[width=0.99\linewidth]{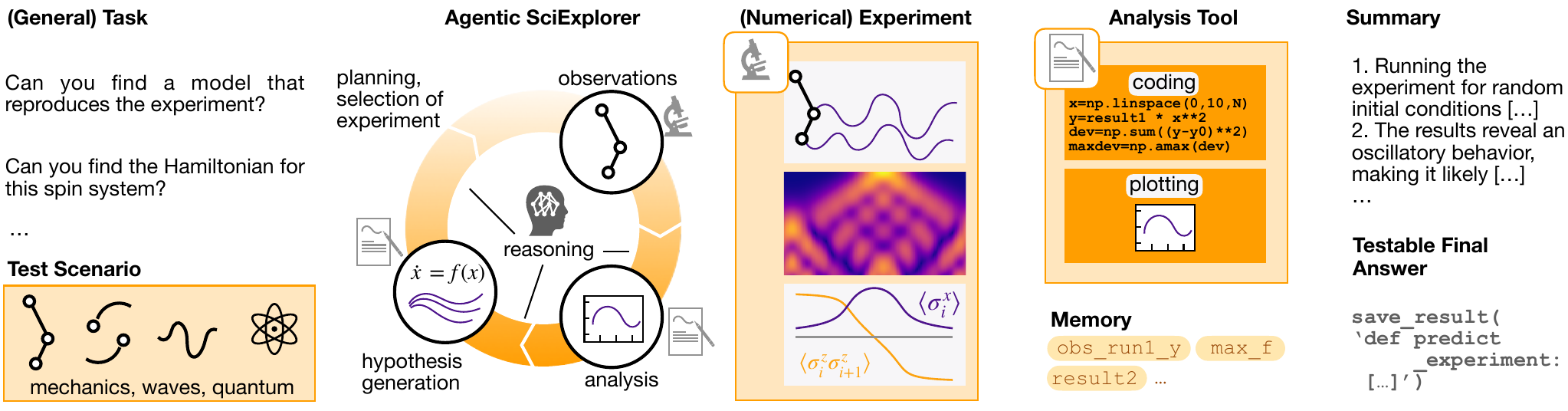}
    \caption{\textbf{Agentic SciExplorer.} The explorer is given a general task, which is then applied to a specific physics scenario. During the scientific exploration cycle, LLM-based reasoning is employed to select a (numerical) experiment to be carried out, calling the appropriate tool. The resulting observations are analyzed using one or several calls to analysis tools, often involving both the coding and the multimodal capabilities of the LLM. This cycle repeats until the LLM agent decides it has acquired sufficient information. Finally, a summary of the reasoning chain is produced for the human user. For benchmarking, the agent is asked to provide a testable final answer, e.g.\ in the form of code that can be run and evaluated.}
    \label{fig:fig_explorer}
\end{figure*}

\section{Introduction}
In recent years, specialized machine learning techniques have been increasingly employed to address various individual aspects of the scientific discovery process, for example, by suggesting new experiments \cite{Alexey2018,2023NathanAutonomousLab, 2024LiBayesianChem}, predicting or analyzing experimental results\,\cite{2021JumperAlphaFold, 2023MerchantGnome, bohrdt2019}, learning meaningful representations of data\,\cite{2025dallaNucleotide, 2025RichterDelineating}, or recovering governing equations as symbolic expressions\,\cite{cranmer2020}. However, these tools are typically designed to solve a specific task in a specific setting.
Large-language models (LLMs) exceed the capabilities of the usual machine learning approaches in several ways: Famously, they can work on a large variety of tasks without any specialized training dataset, often in a zero-shot manner, i.e.\ without even providing a few exemplary solution templates\,\cite{kojima2022large}. They work well with arbitrarily structured input, including multimodal data mixing text and images\,\cite{Alayrac2022} and can reason, producing a textual record of their arguments. As a result, LLMs can work well in situations where heuristic approaches and some level of creativity are required, all the time relying on their general factual knowledge garnered during training. For the tasks we will consider here, this includes, in particular, math and physics knowledge, supplemented by outstanding abilities in coding\,\cite{romera2024mathematical, chen2021evaluatinglargelanguagemodels}. At the same time, the greatest weakness of LLMs is their propensity to hallucinate. One approach to overcome this to some extent, also relied on in the present work, is to focus on scenarios where some level of self-correction is possible, with independent ways to check and repudiate results\,\cite{huang2025survey}.

In the past few years, the power of LLMs has been boosted by agentic behavior (tool use). By allowing the LLM to call up external software tools, one can combine the reliability of those tools with the heuristics and general knowledge of the LLM\,\cite{yao2023react}. First examples of agentic behavior have recently been introduced into several scientific domains. 
In computer science, LLMs conduct complex coding tasks\,\cite{yang2022}, and can, to some extent, also suggest new research ideas and summarize their findings\,\citep{lu2024ai}.
In chemistry, they combine Web search, specialized simulation tools, and access to laboratory interfaces 
to, for example, optimize chemical reactions\,\citep{boiko2023au}, and 
synthesize molecules\,\citep{bran2024au, ruan2024an}. In biology, agentic LLMs are employed to find gene perturbations resulting in specific phenotypes \citep{roohani2025biodiscoveryagent}, and design gene-editing experiments\,\citep{huang2024crispr}. 

In physics, nonagentic LLMs have been used to solve well-defined tasks such as calculating Hartree-Fock Hamiltonians\,\citep{pan2025quant} or answering questions about astrophysics\,\citep{nguyen2023astrollama}. Additionally, their performance has been tested on a suite of benchmarks containing typical exam questions ranging from high school to Olympiad-level difficulties\,\citep{wang2023scibench, qiu2025phybench, zhang2025physreason}, as well as on typical scientific coding tasks\,\citep{tian2024sci}.

While exploring the capabilities of agentic LLMs in physics has been called for\,\cite{lu2025can, barman2025large}, previous work focuses on building specialist agents with specific tools for tasks that require adaptability but typically follow a structured blueprint. Examples include density functional calculations\,\cite{wang2025dreams}, cosmology calculations\,\cite{xu2025opensourceplanning, casas2025clapp}, calculating properties of topological materials\,\cite{zhang2025topomas}, inverse design for metamaterials\,\cite{lu2025agentic}, designing integrated photonic circuits\,\citep{sharma2025aiagentsphotonicintegrated}, cooling of trapped atom experiments\,\cite{sha2025copilotquantum}, and calibrating superconducting quantum processors\,\cite{cao2025automatic}.

In this work, we focus on exploring the capabilities of a general AI physicist in a verifiable setting requiring heuristic, open exploration. Crucially, and in contrast to previous work, the artificial physicist is given only minimal task-specific instructions. Therefore, if it performs well on the benchmark problems, we can expect good performance also on other tasks.
Specifically, we will explore how tool use can enable an LLM to automate the heuristic loop of scientific discovery, including selecting experiments, analyzing results, and forming hypotheses when exploring an unknown system.

The tools we consider include running experiments (here, numerical experiments), performing data analysis through a general coding tool, and visualizing results.
Based on these tools, the LLM can use its reasoning to drive an active learning process, deciding which analysis or experiment to carry out next based on previous results. We will explore in prototypical physics scenarios, drawn from fields like dynamical systems, wave dynamics, and quantum many-body systems, how this approach can successfully solve research-oriented tasks requiring heuristics by mimicking essential aspects of the scientific process. Tasks we will consider include discovering the equations of motion of a system by observing its dynamics and discovering Hamiltonians of quantum many-body systems from measurements.

\begin{figure}[t]
{\centering
    \includegraphics[width=0.99\linewidth]{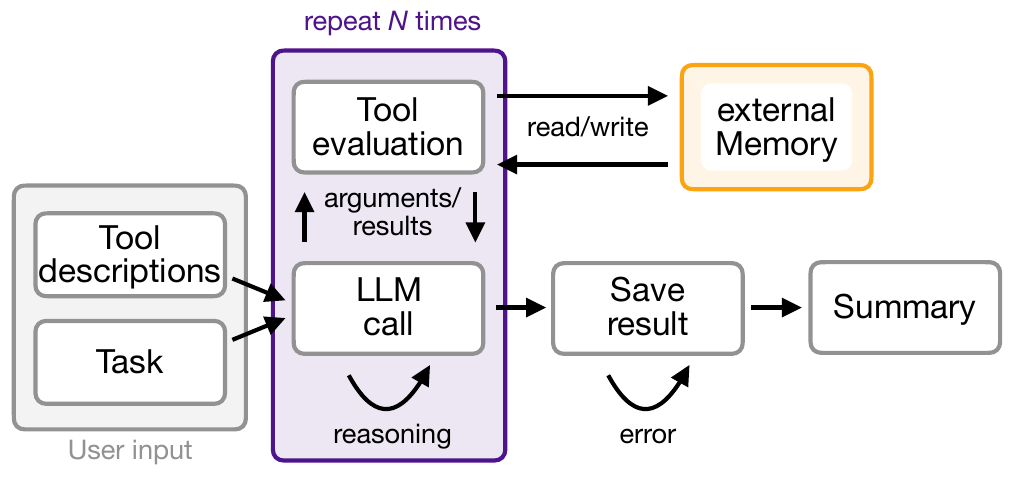}
    \caption{{\bf SciExplorer scheme.} Given task and tool descriptions supplied by the user, SciExplorer executes up to $N$ (in our case 100) steps fully autonomously. In each step, SciExplorer can reason verbally and specify arguments for tools it wants to evaluate. The tools are then executed, reading and writing to an external memory containing all previous tool results. The results are typically returned to SciExplorer in plain text. However, for large numerical arrays, SciExplorer receives a description of their type and shape. When SciExplorer calls the \texttt{save\_result} tool, the exploration stops. In case of a syntax error, SciExplorer can try again, otherwise it is asked to summarize the exploration.}\label{fig:sciexp_scheme}
}
\end{figure}

\begin{figure*}
    \centering
    \includegraphics[width=1.\linewidth]{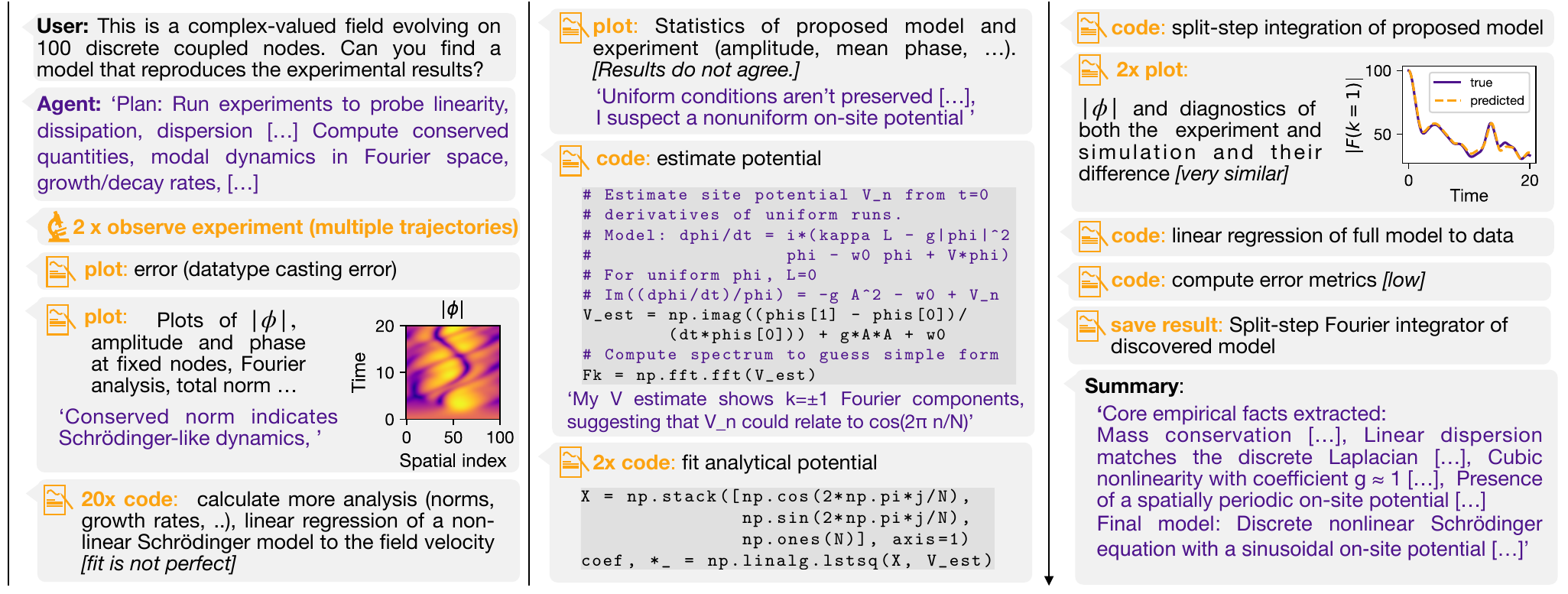}
    \caption{\textbf{Example exploration.} The agent is tasked with discovering the unknown model of a complex field system by observing and analyzing its dynamics. In this figure, we sketch in a compressed fashion some of the steps in the resulting extended autonomous exploration. The agent first runs several experiments with varying initial conditions. Then, it  infers the qualitative model (here, a nonlinear Schr\"odinger equation with external potential) from visualizations and analysis based on the experimental results. It subsequently estimates the external potential and discovers its analytical cosine structure. To validate the proposed model, the agent simulates time evolution using a (self-coded) split-step integrator and compares the results to experimental data. Finally, it saves its result as a Python function that reproduces the experimental results and contains both a simulator of the system and the underlying partial differential equation.
    The sequence of steps depends entirely on the choices of the agent, which are adapted to the given problem, reacting to its own observations and conclusions.}\label{fig:fig_example}
\end{figure*}

\section{Agentic SciExplorer}

The agentic SciExplorer is based on an LLM agent that autonomously selects and analyzes experiments (see \Cref{fig:fig_explorer}), where `experimental' results here are obtained via numerical simulations.  In a typical scenario, the agent specifies initial conditions, observes the time evolution of a system, and adaptively selects subsequent experiments based on its previous analysis. To support this process, the agent can access  
two generic analysis tools. Rather than providing narrowly defined, user-specified tools, we leverage the advanced coding capabilities of modern LLMs\,\cite{tian2024sci}, giving the agent access to a coding tool that executes arbitrary Python code. This choice maximizes flexibility and enables dynamic tool creation by the agent during runtime. Additionally, the agent can access a code-based plotting tool, which, paired with the agent's multimodal capabilities, enables the agent to quickly grasp qualitative insights about the system under study by visual inspection. 
We integrate an external memory that is accessible through the analysis tools, to offer reliable persistence of experimental and analysis results in the form of arrays of numerical values. 
When addressing a user-defined task, the agent proceeds in multiple steps. Each step involves verbal reasoning and may include multiple tool calls for controlling the experiment and analyzing results. The tool results are then communicated to the agent and added to the memory (see \Cref{fig:sciexp_scheme} and \Cref{app:SciExplorer} for a more complete description)

To explore the intrinsic capabilities of state-of-the-art LLMs in successful agentic exploration, we deliberately keep any domain-specific input to a minimum. The relatively short system prompt provides only general advice for scientific exploration (like systematically formulating hypotheses) but does not contain task-specific or even domain-specific information and is applied across all domains explored in the following (see \Cref{app:prompts}). 
When providing information to the agent through the system description or the documentation of the tools executing experimental runs, we follow a simple philosophy: We provide only information that an experimentalist investigating such a system would have access to. This includes, for example, the dimensionality, topology (e.g.\ 1d vs.\ 2d), and constituents (e.g.\ particles vs.\ spins) of the system but not the general shape of its governing laws. Therefore, we do not even tell the agent that mechanical systems are governed by ordinary differential equations or wave systems by partial differential equations.

We evaluate SciExplorer on two qualitatively distinct classes of tasks. In the first scenario, the agent is asked to reproduce the experimental results by implementing a Python function. This setting involves not only model discovery but is closer to the even more challenging task of program discovery, as the solution space consists of the vast set of all valid functions. In this case, correct solutions typically include both a simulator of the system and a description of the system’s governing laws.
In the second scenario, the agent is asked to identify the model of the system within a broad but predefined model class. The advantage of this approach is that the resulting models tend to be more interpretable. However, it requires some limited prior knowledge about the class of admissible models. To evaluate SciExplorer on both task types, we consider the first scenario for mechanical and field systems, and the second scenario for quantum systems, where the agent is asked to determine the Hamiltonian governing the experiment.

If not otherwise specified, we use GPT~5\,\cite{gpt5} as the state-of-the-art LLM backbone of the SciExplorer.
\section{Results}

SciExplorer is able to recover accurate models even when starting from minimal information about the system under consideration. Below, we explicitly present all information provided for a representative example: the damped double pendulum (see Supplemental Material\,\cite{supplement} for details on all systems).
The system and task description reads:
\textit{`You are investigating a dynamical physical system.
Can you find a model that reproduces the \texttt{observe\_experiment} function? After your exploration, save it using the \texttt{save\_result} function'.} The \texttt{observe\_experiment} tool provides only the additional information \textit{`The system has 2 generalized coordinates'}. It also includes a description of the format of its arguments, namely the initial generalized coordinates, initial generalized velocities, the start and end times of the experiment, and the time resolution.

\subsection{Mechanical Systems}
Mechanical systems provide an ideal test bed for the SciExplorer due to their wide diversity spanning conservative, dissipative, and driven regimes, and the wide range of accessible complexity levels, selected via choosing the equations of motion or their dimensionality. An example of an illustrative exploration process is summarized in \Cref{fig:fig_example}. 

\begin{figure}
    \centering
    \includegraphics[width=0.99\linewidth]{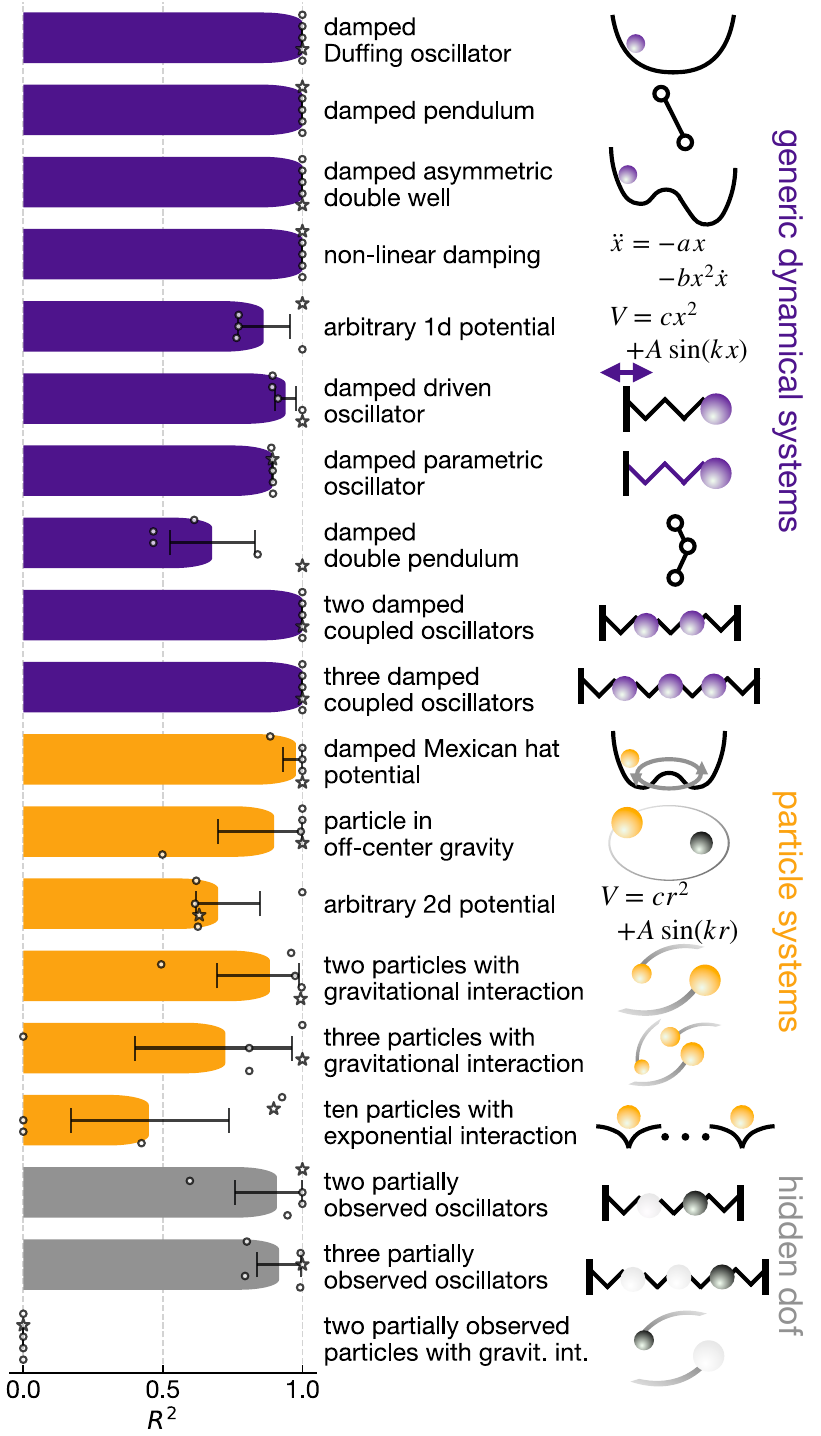}
    \caption{{\bf Mechanical systems.} The agent discovers the equations of motion of black-box physical systems by specifying initial conditions, observing dynamics, and analyzing the results. We consider generic dynamical systems with one to three generalized coordinates, particles moving in 2d, and systems where an observable particle interacts with an unknown number of hidden degrees of freedom (dof). We run 5 independent attempts per system and show the mean coefficient of determination $R^2$ of the agent's proposed model with the true system with 95\% bootstrap confidence intervals. The agent can recover the true model in a large subset of systems. Stars indicate the conversation the agent considers best when asked to rank all attempts.}
    \label{fig:eom_stats}
\end{figure}

For test purposes, we considered generic dynamical systems with up to three generalized coordinates and systems with particles moving in 2d (with up to ten particles) subject to fixed potentials or pairwise interactions (for the full equations of motion see Supplement, Table S1\,\cite{supplement}). \Cref{fig:eom_stats} summarizes the agent's performance across these systems when asked to infer their equations of motion. The agent receives information only about the number of generalized coordinates (or number of particles moving in 2d) and is asked to produce a Python function that reproduces the experimental results.
To quantify the agent’s performance, we compute the coefficient of determination $R^2$ between the 
velocities and accelerations predicted by the agent and the ground-truth dynamics.

As an expanded challenge, we also consider dynamical systems where the agent is allowed to observe only a subset of particles, and is not even told the number of unobserved particles. In those cases, we additionally require the agent to infer the initial conditions of the hidden particles. We then report the mean $R^2$ value across trajectories with random initial conditions of the observable particle (see \Cref{app:Mechanics} for details).

For a large subset of systems, the agent recovers the governing model both qualitatively and quantitatively, often achieving perfect fits ($R^2\approx1$) using only observations of the system's time evolution under initial conditions selected by the agent. Performance degrades for more complex or atypical systems, and the agent only sometimes recovers the correct model. Systems that are not close to any known models (e.g.\ the `arbitrary 2d potential') are particularly challenging, indicating that the LLM agent relies on its broad knowledge base to be successful for other systems. To identify systems correctly in most cases, the agent autonomously implements a broad repertoire of analysis techniques by generating and executing Python code, without requiring specific tools. Typically, the agent first extracts qualitative signatures from visualizations. These include, among others, linearity, mirror symmetry, limit cycles, attractors, and decay. It then constructs candidate basis terms for the governing differential equations and fits their coefficients to the observed accelerations. When applicable, it uses sparse linear regression\,\cite{brunton16sindy}. Otherwise, it fits numerical solutions of the proposed differential equations directly to the experimental data. Depending on the task, the agent uses a large set of additional strategies, such as locating centers of attraction in 2d by maximizing the radial acceleration with respect to a candidate center, testing for angular momentum and energy conservation, and using Fourier transforms to find initial parameter guesses for oscillator systems or to infer the number of hidden particles in partially observed systems (for example explorations, see Supplement S3A and S3B, for the models discovered by the agent, see Supplement Table~S2\,\cite{supplement}).

Furthermore, we test the agent's ability to self-critique by asking it, for each mechanical system, to identify in which of multiple conversations it performed best (stars in \Cref{fig:eom_stats}). When the correct model was found at least once, the agent reliably selected that conversation. However, when none of the proposed models was exactly correct, the agent often did not choose the conversation corresponding to the highest $R^2$ value. Instead, it tends to prioritize qualitative agreement, such as the existence of bound and unbound trajectories, over purely quantitative fit quality.

It is important to note that even in a real experimental setting, one can verify the quality of the solution discovered by the agent without knowing the ground truth by comparing the agent's theory with new experimental data. Therefore, one can boost the effective success rate of the agent by running several explorations and picking the best-performing model (in all but three of the scenarios of  \Cref{fig:eom_stats}, this leads to the correct solution, as it is among the five runs).

To understand which components of the SciExplorer are essential, we perform ablation studies. Depending on the physical system, both the plotting and coding tools are crucial for consistent performance. To test the importance of tool access, we compare a given exploration run of our agent against a conversation where an LLM without access to tools is initially provided with visualizations for the same number of experiments, but now with randomly chosen initial conditions. In this case, the LLM cannot recover the correct model. Interestingly, we also find that GPT~5\,\cite{gpt5} performs much better than Gemini 2.5 pro\,\cite{gemini} (for details see \Cref{app:ablation}).

The superior performance of LLM agents based on state-of-the-art reasoning models relies on significant run-time investment. For the systems studied here, a single exploration ranges from a few minutes to over 1.5 hours, with the bottleneck being the LLM's response time, not the numerical simulations of the system. We also observe that the agent generally requires more time to solve more complex systems than simpler ones (see \Cref{fig:runtim_mechanics}). Nevertheless, based on our experience as working theoretical physicists, we estimate that the overall run-time for a single task is already lower than what even an expert human would typically need for such an initially completely open exploration of an unknown system.

\begin{figure}
    \centering
    \includegraphics[width=0.99\linewidth]{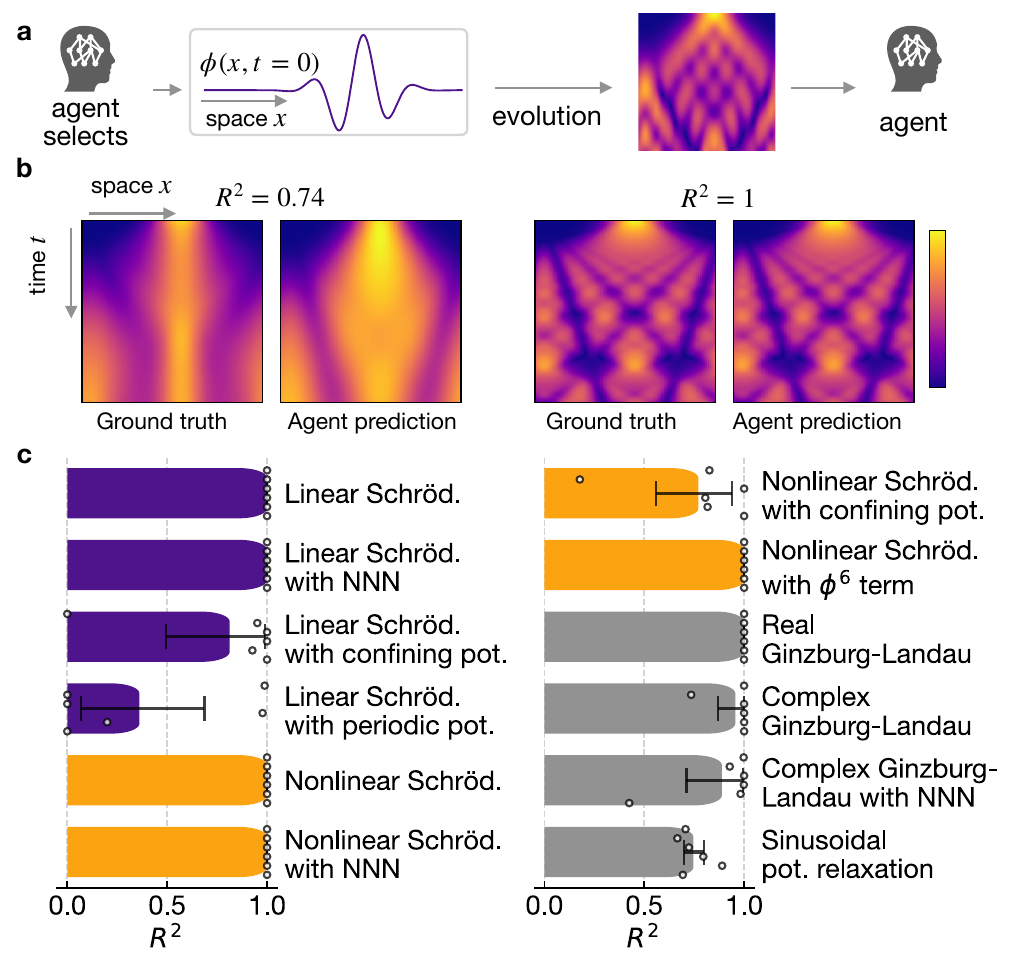}
    \caption{{\bf Waves and fields.} {\bf a} In its exploration of waves and fields, the agent can select the initial field configuration for each experimental run. {\bf b} Evolution of the absolute square $|\phi|^2$ of a Gaussian wave packet for the true model and the agent's discovered model (announced by the agent at the end of the exploration). The true model on the left is a linear Schr\"odinger equation with confining potential (we do not show the most accurate model discovered by the agent in its multiple runs). On the right, the true model is a complex Ginzburg-Landau equation with next-nearest-neighbor (NNN) hopping on a tight-binding lattice. The $R^2$ value is calculated between the evolution equations for the true and the predicted model, for multiple reasonable initial conditions (see \Cref{app:Fields} for details). {\bf c} Statistics for various scenarios. We run six independent explorations, and the agent can recover the true model ($R^2\approx 1$) in all but one scenario in at least one attempt.}
    \label{fig:fields_waves}
\end{figure}

\begin{figure*}
{\centering
    \includegraphics[width=0.99\linewidth]{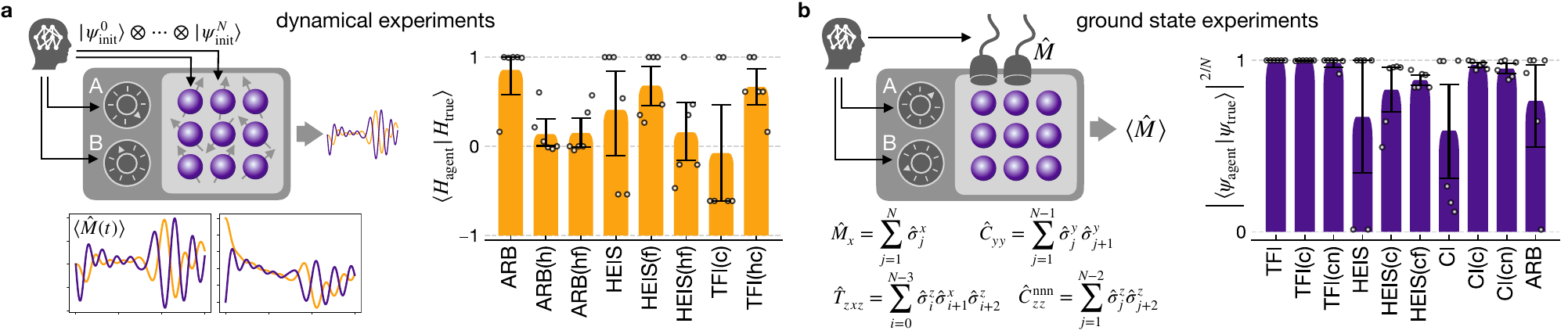}
    }
    \caption{{\bf Quantum many-body physics.} {\bf a} Discovering the Hamiltonian of a system of multiple spin-1/2 particles, based on observing the dynamics for initial conditions selected by the agent. Each experimental run produces the evolution of the single-spin expectation values $\langle {\hat M}(t) \rangle$ (with ${\hat M}={\hat \sigma}^x_j$, ${\hat \sigma}^y_j$, and ${\hat \sigma^z}_j$), for an initial (product) state selected by the agent. In some experiments, we ask the agent to discover a whole class of Hamiltonians by allowing it to control experimental parameters, whose meaning it is not aware of (`A' and `B'). The bottom left shows an example – the complex dynamics of two spins of a Heisenberg model. Right: Performance for various scenarios. Normalized scalar product between the true Hamiltonian and the Hamiltonian proposed by the agent (1 means perfect match, for details see \Cref{app:Quantum}). We consider the following systems: \emph{ARB}: Arbitrarily chosen Hamiltonian acting on 3 spins, \emph{HEIS}: 1d Heisenberg model with 10 spins, \emph{TFI}: 1d transverse field Ising model with 10 spins. The letters in brackets denote whether the agent can vary the value of a field parameter (f), of a coupling parameter (c), or observe only two spins of a larger chain (h). {\bf b}~Hamiltonian discovery by measuring the ground-state expectation values of spin operators $\hat M$ defined by the agent. Some examples are shown on the bottom left. Right: Fidelity scaled by the number of spins $N$ between the ground state of the Hamiltonian predicted by the agent after the exploration and the true ground state. We consider the following systems: \emph{TFI}: 1d transverse field Ising model with 10 spins, \emph{HEIS}: 2d Heisenberg model with 9 spins, \emph{CI}: 1d cluster Ising model with three-body interactions and 10 spins, \emph{ARB}: Arbitrarily chosen, translationally invariant Hamiltonian acting on 10 spins. Letters denote the same as in part a. Additionally, the agent can sometimes vary the number of spins in the chain, as denoted by (n).}\label{fig:quantum_result}
\end{figure*}

\subsection{Dynamics of Waves and Fields}
A fundamental cornerstone of the physical description of nature is the notion of dynamics of many degrees of freedom arranged in an extended space, with locality and translational invariance as guiding principles. These lead naturally to the evolution equations of fields and waves, which also form the basis of the most fundamental theories we have. We ask the LLM agent to explore this domain by providing access to experimental runs where the agent can set up an arbitrary initial configuration of a field $\phi$ and observe the subsequent dynamics $\phi(x,t)$ that is governed by a classical evolution equation unknown to the agent. The agent is provided only with information about the system’s topology: \textit{`This physical system consists of a complex-valued field evolving on 100 discrete coupled nodes.
The nodes are arranged in 1D with periodic boundary conditions, i.e.\ the first and last node are neighbors.'} As before, the agent is asked to implement a function that reproduces the experimental observations. Examples that we tested the agent on include the nonlinear Schrödinger equation, describing the dynamics of matter waves of interacting atoms or light waves subject to nonlinearities, and the time-dependent Ginzburg-Landau equation, describing the evolution of an order parameter field, e.g.\ in a superconductor, relaxing to its thermodynamic equilibrium configuration. In each case, there can be variations like the presence of an arbitrary external potential or of longer-range coupling terms on the underlying lattice. The SciExplorer can discover the true model for any of these variations when given multiple attempts (see \Cref{fig:fields_waves}). However, it struggles with an artificial model describing relaxation into a sinusoidal potential for the field.

In our studies, we observe that the agent often lays out a detailed exploration plan in advance, mentioning the most important field theories as plausible options, sometimes including expected signatures, such as Bragg scattering for linear waves in periodic potentials or solitonlike features for Kerr-type and other nonlinearities. It tends to start with a Gaussian wave packet as its choice of initial condition in the first experiment. If the results of this experiment indicate wave-like dynamics, the agent often adopts a principled approach where it first tries to extract the dispersion relation. One of its strategies is to launch several plane waves of varying wave number and low amplitude to avoid nonlinearities, and to analyze the evolution of the phase of the field $\phi$. Once that is settled, it tries to determine any nonlinearities by checking the amplitude dependence of the evolution. In linear systems, the agent often fits a correct $100\times 100$ Hamiltonian to the data but cannot save it, since it is stored only in the external memory not accessible by the \texttt{save\_result} tool (counted as $R^2=0$).
If the initial dynamics are not wavelike but contractive, with dissipation leading to some steady state, it will adapt its strategy accordingly. In both cases, the agent often visualizes the field $\phi(x,t)$ using space-time density plots, e.g.\ with the declared aim to `spot qualitative structures (breathers, oscillations, solitons)'.To compare its proposed model with observational data, the agent typically implements a split-step integrator or, in linear systems, diagonalizes the Hamiltonian to simulate the proposed field equation. Using these strategies, we find it to quickly make systematic progress (see Supplement Table S4 for the models discovered by the agent and Supplement S3D for an example exploration\,\cite{supplement}).

\subsection{Quantum Many-Body Physics} In domains ranging from condensed matter physics to particle physics, the fundamental language is that of quantum many-body systems. The essence of this domain, with complexities like the exponential growth of the Hilbert space with particle number and the presence of entanglement, is captured well by models of interacting spins or qubits. We therefore let the agent explore and analyze spin Hamiltonians, a task with wide-ranging applications from solid-state physics to quantum technologies\,\cite{Krenn2023, valenti2019}. In terms of experiments, we test two variations: (i) In one setting, the agent can initialize the system in a product state and then observe the temporal evolution of the expectation values of single-spin operators. (ii) In another setting, the agent can construct arbitrary observables (Hermitian operators) and obtain their expectation values in the ground state of the unknown Hamiltonian (see \Cref{fig:quantum_result}). In both settings, the agent is provided with information that the system has a 1d or 2d configuration and that it consists of spins.

We find that the agent has an active working knowledge of quantum many-body physics and spin models in particular, including standard models like transverse Ising, XXZ, and Heisenberg, but also e.g.\ Dzyaloshinskii-Moriya (spin-orbit induced) chiral couplings. It can deploy that knowledge reliably in this challenge.

For the dynamical setting, where the time traces of spin expectation values can look very complex, the agent can nevertheless quickly rule out some hypotheses. For example, it may run short-time dynamics for multiple initial states, which allows it to first check for conserved quantities such as the global magnetization, compute dominant frequency components using Fourier transforms, and discriminate local and many-body interactions. It then often obtains fits to hypothesized families of Hamiltonians (for an example, see Supplement S3E\,\cite{supplement}). We also introduced an even more challenging scenario, where the agent can only observe and control a subset of spins. As expected, this appears to be one of the most complex tasks, and the agent only sometimes makes good progress.

In the ground-state setting, the agent typically first probes the symmetry and structure of the Hamiltonian by selecting a limited set of intelligently chosen operators, obtaining single-spin expectation values as well as nearest-neighbor correlators but also more subtle quantities suited e.g.\ for exploring potential chirality, like $\left\langle {\hat S}^j_{x} {\hat S}^{j+1}_{y} - {\hat S}^j_{y} {\hat S}^{j+1}_{x} \right\rangle$. It quickly deduces aspects like open boundary conditions, translational invariance, and the likely absence of some families of spin-coupling terms in the Hamiltonian. It then selects candidate spin Hamiltonians, conjectures values of their parameters, and runs simulations to confirm or reject these hypotheses. The agent is also cautious enough to keep an open mind (as per its instructions) and lists possibilities that have not yet been refuted, like more subtle, longer-range couplings.

In an extension to both the dynamical and ground-state setting, we let the agent control one or several parameters whose meaning it is unaware of (mimicking closely a potential experimental scenario of exploration), letting it discover a whole family of Hamiltonians. We find that the agent can also successfully solve this extended task, for example, by creating visualizations of expectation values against the strength of the tunable parameters (for an example, see Supplement S3F\,\cite{supplement}).

Overall, as demonstrated in  \Cref{fig:quantum_result}, the agent also performs impressively well in discovering models in quantum many-body physics, again without any fine-tuned instructions or a prescribed exploration loop.

\begin{figure*}[t]
    \centering
    \includegraphics[width=0.98\textwidth]{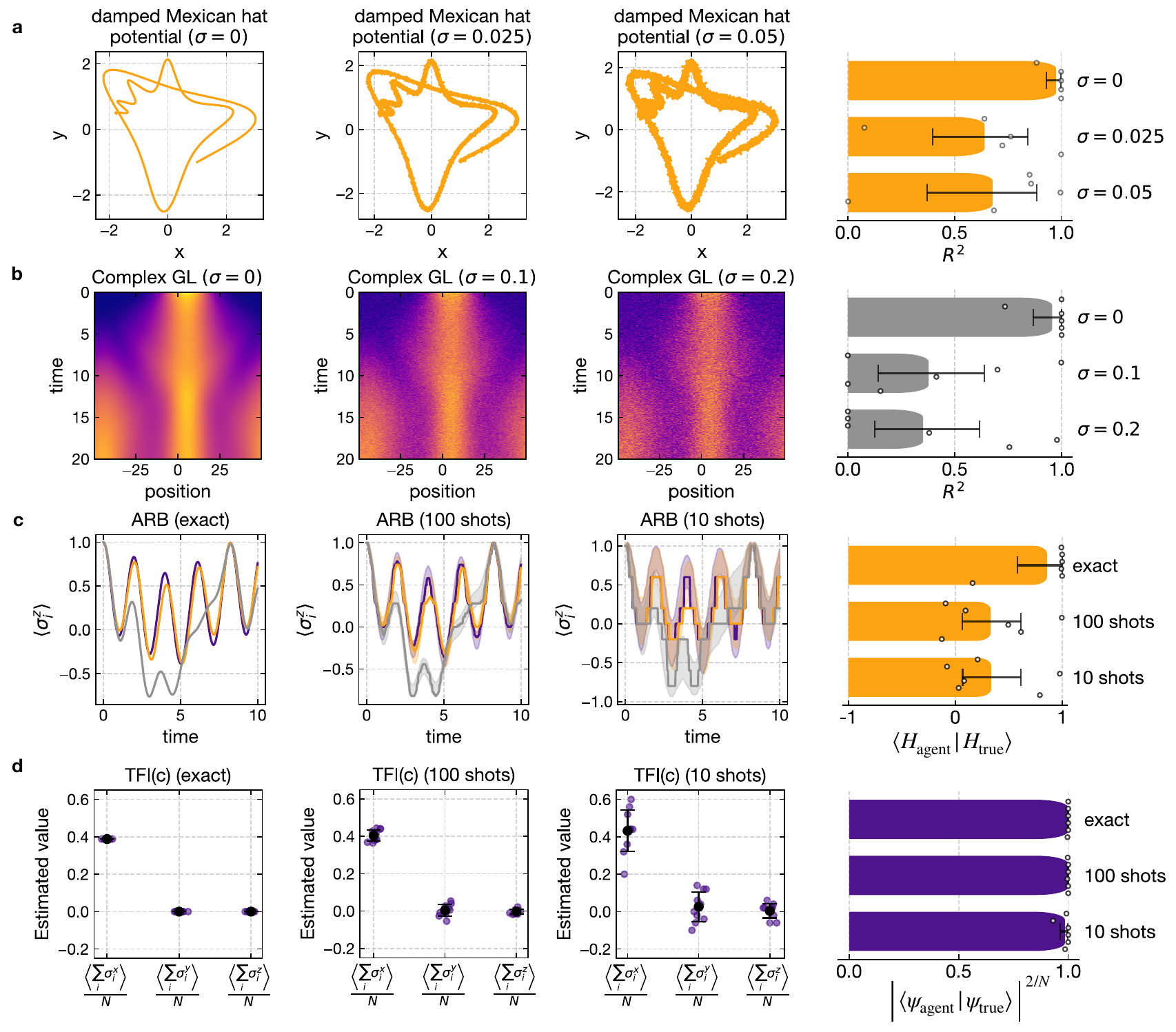}
    \caption{{\bf Experiments with noise.} 
    While the predictions of SciExplorer become less consistent for higher measurement noise levels, it successfully recovers a highly accurate model in at least one of six attempts per noise level and scenario.  
    {\bf a} Left: Example trajectories of a particle in the damped Mexican hat potential under varying Gaussian noise levels. Right: Performance of SciExplorer for different noise strengths.
    {\bf b} Left: Evolution of the absolute square $|\phi|^2$ of a Gaussian wave packet in the complex Ginzburg-Landau scenario under varying Gaussian noise levels. Right: Performance of SciExplorer for different noise strengths.
    {\bf c} Left: Example trajectories of single spin $\sigma^z$ expectation values under the arbitrary Hamiltonian (ARB) in the quantum dynamics setting. Expectation values are estimated from different numbers of single-shot measurements. Shaded areas indicate standard deviations over multiple runs. Right: Performance of SciExplorer for various numbers of measurement shots.
    {\bf d} Left: Ground-state expectation values of the total spin operators in x, y, and z direction in the transverse field Ising model with variable coupling for different numbers of measurement shots. Dots show ten examples of estimated expectation values and bars denote standard deviation. Right: Performance of SciExplorer for various numbers of measurement shots.
    }
    \label{fig:noise_experiments}
\end{figure*}

\subsection{Experiments with noise}\label{noise}

A common challenge in recovering physical models from experimental data is the presence of measurement errors. For classical systems, measurement errors are often modeled as Gaussian noise, which we add to the mechanical and field systems. In quantum mechanics, experimental observables cannot be measured exactly. Instead, they are estimated from a finite number of single-shot measurements. Therefore, we investigate the agent's performance under a finite number of measurement shots per expectation value. We run additional experiments on one prototypical system from each physical scenario presented above, using two different noise levels. We find that while SciExplorer's predictions become less consistent as noise increases, it can recover highly accurate models even under significant measurement noise for all the considered systems in at least one of six attempts per system (see \Cref{fig:noise_experiments}).

\subsection{Common failure modes}

As demonstrated above, the general knowledge and reasoning capabilities of state-of-the-art LLMs are impressive. However, they are by no means perfect and still occasionally hallucinate facts, overlook important clues, or perform superficial analyses.

\begin{figure}[t]
{\centering
    \includegraphics[width=0.85\linewidth]{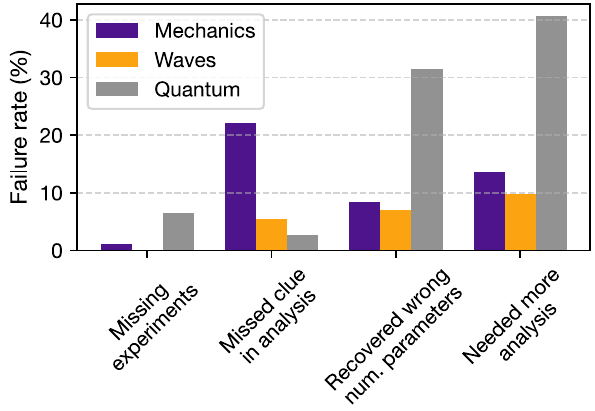}
    \caption{{\bf Common failure modes.} Failure rate per physical domain (as percentage of total runs) categorized by different reasons. Each exploration potentially contributes to multiple bars. Failure reasons were assigned through human judgment after manually analyzing all failed attempts (i.e.\ $R^2$ or overlap $<0.99$).
    \textit{Missing experiments:} The experiments run by the agent did not reveal all the qualitative information needed to extract the model. \textit{Missed clue in analysis:} There was a clue in the agent's analysis results that the agent missed (typically in a plot).  \textit{Recovered wrong num.\ parameters:} Agent found the correct model structure, but failed to find the correct numerical parameters (e.g.\ failed fits, sign errors), \textit{Needed more analysis:} The agent did not miss a clear clue but instead did not implement enough analysis routines to uncover the true model.}\label{fig:failures}
}
\end{figure}

To potentially improve SciExplorer in the future, a detailed understanding of its failure modes is desirable. Therefore, we manually analyze all failed attempts reported in this article to identify recurring mistakes made by the agent. We categorize these errors into four groups, noting that a single failed run may be assigned to multiple categories (see \Cref{fig:failures}). In particular, in the quantum systems, and occasionally in the mechanical systems, the agent does not perform sufficiently diverse experiments to uncover all qualitative signatures of the system. For example, in the quantum ground-state setting, it may fail to consider enough expectation values, leading to an incorrect model that nevertheless fits all observed measurement results.
Another common problem is that the agent discovers the correct qualitative model structure but fails to infer the correct numerical parameters. In the mechanical and wave settings, this often results from unsuccessful parameter fits. In the quantum setting, most errors arise from simple sign errors or from missing factors of two in some Hamiltonian terms. In such cases, the agent fails to even consider the possibility of an overall scaling factor or sign change.
When analyzing mechanical systems, the agent also frequently overlooks qualitative clues that are, in principle, contained in its analysis results, most often in visualizations. Examples include missing fast oscillations modulating an overall trend or failing to detect the breaking of translational invariance by an external potential in field systems (for an example, see Supplement S3C\,\cite{supplement}).
Most runs fail not because the agent missed an obvious clue (at least not obvious to the first author of this paper), but because its analysis was not sufficiently exhaustive to identify the correct model, leading the agent to commit prematurely to an incorrect one.

Many of these failure modes could potentially be mitigated through more sophisticated prompting (e.g.\ to avoid sign errors in quantum many-body systems). However, we deliberately avoided such interventions in the spirit of not introducing task- or system-specific fine-tuning. Other failure modes may only be alleviated by future LLMs with, for example, improved visual capabilities.

SciExplorer relies on an LLM as its backbone and thereby inherits the LLM's intrinsic priors about useful strategies and common physical systems. Because current LLMs are mostly trained on human-generated text, SciExplorer employs heuristics similar to a human physicist. 
While this helps to navigate the exponentially large search space of possible solutions, it also means that there exists a bias toward `likely' models. Therefore, just like for a human, recovering entirely new or uncommon models is difficult for SciExplorer.
However, physically relevant models of an experimental system will typically not contain completely unknown ingredients, but rather combinations of individual, in principle known, parts (e.g.\ different terms in a wave equation). To test SciExplorer's abilities on `uncommon' systems, we included examples such as the `arbitrary potentials', and `arbitrary Hamiltonian', which are uncommon combinations of known ingredients, and the `sinusoidal potential relaxation', which is a completely unphysical system. We find that the agent's success rate is lower on these systems than on more standard ones. Still, it can recover the true underlying model in the nonstandard mechanical and quantum systems in some attempts, while it finds only approximate solutions in the wave system containing an artificial `$\sin\left(0.1|\phi_n|^2\right)\phi_n$' term in the wave equation.
To improve SciExplorer's performance in these uncommon cases, multiple strategies could be employed in the future. In many real applications, some information about the underlying model may be available, which can easily be incorporated into SciExplorer as plain text. Even just the information that one expects uncommon terms in the solution could help to avoid focusing on known models.
Additionally, one could encourage SciExplorer to leverage data-driven model discovery techniques such as SINDy\,\cite{brunton16sindy} that rely less on heuristics and are therefore  also  less influenced by the LLM's priors. Here, SciExplorer could potentially still improve the technique substantially by tailoring it to the specific system, e.g.\ by enforcing previously discovered symmetries.

\subsection{Comparison to symbolic regression}
In this article, we describe a general AI scientist and benchmark its capabilities in a verifiable setting, namely discovering the model of an unknown physical system. The systems considered here are all governed by analytic expressions. Symbolic regression is a research field concerned with automatically extracting such expressions from data\,\cite{symregreview}.
A human attempting to solve the same task used to benchmark SciExplorer (model discovery) would typically use symbolic regression techniques as part of its discovery process. 
These techniques can be categorized into three classes: First, standard regression assumes a known analytical expression in which only numerical constants must be adjusted. Second, sparse regression techniques fit a linear combination of ansatz functions to the data (e.g.\ SINDy\,\cite{brunton16sindy, pysindy} for ordinary differential equations and PDEFIND\,\cite{PDEFIND} for partial differential equations). By including a loss term that encourages sparsity in the resulting prefactors, irrelevant ansatz functions can often be eliminated, leaving only the relevant terms. Finally, some techniques do not require a predefined ansatz and instead navigate the exponentially large space of analytical expressions (e.g.\ AIFeynman\,\cite{aifeynman, aifeynman2} for functions).

When solving model discovery tasks, SciExplorer improves upon these techniques in several qualitative ways:
SciExplorer can control experiments, thereby integrating an active learning component. Moreover, SciExplorer has physical intuition like a human researcher, which it uses to navigate the exponentially large space of possible models. SciExplorer operates with minimal prior information and does not require a predefined ansatz. However, when additional information about the model is known, it can easily be incorporated as plain text. Finally, SciExplorer not only addresses symbolic regression tasks but can also tackle more general challenges, such as program discovery.
When appropriate, SciExplorer uses symbolic regression techniques itself. In fact, after inferring a suitable ansatz from the qualitative signatures of an unknown system, SciExplorer often employs sparse regression as part of its discovery process, implementing it from scratch.

We provide a quantitative comparison between SciExplorer and the symbolic regression techniques SINDy, AIFeynman, and PDEFIND in \Cref{app:comp_reg}. We find that even when a user manually carries out the additional steps required by these techniques (selecting experiments, hyperparameters, and ansatz functions), SciExplorer discovers more accurate and interpretable models.

\section{Conclusion}
We have demonstrated that SciExplorer can successfully address open-ended, research-style tasks by interacting with physical systems through repeated cycles of experimentation and analysis. The LLM at the core of SciExplorer leverages broad knowledge across diverse areas of physics, such as classical mechanics, wave dynamics, and quantum systems, to autonomously characterize systems in a single-shot manner, i.e.\ without example solutions and with only minimal task- or system-specific instructions. Equipped with generic, code-based analysis and visualization tools, the agent can implement an active learning process to solve tasks such as discovering equations of motion from observed dynamics and recovering Hamiltonians, or even Hamiltonian families, from expectation values.

In physics, many modern experiments are controlled through code-based interfaces, which makes SciExplorer directly applicable in experimental settings such as complex fluids, cold atomic gases, strongly correlated electronic and spin systems, or quantum simulators. While model discovery remains highly relevant in these systems, SciExplorer could, for example, also map out phase diagrams or optimize control tasks in unknown or partially known systems.

The iterative process of experiment selection, analysis, and hypothesis generation is not unique to physics but lies at the heart of the natural sciences more broadly. Because LLMs possess substantial knowledge across scientific domains such as chemistry\,\cite{2024JablonkaLeveragingChemistry, 2025HerckAssesmentChem} and biology\,\cite{2019LeeBioBert, 2022LuoBioGPT}, and because SciExplorer requires no task-specific fine-tuning, applying this framework to other scientific fields is a natural next step, for example, to study predator-prey dynamics in biology or chemical nonequilibrium dynamics.

\section*{Data and code availability}
The complete SciExplorer framework, including documentation on adding new experiments and analysis tools, is available as a Python package on GitHub\,\cite{SciExplorerFramework}. Additionally, we provide all code used to run the explorations presented in this article, logs of all explorations, and tools to print or summarize those explorations in a second GitHub repository\,\cite{SciExplorerResults}.

\section*{Acknowledgments}
This research is part of the Munich Quantum Valley, which is supported by the Bavarian state government with funds from the Hightech Agenda Bayern Plus.

\section*{Appendix}
\appendix

\renewcommand{\thefigure}{A\arabic{figure}}
\setcounter{figure}{0}
\makeatletter
\renewcommand{\theHfigure}{A\arabic{figure}} 
\makeatother
\FloatBarrier

\section{Details on the SciExplorer}\label{app:SciExplorer}

After supplying the generic and rather brief system prompt (identical throughout all experiments), the LLM agent receives a short description of the problem class (e.g.\ a dynamical system or a spin system) and the task (e.g.\ discovering equations of motion or the Hamiltonian). In addition, as for any application of agentic tool use, the LLM has access to the documentation of the tools it can access. To facilitate the efficient integration of new analysis tools and experiments into the SciExplorer framework, we automatically generate API-compatible documentation by parsing the docstrings of tool methods.
We list all such inputs to the LLM (prompts, problem class description, task description, tool documentation) in \Cref{app:prompts} and the Supplementary Material\,\cite{supplement}. 

In the following steps of the automated conversation, the agent receives only a hint about the number of remaining conversation steps and tool calls, and a reminder to heed the system prompt.

In all explorations, the agent has access to two generic core tools:
\texttt{execute\_code} for executing generic Python code and \texttt{plot\_from\_code} for creating Matplotlib plots.

The results of each tool call are saved to the external memory under a result label, which the agent is automatically asked to specify to enable descriptive names. These results are then accessible as local variables in the \texttt{execute\_code} and \texttt{plot\_from\_code} tools.
The tool results are communicated to the agent as plain text for arrays with fewer than 10 entries. Otherwise, only the shape and type are returned to prevent excessively large context windows. If a tool call produces an image, it is returned to the agent for visual inspection.

The agent can choose to finish the exploration at any time by calling the \texttt{save\_result} tool. It is then asked to announce its result in the form of Python code that defines the solution hypothesized by the agent (see below).

After a given run of the agent, we save the entire conversation for further inspection.

\section{Mechanical systems}\label{app:Mechanics}
We use a differential equation solver to solve the equations of motion for the experimental runs of the system that is unknown to the agent.
To run an experiment, the agent can specify initial conditions, a start and end time, and the time resolution to receive generalized coordinates and velocities of the system evaluated on a time grid using the \texttt{observe\_experiment} tool. We allow the agent to observe up to 5 experiments within a single tool call to reduce the steps needed to explore the system.

When the agent completes its explorations, it submits the hypothesized model in the form of a Python function \texttt{predict\_experiment} that should reproduce the \texttt{observe\_experiment} function of the unknown system, which typically includes code to solve an ordinary differential equation (ode) and its right-hand side (rhs). However, we never inform the agent that the system is governed by an ODE, and it could, in principle, submit Python code implementing any model.

To evaluate the performance of the agent, we require a single continuous value describing how `close' the agent's prediction is to the truth. We estimate the coefficient of determination $$R^2(\vec{x}, \vec{y}) = 1 - \frac{\sum_i \left(y_i -x_i\right)^2}{\sum_i\left(\bar x-x_i\right)^2}$$
between the true ($\vec x$, with mean $\bar x$) velocities and accelerations of the system (given by the rhs of its ODE) and their prediction by the agent's \texttt{predict\_experiment} function ($\vec y$, estimated from two consecutive data points by choosing a small time resolution). We sample 1000 random position-velocity-time combinations and then compute $R^2$ for each coordinate and velocity separately and average them to obtain a single metric. The $R^2$ value can range from $-\infty$ (bad fit) over $0$ (as good as predicting $\vec y = \mathrm{mean}(\vec x)$) to $1$ (perfect fit). Therefore, we lower bound this value by zero to enable reasonable averaging of multiple runs. For partially observable systems, this method is not applicable, since correctly predicting the evolution of the hidden particles is a crucial part of the task and we cannot just sample random positions in phase space.
Instead, we simulate 100 trajectories with both the true dynamics and the agent's prediction with random initial conditions of the observed particles. Finally, we compute the mean $R^2$ of the trajectory of the observed particle between the true and predicted dynamics, i.e.\ we ask the agent to reproduce the observed dynamics, but do not require it to correctly reproduce the hidden dynamics.

\section{Fields and waves}\label{app:Fields}
We set up a split-step integrator for the evolution of a complex field $\phi$ on a 1d lattice, providing the experimental results for a hidden evolution equation. The agent specifies start and end time, time resolution and initial conditions using JAX code\,\cite{jax} (\texttt{observe\_experiment} tool). As above, the agent saves its model in the form of a \texttt{predict\_experiment} function which should reproduce the experimental results.

To evaluate the performance of the agent, we proceed as follows: We first run the wave evolution for different reasonable initial conditions (multiple Gaussian wave packets with and without momentum and with different amplitudes) to obtain a large set of physically relevant field or wave functions $\phi(x, t)$. We then numerically calculate $\frac{d \phi(x, t)}{dt}$ for all initial conditions and all times $t$ under both the true model and the agent's prediction and calculate the $R^2$ value between them.

\begin{figure*}[ht]
    \centering
    \includegraphics[width=0.65\textwidth]{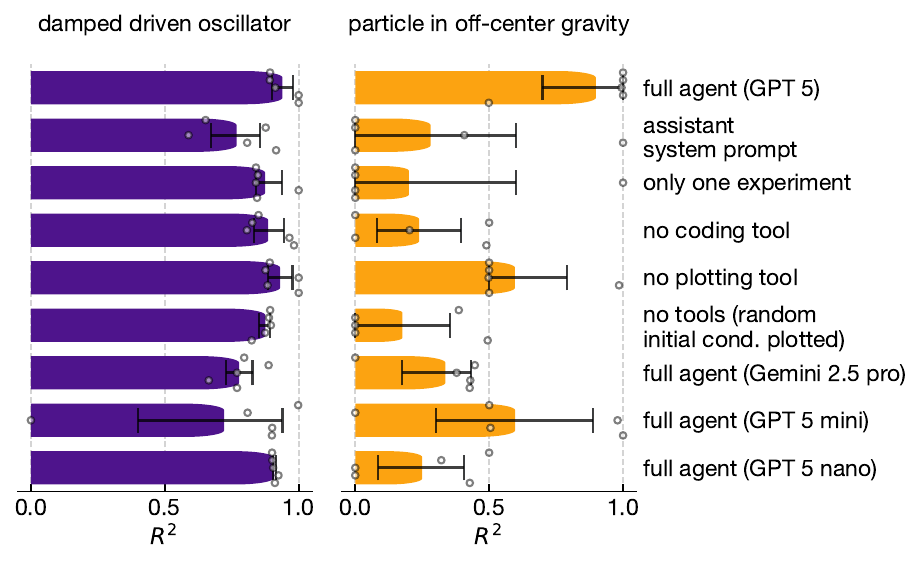}
    \caption{{\bf Ablation studies.} Mean performance of the agent in the damped driven oscillator and the particle in off-center gravity task with different modifications. Error bars are 95\% bootstrap confidence intervals. \emph{Full agent (GPT~5)}: Original SciExplorer with GPT~5\,\cite{gpt5} as an LLM backbone. \emph{Assistant system prompt}: Changed system prompt to 'You are a helpful assistant'. \emph{Only one experiment}: Agent can observe only one trajectory. \emph{No coding tool}: Agent cannot execute arbitrary Python code, but only plot experimental results. \emph{No plotting tool}: The agent cannot generate visualizations. Instead, down-sampled experimental trajectories are stored in the agent's context.  \emph{No tools}: The agent cannot use any tools. In the beginning, it is instead provided with time evolution and phase space plots of ten trajectories with random initial conditions.
    \emph{Full agent (Gemini 2.5 pro)}: Original SciExplorer with Gemini 2.5 pro\,\cite{gemini} as an LLM backbone.
    \emph{Full agent (GPT~5 mini)}: Original SciExplorer with GPT~5 mini as an LLM backbone.
    \emph{Full agent (GPT~5 nano)}: Original SciExplorer with GPT~5 nano as an LLM backbone.}
    \label{fig:ablation}
\end{figure*}

\section{Quantum many-body physics systems}\label{app:Quantum}
We consider a system of $N$ spin-1/2 degrees of freedom (qubits). We predefine the Pauli matrices ${\hat S}_x^j$, ${\hat S}_y^j$, and ${\hat S}_z^j$.
These matrices can be used to define Hamiltonians or operators to observe. An example of such code looks like the following, for the transverse quantum Ising model:
\begin{lstlisting}[language=Python]
H=0*Sz[0]
for j in range(10):
    H+=Sx[j] @ Sx[j+1]
for j in range(10):
    H-=0.5*Sz[j]
\end{lstlisting}
Depending on the scenario, the agent can either retrieve ground-state expectation values of previously specified operators (using the \texttt{set\_operator} and \texttt{observe\_experiment} tools) or observe the dynamics of a (hidden) Hamiltonian by specifying an initial product state of the spins (\texttt{set\_blochvectors} and \texttt{observe\_experiment} tools). In the latter case, we also consider a more challenging scenario in which the agent is only allowed to observe and initialize a selected subset of spins, while all other spins are set to a default state in each experimental run.

For both ground-state and time-dependent experiments, we also consider scenarios where the Hamiltonian depends on one or more tunable parameters (as in ${\hat H}=\sum_j {\hat {\sigma}^x_j} {\hat {\sigma}^x}_{j+1} - A \sum_j {\hat {\sigma}^z_j}$), which the agent can set  before running an experiment. This allows model discovery of a whole parametrized family of Hamiltonians. In a variant of this scenario, the agent may also choose the number $N$ of spins in the experimental system, e.g.\ to explore finite-size effects in translationally invariant models.

\begin{figure}[t]
    \centering
    \includegraphics[width=0.4\textwidth]{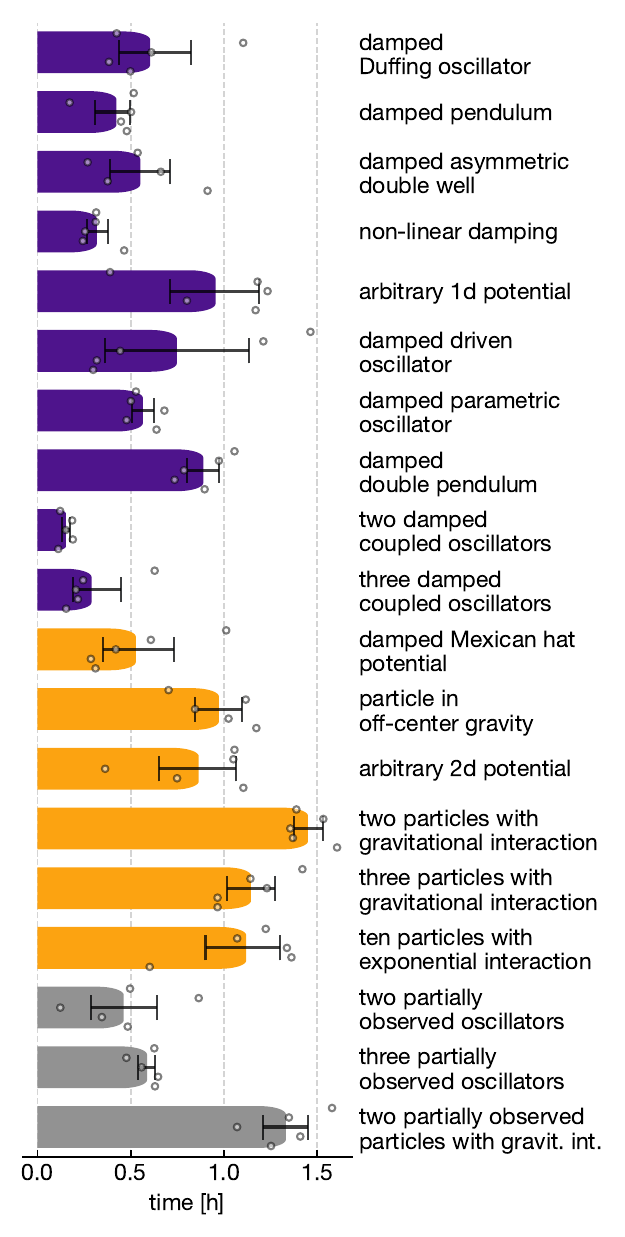}
    \caption{{\bf Run-time analysis.} Run-time of the classical mechanics explorations shown in Figure~3 of the main text in hours. Run times range from a few minutes to over 1.5 hours.}
    \label{fig:runtim_mechanics}
\end{figure}

To save its result, the agent is asked to provide the Hamiltonian of the system in the format described above.
In the ground-state experiments, we evaluate the agent's performance by calculating the fidelity per spin between the ground state predicted by the agent and the true ground state, i.e.\ $\left|\langle \psi_{\mathrm{agent}}|\psi_{\mathrm{true}}\rangle\right|^{2/N}$, where $N$ is the number of spins in the system. This quantity is defined so as to become system-size independent in the thermodynamic limit of large $N$.
For the dynamics tasks, we first shift the Hamiltonians by $H_i^\prime = H_i - \mathrm{tr}(H_i)/2^N$ and then compute the scalar product of these shifted Hamiltonians as $\mathrm{tr}\left[(H_{\mathrm{true}}^\prime)^\dagger H_{\mathrm{agent}}^\prime\right] /\max\left(||H_{\mathrm{true}}^\prime||,||H_{\mathrm{agent}}^\prime||\right)^2$, where $||.||$ denotes the Frobenius norm $||A||=\sqrt{{\rm tr}\left[A^\dagger A\right]}$. Taking the maximum of both norms in the denominator ensures that multiplying the predicted Hamiltonian by a scalar, thereby effectively rescaling time, results in a lower quality score.
For tasks with tunable parameters, we average the quality score over 100 random values of these parameters.

\section{Further analysis of the SciExplorer}\label{app:ablation}

To assess which components of the SciExplorer are essential, we performed ablation experiments on two representative systems: the damped driven oscillator and the particle in off-center gravity (\Cref{fig:ablation}). The relative importance of individual components differs between these systems. A structured system prompt appears to improve performance in both cases, whereas access to multiple experiments and visualization tools is particularly critical in the gravity system. Without the coding tool, the agent sometimes recovers the correct qualitative model but consistently fails to identify accurate numerical parameters. Without any access to tools and instead provided with plots of trajectories under random initial conditions (matching the number of experiments selected by the full agent), the LLM is unable to identify accurate models.

Finally, we observe substantial variation across LLMs: GPT~5\,\cite{gpt5} dramatically outperforms Gemini 2.5 pro\,\cite{gemini}, which fails to recover the true models even when granted access to the full SciExplorer framework. Similarly, GPT~5 nano fails to find the true models in all attempts. GPT~5 mini performs slightly better, recovering the true model in one of five attempts per system.

\section{Prompts}\label{app:prompts}
 Throughout all tasks and physics scenarios, we use the same generic and relatively brief system prompt, essentially instructing the LLM to act like a cautious scientist:
\begin{tcolorbox}[enhanced, breakable, title=System prompt]
\begin{itemize}
    \item Act as a computational physicist dedicated to thoroughly resolving the user's query through careful planning, hypothesis generation, and iterative verification.
    \item In your first message, create a comprehensive plan to solve the users query. Include an extensive list of candidate hypotheses.
    \item Initially, conduct at least 5 different experiments spanning the entire range of reasonable initial conditions. Make sure to cover also extreme cases. Then, create informative plots of your experimental results.
    \item Withhold any final answer until you are sure that no further improvements of your hypothesis are possible.
    \item Before submitting your final answer, simulate your proposed model using the same initial conditions as in your experiments, and compare the results. Only submit your final answer if the simulation results closely match your experimental data.
    
    \item If you have run a tool but still need to extract the results (e.g. via visualization), just briefly explain what tool you will call next to extract the results.
    \item Otherwise, at each step, you must answer the following questions:\\
            1. What can you learn from the new tool results (if any)? \\
            2. Which old hypotheses still fit your data? \\
            3. Which new hypotheses might be worthwhile considering? \\
\end{itemize}
\end{tcolorbox}

Even though we ask the agent to answer the questions above at each step, it does not reliably do so. However, summaries of the internal thinking process of the agent indicate that it considers the questions internally.
We use the following intermediate user message after each response from the agent:
\begin{tcolorbox}[enhanced, breakable, title=Intermediate message]
Keep solving the problem while following your system prompt.
\end{tcolorbox}

\begin{figure*}[t]
    \centering
    \includegraphics[width=\textwidth]{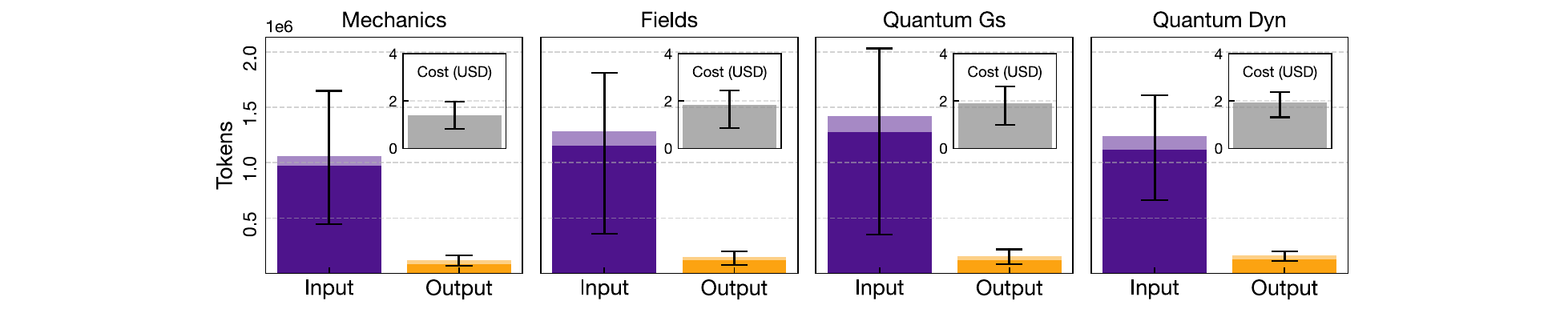}
    \caption{\textbf{Token usage and cost.} Average tokens used in a single exploration for the different domains. Violet bars are input tokens (dark: cached from previous requests, bright: new input), yellow bars are output tokens (dark: thinking tokens, bright: text output tokens). The inset shows the average API costs per exploration for GPT~5. Error bars denote 25 and 75 percentiles across all experiments in the domain.}\label{fig:token_usage}
\end{figure*}

\section{Resource requirements}

In \Cref{fig:runtim_mechanics}, we show the run-time for all explorations of the mechanical systems which typically ranges from a few minutes to over 1.5 hours.

As shown in \Cref{fig:token_usage}, running SciExplorer with a state-of-the-art LLM (in this case, GPT~5) costs around two USD per exploration on average. Although higher than typical specialized machine learning algorithms (e.g., AIFeynman or SINDy), the cost is small compared to the hourly salary of a PhD student in theoretical physics, who would be needed to fully replicate SciExplorer’s capabilities.

\section{Comparison to standard symbolic regression techniques}\label{app:comp_reg}
In the following, we compare SciExplorer to the standard symbolic regression techniques SINDy and AIFeynman\,2 for the mechanical systems and PDEFIND for field systems.

\subsection{SINDy and AIFeynman\,2 for mechanical systems}\label{app:mech_comp}

\begin{figure*}[t]
\centering
\includegraphics[width=0.98\textwidth]{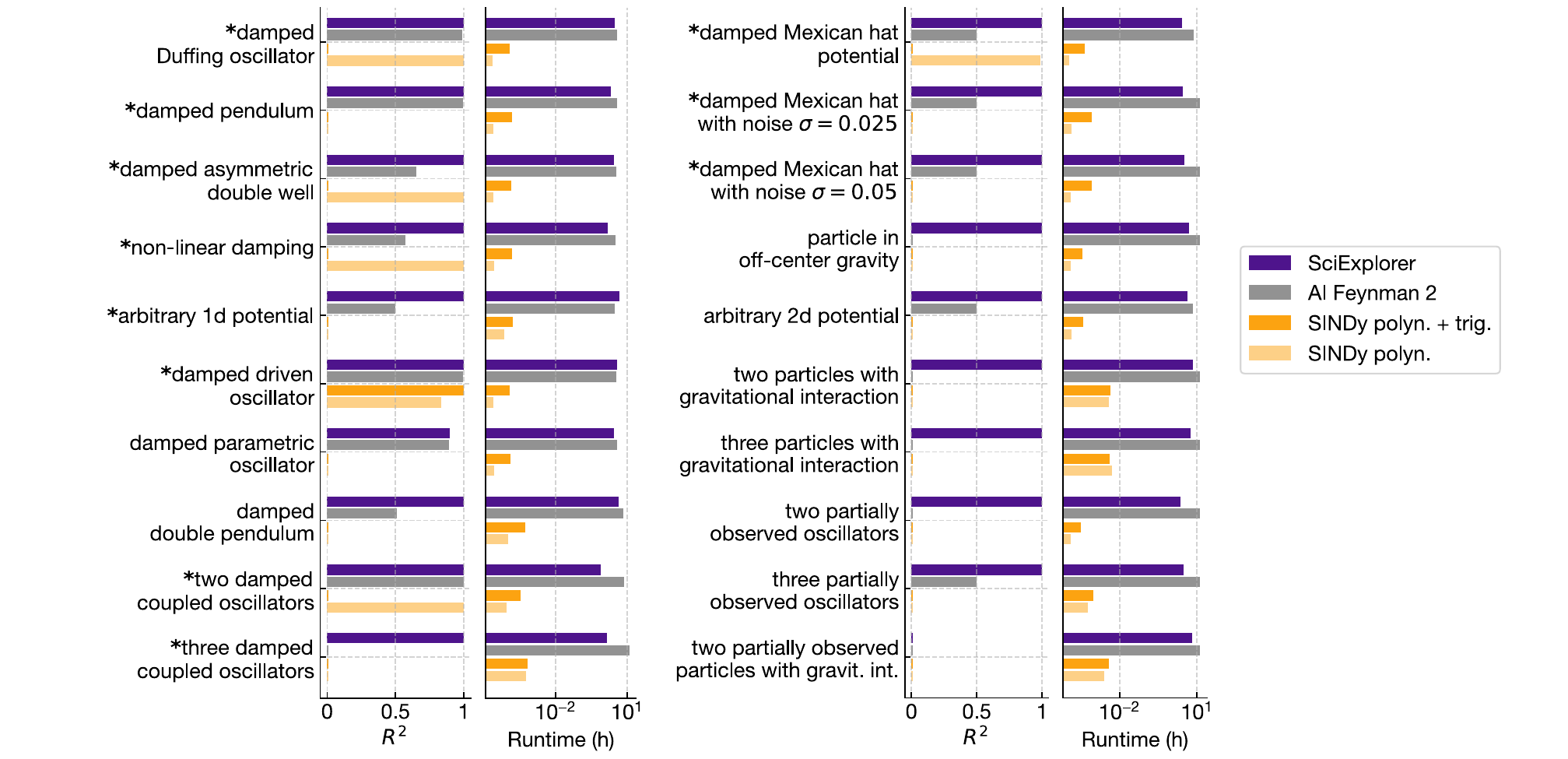}
\caption{{\bf Comparison to SINDy and AIFeynman\,2.} Performance and run-time of SciExplorer (best of 5), AIFeynman\,2, and SINDy with two different function libraries (see main text). Stars mark systems that could, in principle, be exactly recovered with the function libraries provided to SINDy. SINDy requires substantially less run-time but cannot recover many of the models SciExplorer finds. AIFeynman\,2 requires a similar run-time but also falls significantly short of SciExplorer in fit quality. Missing bars indicate an $R^2<0$ or, in the case of AIFeynman\,2, that no result was produced due to the divergence of the neural network trained during its discovery attempt.
The run-time of SciExplorer is the sum of the five independent explorations we ran per system. It could be further reduced by parallelizing these runs. Similarly, AIFeynman\,2 could be sped up by parallelizing over the individual dimensions of the ODE and by running on a GPU.}\label{fig:mech_comp}
\end{figure*}

\begin{figure*}[t]
\centering
\includegraphics[width=0.98\textwidth]{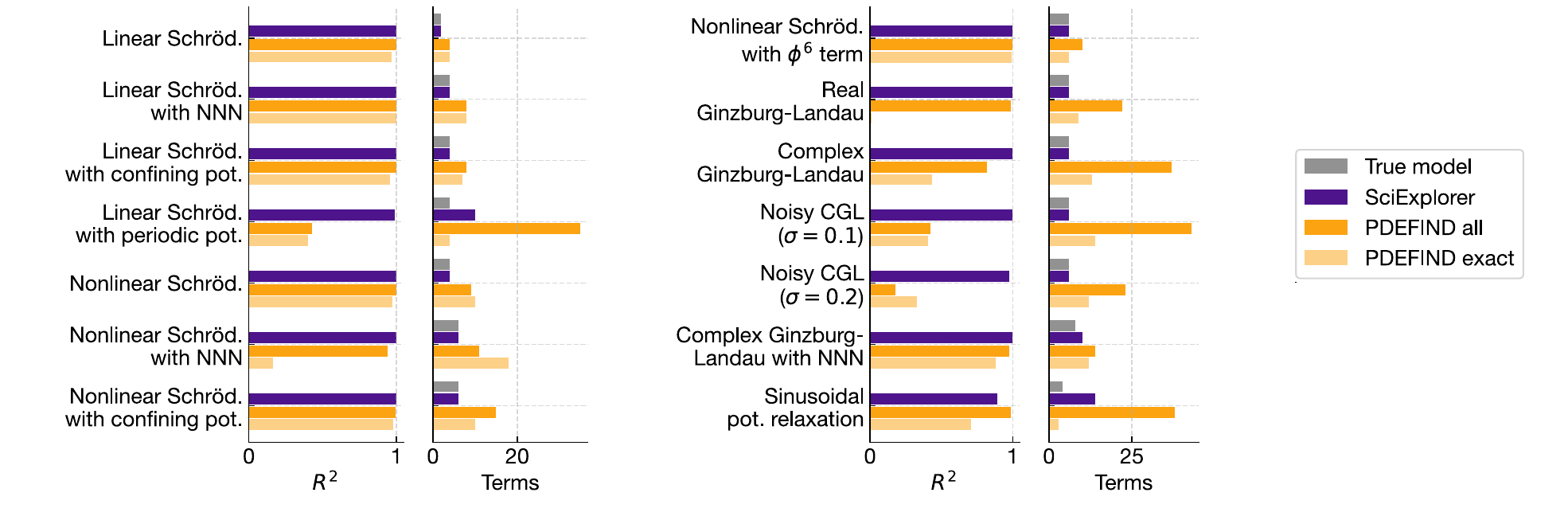}
\caption{{\bf Comparison to PDEFIND.}
Performance and number of terms in the recovered partial differential equation for SciExplorer (best of 5), and PDEFIND for a general ansatz library (all) and a library specialized for each system (exact). SciExplorer consistently finds more accurate models with fewer terms than PDEFIND. }\label{fig:field_comp}
\end{figure*}

SINDy\,\cite{brunton16sindy} is a technique specifically developed to discover ordinary differential equations (odes) from data. It requires the user to propose a set of ansatz functions that may appear in the right-hand side (rhs) of the system's ODE. It then uses sparse linear regression on the system’s acceleration to find the numerical prefactors of these terms. The loss function combines the mean squared error between the true and predicted acceleration with an L1 regularization term on the prefactors of the terms in the predicted rhs. This L1 loss is weighted by a hyperparameter $\alpha$, with higher values promoting sparser solutions. We use the implementation PySINDy\,\cite{pysindy} with a sequentially thresholded Ridge regression optimizer, iteratively setting values below 0.01 to zero. The threshold is chosen such that no constants present in the true models are eliminated (the smallest one being 0.043). The performance of SINDy strongly depends on the sparsity penalty $\alpha$. Therefore, we select the optimal $\alpha \in [0.01, 0.05, 0.1]$ for each experimental system. As the ansatz, we consider two different function libraries. First, we consider a purely polynomial library containing polynomials up to degree three (polyn.). Second, we add $\sin$ and $\cos$ functions with frequencies scanned from 0.5 to 6 in steps of 0.5 to the polynomial library. To even further reduce the difficulty for SINDy, we also include the exact drive frequency for the driven oscillators $\omega=1.551$ (polyn.\ + trig.). These two libraries are chosen since they provide a good trade-off between the number of ansatz functions and the number of systems that could be fitted by them. We fit ten random trajectories of the system with 20,001 time points each. We compare the fit quality and run-time of SINDy with SciExplorer in \Cref{fig:mech_comp}. Although SINDy is much faster than SciExplorer, it can recover only a small subset of solutions with high accuracy. Even some systems whose rhs could be exactly fitted with the provided function libraries (marked by stars) are not recovered.

AIFeynman\,2 is a symbolic regression technique that fits a neural network to the provided data and uses physics-inspired methods, such as testing for translational invariance and symmetries, to simplify the regression task. 
AIFeynman\,2 can recover only the analytical expression of a function rather than an ODE. Therefore, we test how well it can fit the rhs of the mechanical systems' ODEs. We provide the value of the exact rhs at 200,010 points to AIFeynman\,2, which would normally have to be estimated from data, thereby avoiding additional errors that would arise from estimation. A known problem of AIFeynman\,2 is the divergence of its neural network during training. To alleviate this problem, we ran AIFeynman\,2 twice for each system. Once with the direct experimental data and once with the data normalized to zero mean and unit standard deviation, selecting the better of the two results.
We find that AIFeynman,2 can accurately recover only the simplest systems considered (see \Cref{fig:mech_comp}).

Both SINDy and AIFeynman\,2 have hyperparameters and require choices such as the selection of ansatz functions. Our experiments show that even with considerable human effort to choose suitable parameters and algorithms, these techniques do not match SciExplorer in fit quality. However, we do not claim that SINDy is incapable of finding the correct model, provided the correct hyperparameters and a reduced set of suitable ansatz functions. In fact, SciExplorer often employs SINDy-style techniques after inferring an appropriate ansatz from the qualitative signatures of the system's dynamics.

\begin{figure*}[t]
\centering
\includegraphics[width=0.98\textwidth]{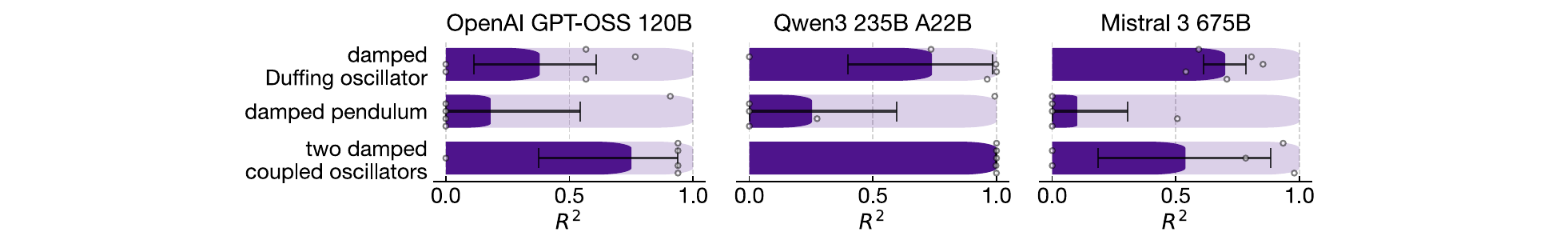}
\caption{{\bf Experiments with open-source models.} Performance of three open-source models on selected mechanical systems. All models struggle even with the easier tasks considered in the main text. Shaded bars indicate the performance of GPT~5, which finds perfect models in all five attempts per system.}\label{fig:opensource}
\end{figure*}

\subsection{PDEFIND for field systems}\label{app:field_comp}

PDEFIND\,\cite{PDEFIND} generalizes SINDy to partial differential equations. As above, we use the PySINDy implementation with a sequentially thresholded Ridge regression optimizer. PDEFIND's performance strongly depends on its hyperparameters.
Therefore, we scan both the L1 loss $\alpha \in [0.01, 0.05, 0.1]$ and the threshold $\in [0.00001, 0.0001, 0.001, 0.01, 0.1]$. We use two function libraries. An exact library tailored to each model, containing only the functions necessary to fit the true underlying model, along with the correct order of spatial derivatives (exact), and a fixed library capable of fitting all models (all). We fit trajectories of four representative initial conditions, using numerical parameters slightly adapted from those used to assess fit quality. Each run of PDEFIND takes only tens of seconds. However, it frequently fails to recover a well-fitting model and often produces models with more terms than required, reducing their interpretability (see \Cref{fig:field_comp}). We stress that we considered a scenario that already requires substantial additional human effort compared to running SciExplorer, i.e.\ identifying a small set of accurate ansatz functions.

\section{Experiments with open-source models}\label{app:opensource}

The experiments in the main text use GPT~5 as the state-of-the-art backbone of SciExplorer. In the long term, it would be desirable to base SciExplorer on open-source models. To this end, we integrate the OpenAI Chat API, which is compatible with many open-source models, into SciExplorer. We present initial experiments with the open-source models OpenAI GPT-OSS 120B (supports reasoning, no vision), Qwen3 235B A22B (supports reasoning, no vision), and Mistral 3 675B (no reasoning, supports vision). Unfortunately, these models already struggle with even the simplest tasks presented in the main text (see \Cref{fig:opensource}). A common problem of GPT-OSS is not following the user's instruction, e.g.\ calling the \texttt{save\_result} in the first step without any information. Mistral~3 can visualize trajectories but often hallucinates image content. While it can write code to fit experimental trajectories, it typically proposes overly simple models and uses inaccurate initial guesses for the fits. Qwen3 performs better than the other open-source models and occasionally employs advanced techniques, such as sparse linear regression of the acceleration, to recover the correct model. Nevertheless, it still often hallucinates, for instance, submitting models with incorrect numerical constants even when its fits returned the true values.
These results suggest that developing a competitive multimodal LLM with robust reasoning capabilities should be a priority for the open-science community.

\FloatBarrier
\bibliography{main}
\FloatBarrier
\clearpage
\def\MAINFILE{}
\ifdefined\MAINFILE
    \renewcommand{\appendixname}{}
    \renewcommand{\thesection}{\arabic{section}}
    \renewcommand{\thesubsection}{\Alph{subsection}}
    \renewcommand{\thesubsubsection}{\arabic{subsubsection}}
    \onecolumngrid
    \begin{center}
    {\Large\bfseries Supplementary Material: Agentic Exploration of Physics Models}\\[2ex]
    {\normalsize Maximilian Nägele\,\orcidlink{0000-0001-6382-2077} and Florian Marquardt\,\orcidlink{0000-0003-4566-1753}}\\
    {\small \textit{Max Planck Institute for the Science of Light, Staudtstra{\ss}e 2, 91058 Erlangen, Germany}}\\
    {\small \textit{Department of Physics, Friedrich-Alexander Universit\"{a}t Erlangen-N\"{u}rnberg, Staudtstra{\ss}e 5, 91058 Erlangen, Germany}}\\[2ex]
    \end{center}
\else
    \documentclass[twocolumn, aps, prx, 10pt, floatfix]{revtex4-2}
    \usepackage{graphicx} 
    \usepackage{parskip}
    \usepackage{amsmath}
    \usepackage{booktabs}
    \usepackage{appendix}
    \usepackage{placeins}
    \usepackage{markdown}
    \usepackage{makecell}
    \usepackage{longtable}
    
    \usepackage{xcolor}
    \definecolor{codepurple}{rgb}{0.58,0,0.82}
    \definecolor{red}{HTML}{E63946}
    \definecolor{blue}{HTML}{457B9D}
    \definecolor{grey}{HTML}{F0F0F0}
    \definecolor{yel}{HTML}{fca311}
    \definecolor{purp}{HTML}{4E148C}
    
    \usepackage{listings}
    \lstdefinestyle{mystyle}{
        backgroundcolor=\color{grey}, 
        commentstyle=\color{purp},
        keywordstyle=\color{yel},
        stringstyle=\color{purp},
        basicstyle=\ttfamily\footnotesize,
        breakatwhitespace=false,         
        breaklines=true,                 
        captionpos=b,                    
        keepspaces=true,                 
        showspaces=false,                
        showstringspaces=false,
        showtabs=false,                  
        tabsize=2,
        deletekeywords={abs},
        deletekeywords=[2]{abs},
    }
    \lstdefinestyle{supstyle}{
        backgroundcolor=\color{grey}, 
        commentstyle=\color{black},
        keywordstyle=\color{black},
        stringstyle=\color{black},
        basicstyle=\ttfamily\footnotesize,
        breakatwhitespace=false,         
        breaklines=true,                 
        captionpos=b,                    
        keepspaces=true,                 
        showspaces=false,                
        showstringspaces=false,
        showtabs=false,                  
        tabsize=2
    }
    \lstset{style=mystyle}
    
    \usepackage{tcolorbox}
    \tcbuselibrary{breakable,skins}
    
    \usepackage{hyperref}
    \hypersetup{
        colorlinks,
        linkcolor={black!50!black},
        citecolor={black!50!black},
        urlcolor={black!80!black},
    }
    \usepackage{orcidlink}
    \usepackage{cleveref}
    \crefname{section}{supplement}{supplements}
    
    \setlength\fboxsep{1pt} 
    \setlength\fboxrule{0.4pt} 
    
    \bibliographystyle{unsrtnat}
    
    \renewcommand{\doi}[1]{DOI: #1} 

    \begin{document}

    \title{Supplementary Material: Agentic Exploration of Physics Models}
    \author{Maximilian Nägele\,\orcidlink{0000-0001-6382-2077}}
    \email{maximilian.naegele@mpl.mpg.de}
    \affiliation{Max Planck Institute for the Science of Light, Staudtstra{\ss}e 2, 91058 Erlangen, Germany}
    \affiliation{Department of Physics, Friedrich-Alexander Universit\"{a}t Erlangen-N\"{u}rnberg, Staudtstra{\ss}e 5, 91058 Erlangen, Germany}
    \author{Florian Marquardt\,\orcidlink{0000-0003-4566-1753}}
    \affiliation{Max Planck Institute for the Science of Light, Staudtstra{\ss}e 2, 91058 Erlangen, Germany}
    \affiliation{Department of Physics, Friedrich-Alexander Universit\"{a}t Erlangen-N\"{u}rnberg, Staudtstra{\ss}e 5, 91058 Erlangen, Germany}
    
    \maketitle

    \newcommand{\todo}[1]{\textcolor{orange}{[#1]}}

\fi

\onecolumngrid
\makeatletter
\setcounter{section}{0}
\setcounter{figure}{0}
\renewcommand \thesection{S\,\@arabic\c@section}
\renewcommand\thetable{S\@arabic\c@table}
\renewcommand \thefigure{S\@arabic\c@figure}
\makeatother

\section{Details on analysis tools}\label{sup:analysis_tools}
\lstset{style=supstyle}
For all tasks and scenarios, we provide the following generic analysis tools to the agent, exploiting the proven power of LLMs in coding. We here display the documentation strings, which are the only information available to the LLM agent:
\begin{tcolorbox}[enhanced, breakable,title=
Analysis: code-based plotting]
\begin{lstlisting}[language=Python]
def plot_from_code(code: str):
    Execute python code that produces a plot from one or more previously saved arrays.
    The following variables are available during evaluation:
        all previosuly saved fields with their previously stated result_key as variable names
        (as global variables).
    You may use the following libraries:
        matplotlib: for plotting.
        matplotlib.pyplot as plt: for plotting.
        jax: for numerical operations.
        jax.numpy: for numerical operations.
        numpy: for numerical operations.
    Args:
        code: python code that produces a plot (without a plt.show() call!).
    Returns:
        Image
\end{lstlisting}
\end{tcolorbox}

\begin{tcolorbox}[enhanced, breakable,title=
Analysis: code based analysis]
\begin{lstlisting}[language=Python]
def execute_code(code: str):
    Evaluate python code.
    This code can be e.g. be used to transform the previously saved fields or to
    calculate or save new fields.
    You can not see any plots created with this tool.
    The following variables are available during evaluation:
        all previosuly saved fields with their previously stated result_key as variable
        names (as global variables).
        jax: jax for numerical operations.
        jnp: jax.numpy for numerical operations.
        np: numpy for numerical operations.
        scipy: scipy for numerical operations including optimization and solving
        differential equations.
        sklearn: scikit-learn.
    IMPORTANT: Your code must set the variable 'result' to a dictionary in the end which
    should contain the newly generated data, for example: result={'<result_key>': <data>,
    ...}.
    Args:
        code: python code that sets the result variable to a dictionary containing some newly
        generated data.
    Returns:
        The result dictionary.
\end{lstlisting}
\end{tcolorbox}

\section{Details on physical systems including experiment tools}\label{sup:details_syst_exp_tools}
Here, we state the governing equations of each physical system, the solutions discovered by the agent, and the experimental tools available to the agent when exploring the system. 
\subsection{Mechanical systems}\label{sup:details_mech}

The agent is asked to produces a Python function that reproduces the \texttt{observe\_experiment} tool.
Depending on the system type, we provide different descriptions of the system, experimental tools, and tools to save the predicted differential equation, as listed in detail below.

In short, for the generic dynamical systems, the agent is told their dimensionality, but coordinates are called 'generalized coordinates' and do not have descriptive names. For the systems with particles moving in 2d, the agent is told the number of particles and that they move in 2d. For the systems with hidden degrees of freedom, the agent is told that it can observe a particle moving in 1d/2d interacting with an unknown number of unobserved particles. 
For each system, we provide a tool where the agent can pass a list of up to 5 initial conditions to observe multiple trajectories with one tool call. The exact equations of motion of the systems are listed in \Cref{tab:eom_mechanics} and the solutions discovered by the agent in \Cref{tab:predicted_eom_mechanics}.

The task given to the agent is the same across all mechanics epxeriments:

\begin{tcolorbox}[enhanced, breakable, title=Task supplied to agent for mechanics tasks]
Can you find a model that reproduces the \texttt{observe\_experiment} function? After your exploration, save it using the \texttt{save\_result} function.
\end{tcolorbox}
For tasks involving noisy measurements, the agent is supplied with the description:
\begin{tcolorbox}[enhanced, breakable, title=Additional description in case of measurement noise]
The observations you are getting may be affected by measurement noise.
\end{tcolorbox}

\subsubsection{Generic dynamical systems}\label[subsubsection]{sup:gendyndescription}
This sections contain the description of the system and tools provided to the agent for the generic dynamical systems (damped Duffing oscillator, damped pendulum, damped asymmetric double well, non-linear damping, arbitrary 1d potential, damped driven oscillator, damped parametric oscillator, damped double pendulum, two damped coupled oscillators, three damped coupled oscillators).
\begin{tcolorbox}[enhanced, breakable, title=Description supplied to agent for generic dynamical systems]
You are investigating a dynamical physical system.
\end{tcolorbox}

Below we show the \texttt{observe\_experiment} and \texttt{save\_result} tools for the case of two generalized coordinates. In other cases, the docstrings are slightly adapted.
\begin{tcolorbox}[enhanced, breakable,title=
Experiment: observe evolution given initial conditions for generic dynamical systems]
\begin{lstlisting}[language=Python]
def observe_experiment(q_inits:str, q_dot_inits:str, t0:float=0.0, T:float=20.0,
    nt:int=20001):
    Observe multiple trajectories of the experiment with given initial conditions.
    The maximum number of trajectories is 5.
    The system has 2 generalized coordinates.
    Args:
        q_inits: string in the form of a list of initial generalized coordinates at time t0, 
        in the form [(q00, q10,), (q01, q11,), ..., (q04, q14,)],
        where qij is the i-th generalized coordinate for the j-th initial condition.
        q_dot_inits: string in the form of a list of initial generalized velocities at
        time t0, in the same form as q_inits.
        t0: float, initial time of the experiment (default 0.0)
        T: float, end time of the experiment (maximum 100.0, default 20.0)
        nt: int, number of time steps where to evaluate the experiment 
        (maximum 20001, default 20001).
    Returns a dictionary with:
        'ts':np.ndarray of shape [nt] with the time steps [t0, t0+dt, ..., T]
        (equal for all trajectories).
        'arrays':np.ndarray of shape [n_trajectories , nt, 4] with the solution.
        Here, the first half of the entries along the last axis correspond to the generalized
        coordinates in the same order as the input, and the second half of the entries
        correspond to the generalized velocities in the same order.
\end{lstlisting}
\end{tcolorbox}
The agent is asked to save its final prediction using the following tool:
\begin{tcolorbox}[enhanced, breakable,title=
Save result for generic dynamical systems]
\begin{lstlisting}[language=Python]
def save_result(code:str):
    Saves your final result.
    You can only call this once. Do not call it when you could potentially further improve your result!
    You should provide code that defines a predict_experiment function which should reproduce 
    trajectories of the experiment.
    The code has access to numpy (as np) and scipy (as scipy).
    Args: 
        code: String defining a function with the following signature:
            def predict_experiment(q_init: np.ndarray, q_dot_init: np.ndarray, t0: float,
                T: float, nt: int) -> Tuple[np.ndarray, np.ndarray]:
                Args (of predict_experiment):
                    q_init: np.ndarray of shape [2] with the initial generalized coordinates
                    at time t0, in the form np.ndarray([q0, q1]).
                    Here, qi is the initial condition for the i-th generalized coordinate.
                    q_dot_init: np.ndarray of shape [2] with the initial generalized 
                    velocities at time t0 in the same order.
                    t0: float, initial time of the experiment.
                    T: float, end time of the experiment.
                    nt: int, number of time steps where to evaluate the experiment.
                Returns (of predict_experiment):
                    Xs: np.ndarray of shape [nt, 4] with the solution. 
                        Here, the first half of the entries along the last axis correspond to 
                        the generalized coordinates in the same order as the input,
                        and the second half of the entries correspond to the generalized 
                        velocities in the same order.
                    ts: np.ndarray of length nt with the time steps [t0, t0+dt, ... , T].
\end{lstlisting}
\end{tcolorbox}

\subsubsection{Particles in 2d}\label{sup:singleparticles}
This section contains the description of the system and tools provided to the agent for the 2d systems with particles (damped Mexican hat potential, particle in off-center gravity, arbitrary 2d potential, two particles with gravity, three particles with gravity, ten particles with exponential potential).

\begin{tcolorbox}[enhanced, breakable, title=Description supplied to agent for particles in 2d]
You are investigating a physical system consisting of particles moving in two dimensions.
\end{tcolorbox}

We use slightly adapted \texttt{observe\_experiment} and \texttt{save\_result} tools. Below are examples for one particle.

\begin{tcolorbox}[enhanced, breakable,title=
Experiment: observe evolution given initial conditions for particles in 2d]
\begin{lstlisting}[language=Python]
def observe_experiment(pos_inits:str, vel_inits:str, t0:float=0.0, T:float=20.0,
    nt:int=20001):
    Observe multiple trajectories of the experiment with given initial conditions.
    The maximum number of trajectories is 5.
    The system contains 1 particle in 2D.
    Args:
        pos_inits: string in the form of a list of initial positions at time t0,
        in the form [(x0_0, y0_0,), (x0_1, y0_1,), ..., (x0_4, y0_4,)],where xi_j and yi_j 
        are the initial x and y positions of the i-th particle in the j-th trajectory.
        vel_inits: string in the form of a list of initial velocities at time t0,
        in the same form as pos_inits.
        t0: float, initial time of the experiment (default 0.0)
        T: float, end time of the experiment (maximum 100.0, default 20.0)
        nt: int, number of time steps where to evaluate the experiment
        (maximum 20001, default 20001).
    Returns a dictionary with:
        'ts':np.ndarray of shape [nt] with the time steps [t0, t0+dt, ..., T]
        (equal for all trajectories).
        'arrays':np.ndarray of shape [n_trajectories , nt, 4] with the solution.
        Here, the first half of the entries along the last axis correspond to the x and y
        positions in the same order as the input,
        and the second half of the entries correspond to the x and y velocities in the same
        order.
\end{lstlisting}
\end{tcolorbox}
\begin{tcolorbox}[enhanced, breakable,title=
Save result for particles in 2d]
\begin{lstlisting}[language=Python]
def save_result(code:str):
    Saves your final result.
    You can only call this once. Do not call it when you could potentially further improve your result!
    You should provide code that defines a predict_experiment function which should reproduce
    trajectories of the experiment.
    The code has access to numpy (as np) and scipy (as scipy).
    Args:
        code: String defining a function with the following signature:
            def predict_experiment(pos_init: np.ndarray, vel_init: np.ndarray, t0: float,
                T: float, nt: int) -> Tuple[np.ndarray, np.ndarray]:
                Args (of predict_experiment):
                    pos_init: np.ndarray of shape [2] with the initial positions at time t0,
                    in the form np.ndarray([x0, y0]). Here, xi and yi are the initial x and y
                    positions of the i-th particle.
                    vel_init: np.ndarray of shape [2] with the initial velocities at time t0
                    in the same order.
                    t0: float, initial time of the experiment.
                    T: float, end time of the experiment.
                    nt: int, number of time steps where to evaluate the experiment.
                Returns (of predict_experiment):
                    Xs: np.ndarray of shape [nt, 4] with the solution. 
                        Here, the first half of the entries along the last axis correspond to
                        the x and y positions in the same order as the input,
                        and the second half of the entries correspond to the x and y
                        velocities in the same order.
                    ts: np.ndarray of length nt with the time steps [t0, t0+dt, ... , T].
\end{lstlisting}
\end{tcolorbox}

\subsubsection{Systems with hidden degrees of freedom}
In case of hidden coordinates we add to the initial description
\begin{tcolorbox}[enhanced, breakable, title=Additional description supplied to agent for hidden generic dynamical systems]
You can only observe the first generalized coordinate of this system. However, there might be additional hidden generalized coordinates influencing the dynamics.
\end{tcolorbox}
or
\begin{tcolorbox}[enhanced, breakable, title=Additional description supplied to agent for hidden particles in 2d]
You can only observe the coordinates of one of the particles. However, there might be additional hidden particles influencing the dynamics.
\end{tcolorbox}

Since starting the experiment at a later time is not possible (as the hidden particles' positions are only fixed at $t=0$), we set the start- and end-time and time-resolution of the experiment. Everything else is kept as in the descriptions and docstrings above.

\begin{table}[h]
\centering
\begin{tabular}{cc}
\toprule
    \multicolumn{2}{c}{\bf Damped Duffing oscillator}  \\
    $\ddot x = -a x^3  - b x - \gamma \dot x$&
    $a = 4.528, b = 1.625, \gamma = 0.043$ \\
\midrule
    \multicolumn{2}{c}{\bf Damped pendulum}  \\
    $\ddot x = -\alpha\sin(x)- \gamma \dot x $&
    $\alpha=1.712, \gamma = 0.043$\\
\midrule
    \multicolumn{2}{c}{\bf Damped asymmetric double well}  \\
    $\ddot x = -a  x^3  + b x + c - \gamma \dot x$&
    $a=4.528, b = 1.625, c =0.1, \gamma = 0.043$\\
\midrule
    \multicolumn{2}{c}{\bf Velocity position coupling}  \\
    $\ddot x = -a x^2 \dot x - k x$&
    $a=1.7, k = 0.4, $\\
\midrule
    \multicolumn{2}{c}{\bf Arbitrary 1d potential}  \\
    $\ddot x = - x - a \cos(k x)$&
    $a=4.8, k =6$\\
\midrule
    \multicolumn{2}{c}{\bf Damped driven oscillator}  \\
    $\ddot x = -k x - \gamma \dot x + A \cos(\omega t)$&
    $k=2.319, \gamma = 0.6, A=1.712$, $\omega=1.551$\\
\midrule
    \multicolumn{2}{c}{\bf Damped parametric oscillator}  \\
    $\ddot x = -\left(k + A \cos(\omega t)\right)x - \gamma \dot x$&
    $k=2.319, A=1.712$, $\omega=1.551$, $\gamma = 0.3$\\
\midrule
    \multicolumn{2}{c}{\bf Damped double pendulum} \\
        $\Delta  = \theta_1 -\theta_2 $ & $m_1 = m_2 = 1$ \\
        $\ddot \theta_1 =-\frac{(2 m_1 + m_2) \sin(\theta_1) -
                   m_2 \sin(\theta_1 - 2 \theta_2) -
                   2 \sin(\Delta) m_2 \left(\omega_2^2 l_2 + \omega_1^2 l_1 \cos(\Delta)\right)}{l_1 \left(2 m_1 + m_2 - m_2  \cos(2 \Delta)\right)}-\gamma \dot \theta_1$ & $ l_1 = 1.712, l_2 = 0.851$ \\
        $\ddot \theta_2 = \frac{2 \sin(\Delta) \left(\omega_1^2 l_1 (m_1 + m_2) + (m_1 + m_2) \cos(\theta_1) +
                   \omega_2^2 l_2 m_2 \cos(\Delta)\right)}{l_2\left(2 m_1 + m_2 - m_2  \cos(2 \Delta)\right)} -\gamma \dot \theta_2$& $\gamma = 0.143$\\
\midrule
    \multicolumn{2}{c}{\bf Two damped coupled oscillators} \\
    $\ddot x_1 = -k_1 x_1 + k_{12} (x_2 - x_1) - \gamma  \dot x_1$&
    $k_1=1.712, k_{12} = 0.15$ \\
    $\ddot x_2 = -k_2 x_2 - k_{12} (x_2 - x_1) - \gamma  \dot x_2$&
    $\gamma = 0.043$ \\
\midrule
    \multicolumn{2}{c}{\bf Three damped coupled oscillators} \\
    \multicolumn{2}{c}{Analogous to two coupled oscillators,  $k_1 = 1, k_2 = 1.5, k3 = 0.5, k_{1,2}=0.2, k_{13} = 0.3, k_{23}=0.4$} \\
\midrule
    \multicolumn{2}{c}{\bf Damped Mexican hat potential}\\
    $\ddot x = \left(a - b \left(x^2+y^2\right)\right) x - \gamma \dot x$&
    $a = 4, b = 2.8$ \\
    $\ddot y = \left(a - b \left(x^2+y^2\right)\right) y - \gamma \dot y$&
    $\gamma = 0.2$ \\
\midrule
    \multicolumn{2}{c}{\bf Particle in off-center gravity} \\
    $r = \sqrt{(x-c_x)^2 + (y-c_y)^2 +\epsilon^2 }$&
    $c_x = -0.7, c_y = 0.2$ \\
    $\ddot x = - \frac{G (x-c_x)}{r^{3}}$, $\ddot y = - \frac{G(y-c_y)}{r^{3}}$ & 
    $G = 2.3$, $\epsilon=0.01$ \\
\midrule
    \multicolumn{2}{c}{\bf Arbitrary 2d potential}\\
    $r =\sqrt{(x^2 + y^2)}$&
    $k = 0.6, a=0.8$ \\
    $\ddot x = -k x - \frac{a x \cos(\omega r)}{r}$,  $\ddot y = -k y - \frac{a y \cos(\omega r)}{r}$ & $\omega = 6$\\
\midrule
    \multicolumn{2}{c}{\bf Two particles with gravity} \\
    $\Delta_x = x_1 - x_2, \Delta_y = y_1-y_2$, $r = \sqrt{\Delta_x^2 + \Delta_y^2 + \epsilon^2}$ & $m_1=8.123$, $m_2 = 0.781$ \\
    $\ddot x_1 = - \frac{m_2 \Delta_x}{r^{3}}$, $\ddot y_1 = - \frac{m_2 \Delta_y}{r^{3}}$,  $\ddot x_2 =  \frac{m_1 \Delta_x}{r^{3}}$,
    $\ddot y_2 =  \frac{m_1 \Delta_y}{r^{3}}$ & $\epsilon=0.01$ \\
\\
\midrule
    \multicolumn{2}{c}{\bf Three particles with gravity} \\
    \multicolumn{2}{c}{Analogous to two particles in gravity, $m_1=1.3, m_2=9.0, m_3=0.2, \epsilon=0.01$}\\
\midrule
    \multicolumn{2}{c}{\bf Ten particles with exponential potential}\\
    $r_{ij} = \sqrt{(x_i-x_j)^2+(y_i-y_j)^2}$& $a=0.8$\\ 
    $\ddot{x_i} = \sum_j -ab\frac{x_i - x_j}{r_{ij}}\exp(-b r_{ij})$, $ \ddot{y_i} = \sum_j -ab\frac{y_i - y_j}{r_{ij}}\exp(-b r_{ij})$ & $b = 1.3$\\
\midrule
   \multicolumn{2}{c} {\bf Two partially observed oscillators} \\
    \multicolumn{2}{c}{Analogous to two coupled oscillators, $k_1=1.1, k_2=1.54, k_{12}=1.2, \gamma =0$}\\
    \multicolumn{2}{c}{Hidden initial conditions, $x_2 = 0.4, \dot x_2 = -0.2$}\\
\midrule
    \multicolumn{2}{c}{\bf Three partially observed oscillators} \\
    \multicolumn{2}{c}{Analogous to three coupled oscillators, $k_1=0.611, k_2=1.907, k_3=1.473, k_{12}=1.317, k_{13} = 0.537,k_{23}=1.811,\gamma =0$}\\
    \multicolumn{2}{c}{Hidden initial conditions, $x_2 = -0.123, \dot x_2 = 0.895, x_3 = 0.068, \dot x_3 = 0.957$}\\
\midrule
    \multicolumn{2}{c}{\bf Two partially observed particles with gravity} \\
    \multicolumn{2}{c}{Analogous to two particles in gravity, $m_1=0.7, m_2=1.8$}\\
    \multicolumn{2}{c}{Hidden initial conditions, $x_2 = 0.5, y_2= 0.5, \dot x_2 = -0.5, \dot y_2 = 0$}\\
\bottomrule
\end{tabular}
\caption{Equations of motion for the considered mechanical systems.}
\label{tab:eom_mechanics}
\end{table}

\FloatBarrier

\begin{longtable}{cc}
\toprule
    \multicolumn{2}{c}{\bf Damped Duffing oscillator}  \\
    \multicolumn{2}{c}{\color{red} [Tiny extra non-linear damping term.]}  \\
    $\ddot x = -\gamma \dot x - \gamma_2 x^2 \dot x - k x - a x^3$&
    $\gamma = 0.046, \gamma_2 = 0.007, k = 1.661, a = 4.512$ \\
\midrule
    \multicolumn{2}{c}{\bf Damped pendulum}  \\
    $\ddot x = -\alpha\sin(x)- \gamma \dot x $&
    $\alpha=1.719, \gamma = 0.042$\\
\midrule
    \multicolumn{2}{c}{\bf Damped asymmetric double well}  \\
    \multicolumn{2}{c}{\color{red} [Tiny extra non-linear damping term.]}  \\
    $\ddot x = -\delta \dot x + \alpha x - \beta x^3 + \kappa x^5 + \gamma x^2 \dot x + F_0$&
    $\delta=0.040, \alpha=1.606, \beta = 4.510,$ \\ & $ \kappa = 0.002, \gamma = -0.003, F_0 = 0.104$\\
\midrule
    \multicolumn{2}{c}{\bf Velocity position coupling}  \\
    $\ddot x = -a x^2 \dot x - k x$&
    $a=1.702, k = 0.400, $\\
\midrule
    \multicolumn{2}{c}{\bf Arbitrary 1d potential}  \\
    $\ddot x = -k\, x + B \cos(\omega x) + c$&
    $k=1, B = -4.819, c = 0.001, \omega=6$\\
\midrule
    \multicolumn{2}{c}{\bf Damped driven oscillator}  \\
    $\ddot x = -\gamma \dot x - k x + F_c \cos(\omega t)$&
    $k=2.320, \gamma = 0.601, F_c=1.713$, $\omega=1.551$\\
\midrule
    \multicolumn{2}{c}{\bf Damped parametric oscillator}  \\
    \multicolumn{2}{c}{\color{red} [Missed external driving.]}  \\
    $\ddot x = -c \dot x - k x + b x^3$&
    $k=2.242, b=0.004$, $c=0.154$\\
\midrule
    \multicolumn{2}{c}{\bf Damped double pendulum} \\
        $\Delta  = \theta_1 -\theta_2 $ & $m_1 = m_2 = 1$ \\
        $\ddot \theta_1 = \frac{-g(2 m_1 + m_2)\sin(\theta_1) - m_2 g \sin(\theta_1 - 2 \theta_2) - 2 m_2 \sin(\Delta)\left(\dot \theta_2^2 l_2 + \dot \theta_1^2 l_1 \cos(\Delta)\right)}{l_1\left(2 m_1 + m_2 - m_2  \cos(2 \Delta)\right)}-d_1 \dot \theta_1$ & $ l_1 = 3.233, l_2 = 1.597$ \\
        $\ddot \theta_2 = \frac{2 \sin(\Delta) \left(\dot \theta_1^2 l_1 (m_1 + m_2) + g (m_1 + m_2) \cos(\theta_1) + \dot \theta_2^2 l_2 m_2 \cos(\Delta)\right)}{l_2\left(2 m_1 + m_2 - m_2  \cos(2 \Delta)\right)} -d_2 \dot \theta_2$& $g = 1.911, d_1 = 0.145, d_2 = 0.134$\\
\midrule
    \multicolumn{2}{c}{\bf Two damped coupled oscillators} \\
    $\vec{\ddot x} = A \vec{x} + B \vec{\dot x}$ &
    $A=\begin{bmatrix}-1.863 & 0.150 \\ 0.150 & -1.001\end{bmatrix}$ ,
    $B=\begin{bmatrix}-0.043 & 0 \\ 0 & -0.043\end{bmatrix}$ \\ \\
\midrule
    \multicolumn{2}{c}{\bf Three damped coupled oscillators} \\
    \multicolumn{2}{c}{Analogous to above with : $A=\begin{bmatrix}-1.500 & 0.200 & 0.300\\ 0.200 & -2.101 & 0.400\\ 0.300 & 0.400 & -1.200\end{bmatrix}$, 
$B=\begin{bmatrix}-0.162 & 0.000 & 0.000\\ 0.000 & -0.162 & 0.000\\ 0.000 & 0.000 & -0.162\end{bmatrix}$} \\
\midrule
    \multicolumn{2}{c}{\bf Damped Mexican hat potential}\\
    $\ddot x = \alpha x + \beta \left(x^2+y^2\right) x + \gamma \dot x$&
    $\alpha = 2, \beta = -0.7$ \\
    $\ddot y = \alpha y + \beta \left(x^2+y^2\right) y + \gamma \dot y$&
    $\gamma = -0.2$ \\
\midrule
    \multicolumn{2}{c}{\bf Particle in off-center gravity} \\
    $r = \sqrt{(x-c_x)^2 + (y-c_y)^2 +\epsilon^2 }$&
    $c_x = -0.700, c_y = 0.200$ \\
    $\ddot x = - \frac{G (x-c_x)}{r^{3}}$, $\ddot y = - \frac{G(y-c_y)}{r^{3}}$ & 
    $G = 2.297$, $\epsilon=0.000$ \\
\midrule
    \multicolumn{2}{c}{\bf Arbitrary 2d potential}\\
    \multicolumn{2}{c}{\color{red} [Missed analytic description of radial force. However, the numerical fit closely follows the true law.]}\\
    $r =\sqrt{(x^2 + y^2)}$&
    $g(r)$: radial acceleration from lookup table \\
    $\ddot x = -\frac{g(r)}{r}\, x$,  $\ddot y = -\frac{g(r)}{r}\, y$ \\
\midrule
    \multicolumn{2}{c}{\bf Two particles with gravity} \\
    $\Delta_x = x_1 - x_2, \Delta_y = y_1-y_2$, $r = \sqrt{\Delta_x^2 + \Delta_y^2 + \epsilon^2}$ & $m_1=8.338$, $m_2 = 0.794$ \\
    $\ddot x_1 = - \frac{m_2 \Delta_x}{r^{3}}$, $\ddot y_1 = - \frac{m_2 \Delta_y}{r^{3}}$,  $\ddot x_2 =  \frac{m_1 \Delta_x}{r^{3}}$,
    $\ddot y_2 =  \frac{m_1 \Delta_y}{r^{3}}$ & $\epsilon=0.031$ \\
\\
\midrule
    \multicolumn{2}{c}{\bf Three particles with gravity} \\
    \multicolumn{2}{c}{Analogous to two particles in gravity, $m_1=1.298, m_2=9.003, m_3=0.200, \epsilon=0.010$}\\
\midrule
    \multicolumn{2}{c}{\bf Ten particles with exponential potential}\\
    \multicolumn{2}{c}{\color{red} [Understood that it is a pairwise force but missed the exact form.]}\\
    $r_{ij}^2 = (x_j - x_i)^2+(y_j - y_i)^2$& $G=0.247$\\ 
    $\ddot{x_i} = G \sum_{j \ne i} \frac{x_j - x_i}{\left(r_{ij}^2 + s^2\right)^{(p+1)/2}}$, $ \ddot{y_i} = G \sum_{j \ne i} \frac{y_j - y_i}{\left(r_{ij}^2 + s^2\right)^{(p+1)/2}}$ & $s = 0.290, p = 2.160$\\
\midrule
   \multicolumn{2}{c} {\bf Two partially observed oscillators} \\
    \multicolumn{2}{c}{Analogous to two coupled oscillators, $k_1=1.1, k_2=1.54, k_{12}=1.2, \gamma =0$}\\
    \multicolumn{2}{c}{Hidden initial conditions, $x_2 = 0.4, \dot x_2 = -0.2$}\\
\midrule
    \multicolumn{2}{c}{\bf Three partially observed oscillators} \\
    \multicolumn{2}{c}{\color{red} [Exact analytical solution with coefficients fitted from data instead of simulating hidden particle.]}\\
    \multicolumn{2}{c}{$x(t) = \sum_{k=0}^2 \left[ \left(B^{(c)}_k + A^{(c)}_k x_0 + V^{(c)}_k \dot q_0\right) \cos(\omega_k t) + \left(B^{(s)}_k + A^{(s)}_k x_0 + V^{(s)}_k \dot q_0\right) \sin(\omega_k t) \right] + C_0$}\\
\midrule
    \multicolumn{2}{c}{\bf Two partially observed particles with gravity} \\
    \multicolumn{2}{c}{\color{red} [Completely wrong model. Not an ode.]}\\
    \multicolumn{2}{c}{Specular reflection model on five infinite straight lines.}\\
    \multicolumn{2}{c}{Dynamics: free motion (constant velocity) except when crossing a line y = m x + b,}\\
    \multicolumn{2}{c}{where velocity reflects across the line's normal.}\\
\bottomrule
\\
\caption{Best fitting equations of motion for the mechanical systems extracted from the \texttt{predict\_experiment} function saved by the agent. We include comments on the system for which the agent failed to recover the true model in red.}
\label{tab:predicted_eom_mechanics}
\end{longtable}

\FloatBarrier
\subsection{Field systems}\label{sup:details_waves}
\begin{figure*}[t]
    \centering
    \includegraphics[width=0.98\linewidth]{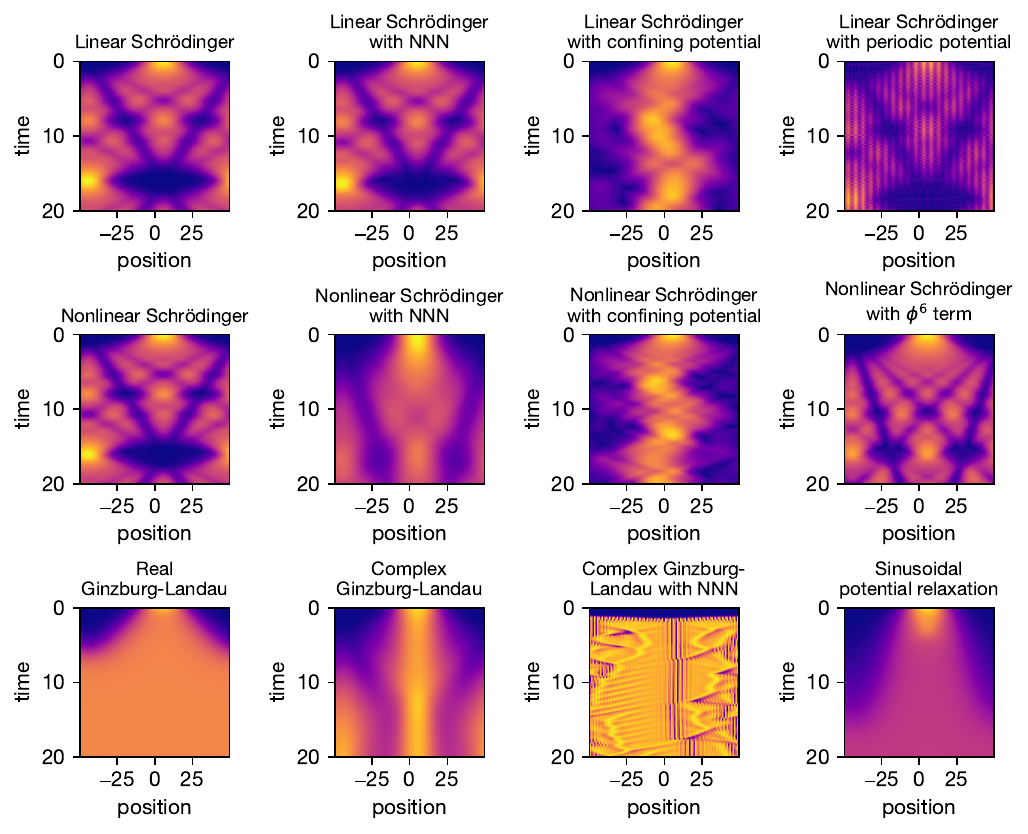}
    \caption{\textbf{Field dynamics.} Dynamics of the different field systems with a Gaussian wave packet as the initial condition. The squared absolute value of the field $|\phi|^2$ is depicted with the color scale ranging from blue (zero) to yellow (high).}\label{fig:wave_dynamics}
\end{figure*}

The agent is tasked to reproduce the time evolution of a complex-valued field on a 1d lattice. Additionally, it is given the number of lattice points, the maximum experiment time, and the space resolution of the system. We show the dynamics of the considered fields in \Cref{fig:wave_dynamics}, the exact equations of motion in \Cref{tab:eom_waves}, and the solutions discovered by the agent in \Cref{tab:agent_eom_waves}.

The task of the agent is exactly the same as for the mechanical systems above:
\begin{tcolorbox}[enhanced, breakable, title=Task supplied to agent for field tasks]
Can you find a model that reproduces the \texttt{observe\_experiment} function? After your exploration, save it using the \texttt{save\_result} function.
\end{tcolorbox}
The description of the system is focused purely on the topology of the system:
\begin{tcolorbox}[enhanced, breakable, title=Default description supplied to agent in field tasks]
This physical system consists of a complex-valued field evolving on 100 discrete coupled nodes.
The nodes are arranged in 1D with periodic boundary conditions, i.e.\ the first and last node are neighbors.
\end{tcolorbox}

\begin{tcolorbox}[enhanced, breakable, title=Additional description in case of measurement noise]
The observations you are getting may be affected by measurement noise.
\end{tcolorbox}

The agent can run an experiment via the following tool:
\begin{tcolorbox}[enhanced, breakable, title=
Experiment: observe the evolution of a field with given initial conditions]
\begin{lstlisting}[language=Python]
def observe_experiment(initial_condition_code:str, t0:float=0., T:float=100., nt:float=1001):
    Run one experiment of the evolution of the system, where you can choose the initial condition.
    Args:
        initial_condition_code: python code with jax.numpy syntax that sets the variable
        phi0, which represents the complex field at time t0. Use np.sin(...) etc.
        phi0 needs to be of shape (n_nodes,) and can be complex-valued.
        t0: float, initial time of the experiment (default 0.0)
        T: float, end time of the experiment (default 20.0, maximum 100.0)
        nt: int, number of time steps where to evaluate the experiment (default 2001,
        maximum 10001)
    Returns a dict of the form:
        ts: jax.Array of shape (nt,), the time points
        phis: jax.Array of shape (nt, n_nodes), the complex field solution
    
\end{lstlisting}
\end{tcolorbox}

\begin{table}[ht]
\centering
\begin{tabular}{cc}
\toprule
    \bf Linear Schrödinger & 
    $i\dfrac{d\phi_n}{dt} = -\tfrac{1}{2}(\Delta_d\phi)_n$ \\
    ${U_{\text{kin}}(\phi_k,k,t,\Delta t)}=\exp\!\big(-i[1-\cos(k)]\Delta t\big)\,\phi_k$ &  \\
    ${U_{\text{pot}}(\phi,x,t,\Delta t)}=\phi$ & \\
\midrule
    \bf Linear Schrödinger with NNN & 
    $i\dfrac{d\phi_n}{dt}
    = (A-1+B)\phi_n
    -\tfrac{1}{2}(\Delta_d\phi)_n
    +\tfrac{B}{2}(\Delta_{d,2}\phi)_n$ \\
    ${U_{\text{kin}}(\phi_k,k,t,\Delta t)}=\exp\!\big(-i[A-\cos(k)+B\cos(2k)]\Delta t\big)\,\phi_k$ & $A=0.5,\ B=0.5$ \\
    ${U_{\text{pot}}(\phi,x,t,\Delta t)}=\phi$ & \\
\midrule
    \bf Linear Schrödinger with confining potential & 
    $i\dfrac{d\phi_n}{dt}
    = -\tfrac{1}{2}(\Delta_d\phi)_n
    + B\cos\!\big(\tfrac{\pi (n-x_{\max})}{x_{\max}}\big)\phi_n$ \\
    ${U_{\text{kin}}(\phi_k,k,t,\Delta t)}=\exp\!\big(-i[1-\cos(k)]\Delta t\big)\,\phi_k$ &  \\
    ${U_{\text{pot}}(\phi,x,t,\Delta t)}=\exp\!\big(-iB\cos(\pi x/x_\mathrm{max})\Delta t\big)\phi$ & $B=-0.5,\ x_\mathrm{max}=49.5$ \\
\midrule
    \bf Linear Schrödinger with periodic potential & 
    $i\dfrac{d\phi_n}{dt}
    = -\tfrac{1}{2}(\Delta_d\phi)_n
    + B\cos\!\big(\tfrac{2\pi N (n-x_{\max})}{x_{\max}}\big)\phi_n$ \\
    ${U_{\text{kin}}(\phi_k,k,t,\Delta t)}=\exp\!\big(-i[1-\cos(k)]\Delta t\big)\,\phi_k$ & \\
    ${U_{\text{pot}}(\phi,x,t,\Delta t)}=\exp\!\big(-iB\cos(N\,2\pi x/x_\mathrm{max})\Delta t\big)\phi$ & $B=0.2,\ N=10,\ x_\mathrm{max}=49.5$ \\
\midrule
    \bf Nonlinear Schrödinger & 
    $i\dfrac{d\phi_n}{dt}
    = -\tfrac{A}{2}(\Delta_d\phi)_n
    + B|\phi_n|^2\phi_n$ \\
    ${U_{\text{kin}}(\phi_k,k,t,\Delta t)}=\exp\!\big(-iA[1-\cos(k)]\Delta t\big)\,\phi_k$ & $A=0.5$ \\
    ${U_{\text{pot}}(\phi,x,t,\Delta t)}=\exp\!\big(-iB|\phi|^2\Delta t\big)\phi$ & $B=0.25$ \\
\midrule
    \bf Nonlinear Schrödinger with NNN & 
    $\begin{aligned}i\dfrac{d\phi_n}{dt}
    =&(AB-1+AC)\phi_n
    -\tfrac{A}{2}(\Delta_d\phi)_n \\
    &+\tfrac{AC}{2}(\Delta_{d,2}\phi)_n
    + D|\phi_n|^2\phi_n\end{aligned}$\\
    ${U_{\text{kin}}(\phi_k,k,t,\Delta t)}=\exp\!\big(-i[A(B-\cos(k)+C\cos(2k))]\Delta t\big)\,\phi_k$ & $A=0.5,\ B=1.5,\ C=0.5$ \\
    ${U_{\text{pot}}(\phi,x,t,\Delta t)}=\exp\!\big(-iD|\phi|^2\Delta t\big)\phi$ & $D=0.25$ \\
\midrule
    \bf Nonlinear Schrödinger with confining potential &
    $i\dfrac{d\phi_n}{dt}
    = -\tfrac{1}{2}(\Delta_d\phi)_n
    + \Big(-\cos\!\big(\tfrac{\pi (n-x_{\max})}{x_{\max}}\big)+|\phi_n|^2\Big)\phi_n$ \\
    ${U_{\text{kin}}(\phi_k,k,t,\Delta t)}=\exp\!\big(-i[1-\cos(k)]\Delta t\big)\,\phi_k$ & \\
    ${U_{\text{pot}}(\phi,x,t,\Delta t)}=\exp\!\big(-i[-\cos(\pi x/x_\mathrm{max})+|\phi|^2]\Delta t\big)\phi$ & $N=10,\ x_\mathrm{max}=49.5$ \\
\midrule
    \bf Nonlinear Schrödinger with $\phi^6$ term &
    $i\dfrac{d\phi_n}{dt}
    = -\tfrac{A}{2}(\Delta_d\phi)_n
    + B(|\phi_n|^2+2|\phi_n|^4)\phi_n$ \\
    ${U_{\text{kin}}(\phi_k,k,t,\Delta t)}=\exp\!\big(-iA[1-\cos(k)]\Delta t\big)\,\phi_k$ & $A=0.5$ \\
    ${U_{\text{pot}}(\phi,x,t,\Delta t)}=\exp\!\big(-iB(|\phi|^2+2|\phi|^4)\Delta t\big)\phi$ & $B=0.25$ \\
\midrule
    \bf Real Ginzburg--Landau &
    $\dfrac{d\phi_n}{dt}
    = -A(\Delta_d\phi)_n
    - B(2|\phi_n|^2-1)\phi_n$ \\
    ${U_{\text{kin}}(\phi_k,k,t,\Delta t)}=\exp\!\big(-A[1-\cos(k)]\Delta t\big)\,\phi_k$ & $A=0.5$ \\
    ${U_{\text{pot}}(\phi,x,t,\Delta t)}=\exp\!\big(-B(2|\phi|^2-1)\Delta t\big)\phi$ & $B=0.5$ \\
\midrule
    \bf Complex Ginzburg--Landau &
    $\dfrac{d\phi_n}{dt}
    = -A(\Delta_d\phi)_n
    - B(2C|\phi_n|^2-1)\phi_n$ \\
    ${U_{\text{kin}}(\phi_k,k,t,\Delta t)}=\exp\!\big(-A[1-\cos(k)]\Delta t\big)\,\phi_k$ & $A=0.2(0.5+2i)$ \\
    ${U_{\text{pot}}(\phi,x,t,\Delta t)}=\exp\!\big(-B(2C|\phi|^2-1)\Delta t\big)\phi$ & $B=0.2,\ C=(1-1.5i)$ \\
\midrule
    \bf Complex Ginzburg--Landau with NNN &
    $\begin{aligned}\dfrac{d\phi_n}{dt}
    =&-A\!\Big[(B-1+C)\phi_n
    -\tfrac{1}{2}(\Delta_d\phi)_n
    +\tfrac{C}{2}(\Delta_{d,2}\phi)_n\Big]\\
    &- D(2E|\phi_n|^2-1)\phi_n\end{aligned}$ \\
    ${U_{\text{kin}}(\phi_k,k,t,\Delta t)}=\exp\!\big(-A[B-\cos(k)+C\cos(2k)]\Delta t\big)\,\phi_k$ & $A=0.8(0.5+0.5i),\ B=0.65,\ C=0.8$ \\
    ${U_{\text{pot}}(\phi,x,t,\Delta t)}=\exp\!\big(-D(2E|\phi|^2-1)\Delta t\big)\phi$ & $D=0.2,\ E=(1-1.5i)$  \\
\midrule
    \bf Sinusoidal potential relaxation &
    $\dfrac{d\phi_n}{dt}
    = -A(\Delta_d\phi)_n
    + B\sin(C|\phi_n|^2)\phi_n$ \\
    ${U_{\text{kin}}(\phi_k,k,t,\Delta t)}=\exp\!\big(-A[1-\cos(k)]\Delta t\big)\,\phi_k$ & $A=0.5$ \\
    ${U_{\text{pot}}(\phi,x,t,\Delta t)}=\exp\!\big(B\sin(C|\phi|^2)\Delta t\big)\phi$ & $B=0.1,\ C=15$ \\
\bottomrule
\end{tabular}
\caption{Field equation and corresponding kinetic and potential propagators for field systems. 
Here $(\Delta_d\phi)_n=\phi_{n+1}-2\phi_n+\phi_{n-1}$ and $(\Delta_{d,2}\phi)_n=\phi_{n+2}-2\phi_n+\phi_{n-2}$.}
\label{tab:eom_waves}
\end{table}

\FloatBarrier
\begin{longtable}{cc}
\toprule
\multicolumn{2}{c}{\textbf{Linear Schr\"odinger}}\\
$ i\dfrac{d\phi_n}{dt}=-\kappa\,(\Delta_d\phi)_n $ & $\kappa = 50.0$ \\
\midrule
\multicolumn{2}{c}{\textbf{Linear Schr\"odinger with NNN}}\\
$ i\dfrac{d\phi_n}{dt}= \alpha\,(\Delta_d\phi)_n + \beta\,(\Delta_{d,2}\phi)_n $ & \begin{minipage}[t]{0.3\linewidth}$ \alpha = 50.0 $\\$ \beta = 25.0 $\end{minipage} \\
\midrule
\multicolumn{2}{c}{\textbf{Linear Schr\"odinger with confining potential}}\\
$ i\dfrac{d\phi_n}{dt}=-\kappa\,(\Delta_d\phi)_n - V_n\,\phi_n,\quad V_n = V_0\cos\!\big(\tfrac{2\pi n}{N}+\Phi\big) $ & \begin{minipage}[t]{0.3\linewidth}$ \kappa = 50.0 $\\$ V_0 = 0.5 $\\$ N = \text{number of lattice sites} $\\$ \Phi = -\pi + \tfrac{\pi}{N} $\end{minipage} \\
\midrule
\multicolumn{2}{c}{\textbf{Linear Schr\"odinger with periodic potential}}\\
\multicolumn{2}{c}{\color{red} [Included unnecessary couplings beyond nearest-neighbors.]}\\
\multicolumn{2}{c}{Linear Schr\"odinger equation with up to fourth nearest-neighbor coupling fitted from data and onsite potential $V_n$}\\
$V_n=A + B \cos(2\pi \frac{n}{P}) + C \sin(2\pi \frac{n}{P})$ & \begin{minipage}[t]{0.3\linewidth}$A=70.4$,\\ $B=-14.2$,\\ $C=-10.1, \\ P=5$\end{minipage}\\
\midrule
\multicolumn{2}{c}{\textbf{Nonlinear Schr\"odinger}}\\
$ i\dfrac{d\phi_n}{dt}= -\kappa\,(\Delta_d\phi)_n + g\,|\phi_n|^2\phi_n $ & \begin{minipage}[t]{0.3\linewidth}$ \kappa = 50.0 $\\$ g = 0.25 $\end{minipage} \\
\midrule
\multicolumn{2}{c}{\textbf{Nonlinear Schr\"odinger with NNN}}\\
$ i\dfrac{d\phi_n}{dt}= \omega_0\,\phi_n + \alpha\,(\Delta_d\phi)_n + \beta\,(\Delta_{d,2}\phi)_n + g\,|\phi_n|^2\phi_n $ & \begin{minipage}[t]{0.3\linewidth}$ \omega_0 = 50.0 $\\$ \alpha = -25.0 $\\$ \beta = 12.5 $\\$ g = 0.25 $\end{minipage} \\
\midrule
\multicolumn{2}{c}{\textbf{Nonlinear Schr\"odinger with confining potential}}\\
$ i\dfrac{d\phi_n}{dt}= -\kappa\,(\Delta_d\phi)_n - V_n\,\phi_n + \gamma\,|\phi_n|^2\phi_n$,\\
$ V_n = a\cos\!\big(\tfrac{2\pi n}{N}\big) + b\sin\!\big(\tfrac{2\pi n}{N}\big) $ & \begin{minipage}[t]{0.3\linewidth}$ \kappa = 50.01 $\\$ \gamma = 1.00 $\\$ a = -1.00 $\\$ b = 0.03 $\\$ N = \text{lattice size (ring)} $\end{minipage} \\
\midrule
\multicolumn{2}{c}{\textbf{Nonlinear Schr\"odinger with $\phi^6$ term}}\\
$ i\dfrac{d\phi_n}{dt}= -\kappa\,(\Delta_d\phi)_n + g\,|\phi_n|^2\phi_n + h\,|\phi_n|^4\phi_n $ & \begin{minipage}[t]{0.3\linewidth}$ \kappa = 25.0 $\\$ g = 0.25 $\\$ h = 0.5 $\end{minipage} \\
\midrule
\multicolumn{2}{c}{\textbf{Real Ginzburg-Landau}}\\
$ \dfrac{d\phi_n}{dt}= \mu\,\phi_n + D\,(\Delta_d\phi)_n - g\,|\phi_n|^2\phi_n $ & \begin{minipage}[t]{0.3\linewidth}$ \mu = 0.5 $\\$ D = 25.0 $\\$ g = 1.0 $\end{minipage} \\
\midrule
\multicolumn{2}{c}{\textbf{Complex Ginzburg-Landau}}\\
$ \dfrac{d\phi_n}{dt}= a\,\phi_n + b\,(\Delta_d\phi)_n - c\,|\phi_n|^2\phi_n $ & \begin{minipage}[t]{0.3\linewidth}$ a = 0.2 $\\$ b = 5 + 20i $\\$ c = 0.4 - 0.6i $\end{minipage} \\
\midrule
\multicolumn{2}{c}{\textbf{Complex Ginzburg-Landau with NNN}}\\
\multicolumn{2}{c}{\color{red} [Included unnecessary Laplacian-nonlinear term (prefactor E)]}\\
$ \dfrac{d\phi_n}{dt}= A\,\phi_n + B\,(\Delta_d\phi)_n + C\,(\Delta_{d,2}\phi)_n + D\,|\phi_n|^2\phi_n + E\,|\phi_n|^2(\Delta_d\phi)_n $ & 
\begin{minipage}[t]{0.3\linewidth}$ A = -15.22-17.39i $\\$ B = -42.85-40.58i $\\$ C = -15.89 - 14.63i $\\$ D = -0.49 + 0.66i $\\$ E = -0.037 + 0.012i $\end{minipage} \\
\midrule
\multicolumn{2}{c}{\textbf{Sinusoidal potential relaxation}}\\
\multicolumn{2}{c}{\color{red} [Found completely different model.]}\\
\begin{minipage}[t]{0.5\linewidth}
    $ \dfrac{d\phi_n}{dt}= a\,\phi_n + b\,(\Delta_d\phi)_n - c\,|\phi_n|^2\phi_n + \alpha\,|\phi_n|^2(\Delta_d\phi)_n + \beta\,\big(\Delta_d(|\phi|^2\phi)\big)_n  + \gamma\,(\Delta_{d,2}\phi)_n + \mu\,\langle|\phi|^2\rangle\,\phi_n $\\
    Here $\langle|\phi|^2\rangle$ is the mean value at a given time.
\end{minipage} 
& \begin{minipage}[t]{0.3\linewidth}$ a = 0.05 $\\$ b = 25.08 $\\$ c = 0.16 $\\$ \alpha = 2.88 $\\$ \beta = -3.00 $\\$ \gamma = -1.07 $\\$ \mu = 0.11 $\end{minipage} \\
\bottomrule
\caption{Best fitting equations of motion for the field systems extracted from the \texttt{predict\_experiment} function saved by the agent.
We include comments on the system for which the agent failed to recover the true model in red.}
\label{tab:agent_eom_waves}
\end{longtable}

The agent saves its result via:
\begin{tcolorbox}[enhanced, breakable, title=
Save prediction for field systems]
\begin{lstlisting}[language=Python]
def save_result(code:str):
    Saves your final result.
    You can only call this once. Do not call it when you could potentially further improve your result!
    You should provide code that defines a predict_experiment function which should reproduce
    trajectories of the experiment.
    Your code does not have access to previously computed arrays, so make sure your model does not need access to a large set of numerical constants.
    The code has access to numpy (as np), jax (as jax), jax.numpy (as jnp), and scipy (as scipy).
    Args:
        code:  String defining a function with the following signature:
            def predict_experiment(initial_cond:jax.Array, t0:float, T:float, nt:int)
                -> Tuple[jax.Array, jax.Array]:
                Args (of predict_experiment):
                    initial_cond : complex jax.Array of shape (n_nodes,), the initial 
                    condition of the complex field at time t=t0
                    t0 : float, initial time of the experiment
                    T : float, end time of the experiment
                    nt : int, number of time steps where to evaluate the experiment
                Returns a tuple of the form (ts, phis):
                    ts : jax.Array of shape (nt,) with time points [t0, t0+dt, ..., T]
                    phis : complex jax.Array of shape (nt, n_nodes), the complex field 
                    solution
\end{lstlisting}
\end{tcolorbox}

\FloatBarrier

\subsection{Quantum systems}

The agent is asked to discover the underlying Hamiltonian for an unknown quantum many-body system composed of spins (spin 1/2, i.e.\ qubits). In some cases, the Hamiltonian may contain one or several parameters (fields, couplings). In some cases, the agent can change the size $N$ of the experimental system as well. The agent does not have any information except that it is dealing with a spin 1/2 system and that the number of spins is a certain value (or variable) and the connectivity of the system (1d vs 2d grid). If there is a parameter, the agent is just being told the name of that parameter (e.g.\ `A') without reference to its meaning. The exact Hamiltonians are specified in \Cref{tab:eom_quantum} and the Hamiltonians predicted by the agent in \Cref{tab:predicted_ham_quantum}.

The system is described in the following way:

\begin{tcolorbox}[enhanced, breakable, title=Default description supplied to agent in quantum tasks]
You do have access to an experimental system of ... spins.
\end{tcolorbox}

\begin{tcolorbox}[enhanced, breakable, title=Additional description in case of 2d systems]
The system is a 2D system of spins on a 3x3 grid, with spin index j=jx+jy*M, for M=3.
\end{tcolorbox}

\begin{tcolorbox}[enhanced, breakable, title=Additional description in case of variable system size]
You can change the number of spins N in each experiment call. You should not choose values larger than about N=12.
\end{tcolorbox}

\begin{tcolorbox}[enhanced, breakable, title=Additional description for partially observed dynamics]
You can only observe and initialize the spins at the indices ..., whose spin expectation values will be returned in this order.
\end{tcolorbox}

\begin{tcolorbox}[enhanced, breakable, title=Additional description in case of shot noise]
Expectation values are estimated from ... measurement shots, leading to some statistical noise.
\end{tcolorbox}

The task for the agent is:

\begin{tcolorbox}[enhanced, breakable, title=Task supplied to agent for field tasks]
Can you find the Hamiltonian of this system? After your exploration, save it using the \texttt{save\_result} function.
\end{tcolorbox}

To save its prediction, the agent uses:
\begin{tcolorbox}[enhanced, breakable, title=
Save result for quantum systems]
\begin{lstlisting}[language=Python]
def save_result(code:str)
    Saves your final result.
    You can only call this once. Do not call it when you could potentially further improve your result!
    You pass a python code that must produce an operator H, constructing it out of provided spin operators. Here Sx is a list of spin operator x-components, Sy likewise for the y-components, and Sz for the z-components. These are Pauli matrices. They can be accessed like Sx[2] etc.  Remember to use the "@" matrix multiplication operator when taking the product of several spin operators. Otherwise you can use jax.numpy syntax in the form "jnp.sin(...)".
    You may also use the current variable N (number of spins in the experiment), but you cannot set a new value for it here.
    Args:
        code: the python code defining the Hamiltonian.
    Returns:
        Message that the Hamiltonian has been stored.
\end{lstlisting}
\end{tcolorbox}

\begin{table}[t]
\centering
\begin{tabular}{cc}
\toprule
    \multicolumn{2}{c}{\bf GS Transverse-field Ising chain}\\
    ${\hat H}=J\sum_{j=0}^{N-2}{\hat \sigma}^{z}_j{\hat \sigma}^{z}_{j+1}-h\sum_{j=0}^{N-1}{\hat \sigma}^{x}_j$ &
    $N=10,\ J=1.5,\ h=0.6$ \\
\midrule
    \multicolumn{2}{c}{\bf GS Transverse-field Ising chain with tunable coupling} \\
    ${\hat H}=A\sum_{j=0}^{N-2}{\hat \sigma}^{z}_j{\hat \sigma}^{z}_{j+1}-h\sum_{j=0}^{N-1}{\hat \sigma}^{x}_j$ &
    $N=10,\ h=0.6,\ A\ \text{tunable}$ \\
\midrule
    \multicolumn{2}{c}{\bf GS Transverse-field Ising chain with tunable coupling and number of spins} \\
    ${\hat H}=A\sum_{j=0}^{N-2}{\hat \sigma}^{z}_j{\hat \sigma}^{z}_{j+1}-h\sum_{j=0}^{N-1}{\hat \sigma}^{x}_j$ &
    $N\ \text{tunable},\ h=0.6,\ A\ \text{tunable}, $ \\
\midrule
    \multicolumn{2}{c}{\bf GS 2d Heisenberg model} \\
    ${\hat H}=J\sum_{\langle r,r'\rangle}\big({\hat \sigma}^{x}_{r}{\hat \sigma}^{x}_{r'}+{\hat \sigma}^{y}_{r}{\hat \sigma}^{y}_{r'}+{\hat \sigma}^{z}_{r}{\hat \sigma}^{z}_{r'}\big)-h\sum_{r}{\hat \sigma}^{x}_{r}$ &
    $N=9,\ J=1,\ h=2$ \\
\midrule
    \multicolumn{2}{c}{\bf GS 2d Heisenberg model with tunable coupling} \\
    ${\hat H}=2A\sum_{\langle r,r'\rangle}\big({\hat \sigma}^{x}_{r}{\hat \sigma}^{x}_{r'}+{\hat \sigma}^{y}_{r}{\hat \sigma}^{y}_{r'}+{\hat \sigma}^{z}_{r}{\hat \sigma}^{z}_{r'}\big)-h\sum_{r}{\hat \sigma}^{x}_{r}$ &
    $N=9,\ h=2,\ A\ \text{tunable}$ \\
\midrule
    \multicolumn{2}{c}{\bf GS 2d Heisenberg model with tunable coupling and field strength} \\
    ${\hat H}=A\sum_{\langle r,r'\rangle}\big({\hat \sigma}^{x}_{r}{\hat \sigma}^{x}_{r'}+{\hat \sigma}^{y}_{r}{\hat \sigma}^{y}_{r'}+{\hat \sigma}^{z}_{r}{\hat \sigma}^{z}_{r'}\big)-2B\sum_{r}{\hat \sigma}^{x}_{r}$ &
    $N=9,\ A\ \text{tunable},\ B\ \text{tunable}$ \\
\midrule
    \multicolumn{2}{c}{\bf GS Cluster Ising chain} \\
    ${\hat H}=K\sum_{j=0}^{N-3}{\hat \sigma}^{z}_j{\hat \sigma}^{x}_{j+1}{\hat \sigma}^{z}_{j+2}-J\sum_{j=0}^{N-2}{\hat \sigma}^{z}_j{\hat \sigma}^{z}_{j+1}-h\sum_{j=0}^{N-1}{\hat \sigma}^{x}_j$ &
    $N=10,\ K=0.5,\ J=1,\ h=0.3$ \\
\midrule
    \multicolumn{2}{c}{\bf GS Cluster Ising chain with tunable coupling} \\
    ${\hat H}=A\sum_{j=0}^{N-3}{\hat \sigma}^{z}_j{\hat \sigma}^{x}_{j+1}{\hat \sigma}^{z}_{j+2}-J\sum_{j=0}^{N-2}{\hat \sigma}^{z}_j{\hat \sigma}^{z}_{j+1}-h\sum_{j=0}^{N-1}{\hat \sigma}^{x}_j$ &
    $N=10,\ A\ \text{tunable},\ J=1,\ h=1$ \\
\midrule
    \multicolumn{2}{c}{\bf GS Topological Ising chain with tunable coupling and number of spins} \\
    ${\hat H}=A\sum_{j=0}^{N-3}{\hat \sigma}^{z}_j{\hat \sigma}^{x}_{j+1}{\hat \sigma}^{z}_{j+2}-J\sum_{j=0}^{N-2}{\hat \sigma}^{z}_j{\hat \sigma}^{z}_{j+1}-h\sum_{j=0}^{N-1}{\hat \sigma}^{x}_j$ &
    $N\ \text{tunable},\ A\ \text{tunable},\  J=1,\ h=1$ \\
\midrule
    \multicolumn{2}{c}{\bf GS Arbitrary Hamiltonian} \\
    ${\hat H}=J_{xz}\sum_{j=0}^{N-2}{\hat \sigma}^{x}_j{\hat \sigma}^{z}_{j+1}-J_{yx}\sum_{j=0}^{N-2}{\hat \sigma}^{y}_j{\hat \sigma}^{x}_{j+1}-h_x\sum_{j=0}^{N-1}{\hat \sigma}^{x}_j+h_y\sum_{j=0}^{N-1}{\hat \sigma}^{y}_j$ &
    $N=10,\ J_{xz}=1.5,\ J_{yx}=0.7,\ $ \\
    & $h_x=0.6,\ h_y=0.4$ \\
\midrule
    \multicolumn{2}{c}{\bf DYN Arbitrary Hamiltonian} \\
    ${\hat H}=J_{1}{\hat \sigma}^{x}_{0}{\hat \sigma}^{z}_{1}+J_{2}{\hat \sigma}^{y}_{0}{\hat \sigma}^{x}_{2}-h_{1}{\hat \sigma}^{y}_{1}+h_{2}{\hat \sigma}^{y}_{2}-K{\hat \sigma}^{x}_{1}{\hat \sigma}^{y}_{2}$ &
    $N=3,\ J_{1}=1,\ J_{2}=0.5,\  $ \\
    & $h_{1}=0.7,\ h_{2}=0.3,\ K=0.8$ \\
\midrule
    \multicolumn{2}{c}{\bf DYN Arbitrary Hamiltonian with access to two spins} \\
    ${\hat H}=J_{1}{\hat \sigma}^{x}_{0}{\hat \sigma}^{z}_{1}+J_{2}{\hat \sigma}^{y}_{0}{\hat \sigma}^{x}_{2}-h_{1}{\hat \sigma}^{y}_{1}+h_{2}{\hat \sigma}^{y}_{2}-K{\hat \sigma}^{x}_{1}{\hat \sigma}^{y}_{2}$ &
    $N=3,\ J_{1}=1,\ J_{2}=0.5,\ $ \\
    & $h_{1}=0.7,\ h_{2}=0.3,\ K=0.8$ \\
\midrule
    \multicolumn{2}{c}{\bf DYN Arbitrary Hamiltonian with access to two spins with tunable field} \\
    ${\hat H}=J_{1}{\hat \sigma}^{x}_{0}{\hat \sigma}^{z}_{1}+J_{2}{\hat \sigma}^{y}_{0}{\hat \sigma}^{x}_{2}-A{\hat \sigma}^{y}_{1}+h_{2}{\hat \sigma}^{y}_{2}-K{\hat \sigma}^{x}_{1}{\hat \sigma}^{y}_{2}$ &
    $N=3,\ J_{1}=1,\ J_{2}=0.5,\ $ \\
    & $h_{2}=0.3,\ K=0.8,\ A\ \text{tunable}$ \\
\midrule
    \multicolumn{2}{c}{\bf DYN Heisenberg chain} \\
    ${\hat H}=J\sum_{j=0}^{N-2}\!\left({\hat \sigma}^{x}_{j}{\hat \sigma}^{x}_{j+1}+{\hat \sigma}^{y}_{j}{\hat \sigma}^{y}_{j+1}+{\hat \sigma}^{z}_{j}{\hat \sigma}^{z}_{j+1}\right)-h\sum_{j=0}^{N-1}{\hat \sigma}^{x}_{j}$ &
    $N=10,\ J=0.5,\ h=1.5$ \\
\midrule
    \multicolumn{2}{c}{\bf DYN Heisenberg chain with tunable field} \\
    ${\hat H}=J\sum_{j=0}^{N-2}\!\left({\hat \sigma}^{x}_{j}{\hat \sigma}^{x}_{j+1}+{\hat \sigma}^{y}_{j}{\hat \sigma}^{y}_{j+1}+{\hat \sigma}^{z}_{j}{\hat \sigma}^{z}_{j+1}\right)-A\sum_{j=0}^{N-1}{\hat \sigma}^{x}_{j}$ &
    $N=10,\ J=0.5,\  A\ \text{tunable}$ \\
\midrule
    \multicolumn{2}{c}{\bf DYN Heisenberg chain with access to two spins with tunable field} \\
    ${\hat H}=J\sum_{j=0}^{N-2}\!\left({\hat \sigma}^{x}_{j}{\hat \sigma}^{x}_{j+1}+{\hat \sigma}^{y}_{j}{\hat \sigma}^{y}_{j+1}+{\hat \sigma}^{z}_{j}{\hat \sigma}^{z}_{j+1}\right)-A\sum_{j=0}^{N-1}{\hat \sigma}^{x}_{j}$ &
    $N=10,\ J=0.5,\  A\ \text{tunable}$ \\
\midrule
    \multicolumn{2}{c}{\bf DYN Transverse-field Ising chain with tunable coupling} \\
    ${\hat H}=A\sum_{j=0}^{N-2}{\hat \sigma}^{z}_{j}{\hat \sigma}^{z}_{j+1}-h\sum_{j=0}^{N-1}{\hat \sigma}^{x}_{j}$ &
    $N=10,\ A\ \text{tunable},\ h=1$ \\
\midrule
    \multicolumn{2}{c}{\bf DYN Transverse-field Ising chain with access to two spins with tunable coupling} \\
    ${\hat H}=A\sum_{j=0}^{N-2}{\hat \sigma}^{z}_{j}{\hat \sigma}^{z}_{j+1}-h\sum_{j=0}^{N-1}{\hat \sigma}^{x}_{j}$ &
    $N=10,\ A\ \text{tunable},\ h=1$ \\
\bottomrule
\end{tabular}
\caption{Hamiltonians for the considered quantum ground state (GS) and dynamics tasks (DYN).}
\label{tab:eom_quantum}
\end{table}

\begin{table}[t]
\centering
\begin{tabular}{cc}
\toprule
    \multicolumn{2}{c}{\bf GS Transverse-field Ising chain}\\
    ${\hat H}=\sum_{j=0}^{N-2}{\hat \sigma}^{z}_j{\hat \sigma}^{z}_{j+1}-h\sum_{j=0}^{N-1}{\hat \sigma}^{x}_j$ &
    $N =10 ,\ h=0.4$ \\
\midrule
    \multicolumn{2}{c}{\bf GS Transverse-field Ising chain with tunable coupling} \\
    ${\hat H}=A\sum_{j=0}^{N-2}{\hat \sigma}^{z}_j{\hat \sigma}^{z}_{j+1}-0.6\sum_{j=0}^{N-1}{\hat \sigma}^{x}_j$ &
    $N\ \text{tunable},\ A\ \text{tunable},\ h=0.6$ \\
\midrule
    \multicolumn{2}{c}{\bf GS Transverse-field Ising chain with tunable coupling and number of spins} \\
    ${\hat H}=A\sum_{j=0}^{N-2}{\hat \sigma}^{z}_j{\hat \sigma}^{z}_{j+1}-g\sum_{j=0}^{N-1}{\hat \sigma}^{x}_j$ &
    $N = 10,\ g=0.6,\ A\ \text{tunable}, $ \\
\midrule
    \multicolumn{2}{c}{\bf GS 2d Heisenberg model} \\
    ${\hat H}=J\sum_{\langle r,r'\rangle}\big({\hat \sigma}^{x}_{r}{\hat \sigma}^{x}_{r'}+{\hat \sigma}^{y}_{r}{\hat \sigma}^{y}_{r'}+{\hat \sigma}^{z}_{r}{\hat \sigma}^{z}_{r'}\big)-\sum_{r}{\hat \sigma}^{x}_{r}$ &
    $N=9,\ J=0.5$ \\
\midrule
    \multicolumn{2}{c}{\bf GS 2d Heisenberg model with tunable coupling} \\
    \multicolumn{2}{c}{\color{red}[Missed factor of two for $\sigma_x$ terms. Wrong boundary conditions.]}\\
    ${\hat H}=A\sum_{\langle r,r'\rangle}^{\mathrm{PBC}}\big({\hat \sigma}^{x}_{r}{\hat \sigma}^{x}_{r'}+{\hat \sigma}^{y}_{r}{\hat \sigma}^{y}_{r'}+{\hat \sigma}^{z}_{r}{\hat \sigma}^{z}_{r'}\big)-\sum_{r}{\hat \sigma}^{x}_{r}$ &
    $N=9,\ A\ \text{tunable}$ \\
\midrule
    \multicolumn{2}{c}{\bf GS 2d Heisenberg model with tunable coupling and field strength} \\
    \multicolumn{2}{c}{\color{red}[Missed factor of two for $\sigma_x$ terms. Wrong boundary conditions.]}\\
    ${\hat H}=A\sum_{\langle r,r'\rangle}^{\mathrm{PBC}}\big({\hat \sigma}^{x}_{r}{\hat \sigma}^{x}_{r'}+{\hat \sigma}^{y}_{r}{\hat \sigma}^{y}_{r'}+{\hat \sigma}^{z}_{r}{\hat \sigma}^{z}_{r'}\big)-B\sum_{r}{\hat \sigma}^{x}_{r}$ &
    $N=9,\ A\ \text{tunable},\ B\ \text{tunable}$ \\
\midrule
    \multicolumn{2}{c}{\bf GS Cluster Ising chain} \\
    ${\hat H}=K\sum_{j=0}^{N-3}{\hat \sigma}^{z}_j{\hat \sigma}^{x}_{j+1}{\hat \sigma}^{z}_{j+2}-J\sum_{j=0}^{N-2}{\hat \sigma}^{z}_j{\hat \sigma}^{z}_{j+1}-h\sum_{j=0}^{N-1}{\hat \sigma}^{x}_j$ &
    $N=10,\ K=0.5,\ J=1,\ h=0.3$ \\
\midrule
    \multicolumn{2}{c}{\bf GS Cluster Ising chain with tunable coupling} \\
    ${\hat H}=-\sum_{j=0}^{N-1}{\hat \sigma}^{x}_{j}-\sum_{j=0}^{N-2}{\hat \sigma}^{z}_{j}{\hat \sigma}^{z}_{j+1}+A\sum_{j=0}^{N-3}{\hat \sigma}^{z}_{j}{\hat \sigma}^{x}_{j+1}{\hat \sigma}^{z}_{j+2}$ &
    $N=10,\ A\ \text{tunable},\ $ \\
\midrule
    \multicolumn{2}{c}{\bf GS Topological Ising chain with tunable coupling and number of spins} \\
    ${\hat H}=A\sum_{j=0}^{N-3}{\hat \sigma}^{z}_j{\hat \sigma}^{x}_{j+1}{\hat \sigma}^{z}_{j+2}-\sum_{j=0}^{N-2}{\hat \sigma}^{z}_j{\hat \sigma}^{z}_{j+1}-\sum_{j=0}^{N-1}{\hat \sigma}^{x}_j$ &
    $N\ \text{tunable},\ A\ \text{tunable},\  $ \\
\midrule
    \multicolumn{2}{c}{\bf GS Arbitrary Hamiltonian} \\
    ${\hat H}=\sum_{j=0}^{N-1}\!\left(h_x\,{\hat \sigma}^{x}_{j}+h_y\,{\hat \sigma}^{y}_{j}\right)+\sum_{j=0}^{N-2}\!\Big(J_{yx}\,{\hat \sigma}^{y}_{j}{\hat \sigma}^{x}_{j+1}+J_{xz}\,{\hat \sigma}^{x}_{j}{\hat \sigma}^{z}_{j+1}\Big)$ &
    $N=10,\, J_{yx}=-0.39,\ J_{xz}=0.83,$ \\
    & $h_x=-0.33,\ h_y=0.22$ \\
\midrule
    \multicolumn{2}{c}{\bf DYN Arbitrary Hamiltonian} \\
    ${\hat H}=J_{1}{\hat \sigma}^{x}_{0}{\hat \sigma}^{z}_{1}+J_{2}{\hat \sigma}^{y}_{0}{\hat \sigma}^{x}_{2}-h_{1}{\hat \sigma}^{y}_{1}+h_{2}{\hat \sigma}^{y}_{2}-K{\hat \sigma}^{x}_{1}{\hat \sigma}^{y}_{2}$ &
    $N=3,\ J_{1}=1.0,\ J_{2}=0.5,\  $ \\
    & $h_{1}=0.7,\ h_{2}=0.3,\ K=0.8$ \\
\midrule
    \multicolumn{2}{c}{\bf DYN Arbitrary Hamiltonian with access to two spins} \\
    \multicolumn{2}{c}{\color{red} [Overly complicated model. Wrong constants in fit.]} \\
    ${\hat H}=\sum_{a,b\in\{x,y,z\}}\Big(J_{01}^{ab}\,\sigma^{a}_{0}\sigma^{b}_{1}+J_{12}^{ab}\,\sigma^{a}_{1}\sigma^{b}_{2}+J_{02}^{ab}\,\sigma^{a}_{0}\sigma^{b}_{2}\Big)$ &
    $N=3,\ J_{01}\in\mathbb{R}^{3\times 3}\ (\text{fitted}),\ J_{12}\in\mathbb{R}^{3\times 3}\ (\text{fitted}),\ $\\
    $+\sum_{i=0}^{2}\big(h_{i}^{x}\,\sigma^{x}_{i}+h_{i}^{y}\,\sigma^{y}_{i}+h_{i}^{z}\,\sigma^{z}_{i}\big)$ 
    & $J_{02}\in\mathbb{R}^{3\times 3}\ (\text{fitted}),\ h \in\mathbb{R}^{3\times 3}\ (\text{fitted})$ \\
\midrule
    \multicolumn{2}{c}{\bf DYN Arbitrary Hamiltonian with access to two spins with tunable field} \\
    \multicolumn{2}{c}{\color{red} [Wrong model.]} \\
    ${\hat H}=J_{xz}^{01}{\hat \sigma}^{x}_{0}{\hat \sigma}^{z}_{1}+J_{xy}^{12}\left({\hat \sigma}^{x}_{1}{\hat \sigma}^{x}_{2}+{\hat \sigma}^{y}_{1}{\hat \sigma}^{y}_{2}\right)+B_{2y}{\hat \sigma}^{y}_{2}-A{\hat \sigma}^{y}_{1}$ &
    $N=3,\ J_{xz}^{01}=0.85,\ J_{xy}^{12}=0.38,\ $ \\
    & $B_{2y}=0.52,\ A\ \text{tunable}$ \\
\midrule
    \multicolumn{2}{c}{\bf DYN Heisenberg chain} \\
    ${\hat H}=J\sum_{j=0}^{N-2}\!\left({\hat \sigma}^{x}_{j}{\hat \sigma}^{x}_{j+1}+{\hat \sigma}^{y}_{j}{\hat \sigma}^{y}_{j+1}+{\hat \sigma}^{z}_{j}{\hat \sigma}^{z}_{j+1}\right)+h_x\sum_{j=0}^{N-1}{\hat \sigma}^{x}_{j}$ &
    $N=10,\ J=0.5,\ h_x=-1.5$ \\
\midrule
    \multicolumn{2}{c}{\bf DYN Heisenberg chain with tunable field} \\
    ${\hat H}=J\sum_{j=0}^{N-2}\!\left({\hat \sigma}^{x}_{j}{\hat \sigma}^{x}_{j+1}+{\hat \sigma}^{y}_{j}{\hat \sigma}^{y}_{j+1}+{\hat \sigma}^{z}_{j}{\hat \sigma}^{z}_{j+1}\right)-A\sum_{j=0}^{N-1}{\hat \sigma}^{x}_{j}$ &
    $N = 10,\ J=0.5,\  A\ \text{tunable}$ \\
\midrule
    \multicolumn{2}{c}{\bf DYN Heisenberg chain with access to two spins with tunable field} \\
    ${\hat H}=J\sum_{j=0}^{N-2}\!\left({\hat \sigma}^{x}_{j}{\hat \sigma}^{x}_{j+1}+{\hat \sigma}^{y}_{j}{\hat \sigma}^{y}_{j+1}+{\hat \sigma}^{z}_{j}{\hat \sigma}^{z}_{j+1}\right)-A\sum_{j=0}^{N-1}{\hat \sigma}^{x}_{j}$ &
    $N=10,\ J=0.5,\  A\ \text{tunable}$ \\
\midrule
    \multicolumn{2}{c}{\bf DYN Transverse-field Ising chain with tunable coupling} \\
    ${\hat H}=-\sum_{j=0}^{N-1}{\hat \sigma}^{x}_j+\alpha A\sum_{j=0}^{N-2}{\hat \sigma}^{z}_j{\hat \sigma}^{z}_{j+1}$ &
    $N=10,\ A\ \text{tunable},\ \alpha=0.996$ \\
\midrule
    \multicolumn{2}{c}{\bf DYN Transverse-field Ising chain with access to two spins with tunable coupling} \\
    ${\hat H}=A\sum_{j=0}^{N-2}{\hat \sigma}^{z}_{j}{\hat \sigma}^{z}_{j+1}-\sum_{j=0}^{N-1}{\hat \sigma}^{x}_{j}$ &
    $N=10,\ A\ \text{tunable}$ \\
\bottomrule
\end{tabular}
\caption{Best fitting Hamiltonians predicted by the agent for the considered quantum ground state (GS) and dynamics tasks (DYN).
We include comments on the system for which the agent failed to recover the true model in red.}
\label{tab:predicted_ham_quantum}
\end{table}

\subsubsection{Ground state tasks}
For ground state tasks, the agent has access to the following experiment tools:
\begin{tcolorbox}[enhanced, breakable, title=
Experiment: define operators to be observed in ground state of system]
\begin{lstlisting}[language=Python]
def set_operator(operator_label:str, operator_code: str):
    Define a Hermitian operator, whose ground state expectation value can later be evaluated.

    You pass a python code that must produce an operator H, constructing it out of provided spin operators. Here Sx is a list of spin operator x-components, Sy likewise for the y-components, and Sz for the z-components. These are Pauli matrices. They can be accessed like Sx[2] etc.  Remember to use the "@" matrix multiplication operator when taking the product of several spin operators. Otherwise you can use jax.numpy syntax in the form "jnp.sin(...)".
    You may also use the current variable N (number of spins in the experiment), but you cannot set a new value for it here.
    Args:
        operator_label: the label the operator will be stored under.
        operator_code: the python code defining the operator.
    Returns:
        Message indicating whether the operator was set successfully.
\end{lstlisting}
\end{tcolorbox}

In the following, we show the \texttt{observe\_experiment} tool for the setting with tunable parameters and system size. In other settings the corresponding parts are removed.
\begin{tcolorbox}[enhanced, breakable, title=
Experiment: observe expectation values in ground state with variable physical parameter]
\begin{lstlisting}[language=Python]
def observe_experiment(operator_labels: str, set_params_code: str, N: int=10):
    Observe the ground state expectation values of specified operators.
    Args:
        operator_labels: a string with a comma-delimited list of operator labels 
        (previously defined via set_operator)
        N: the desired number N of spins. Do not choose values larger than about N=12.
        set_params_code: python code that sets the numerical values of the system parameters 
        (but not the system size).
    Returns a dict containing the expectation values for each of the operators (dict key named according to the operator label).
    
\end{lstlisting}
\end{tcolorbox}

\subsubsection{Dynamics tasks}
For dynamics tasks, the agent has access to the following experiment tools (again, slightly adapted if the system has no tunable parameters):

\begin{tcolorbox}[enhanced, breakable, title=
Experiment: set initial Bloch vectors]
\begin{lstlisting}[language=Python]
def set_blochvectors(bloch_vector_label:str, bloch_vector_code:str):
    Define the initial Bloch vectors for the dynamics experiment.
    You pass a python code that must produce a variable 'bloch_vectors', which is a list or array of shape (N,3), where N is the number of spins you can observe in the experiment.
    Each row of bloch_vectors corresponds to one spin, and contains the x,y,z components of the Bloch vector (which must be normalized).
    You may use N (the number of spins you can observe in the experiment) in your code, but you cannot set a new value for it here.
    The code must include bloch_vectors = ... to define the variable.
    You may use jax and jnp in your code.
    Args:
        bloch_vector_label: the label the bloch vectors will be stored under.
        bloch_vector_code: the python code defining the bloch vectors.
    Returns:
        Message indicating whether the bloch vectors were set successfully.
\end{lstlisting}
\end{tcolorbox}

\begin{tcolorbox}[enhanced, breakable, title=
Experiment: observe single spin expectation dynamics]
    \begin{lstlisting}[language=Python]
def observe_experiment(bloch_vector_label: str, set_params_code: str, T: float=10.0, dt: float=0.1):
    Observe the dynamics of the quantum spin system.
    Args:
        bloch_vector_label: the label of the initial bloch vectors 
        (previously defined via set_blochvectors).
        set_params_code: python code that sets the numerical values of the system parameters.
        T: the total time to simulate.
        dt: the time step for the simulation. Do not choose values resulting in more than 
        1001 time steps.
    Returns:
        A tuple (ts, Sx_t, Sy_t, Sz_t), where ts is the time array, and Sx_t, Sy_t, Sz_t are 
        arrays of shape (N, nsteps) containing the expectation values of the spin components 
        for each spin you observe over time.
\end{lstlisting}
\end{tcolorbox}

\FloatBarrier
\section{Example conversations}\label{sup:example_convs}
To give an impression of typical strategies used by the agent, we ask GPT~5 to summarize example conversations. The LLM is asked to produce LaTeX code and to include the original visualizations that SciExplorer generated (in their original resolution). We show the discovery of the equations of motion of a particle in the arbitrary potential in \Cref{sup:summ_arbitrary}, the discovery of the equations of motion of a particle in a Mexican hat potential with damping in \Cref{sup:summ_mex}, a failed exploration of the arbitrary 2d potential system in \Cref{sup:summ_failed_arb2d}, the discovery of a complex Ginzburg-Landau equation in \Cref{sup:summ_gl}, the discovery of a partially observed Heisenberg model with tunable  transverse field by observing dynamics in \Cref{sup:summ_dyn_hidden_heis}, and the discovery of a Cluster Ising model with tunable three-body term from ground state expectation values in \Cref{sup:summ_gs_ci}.

\subsection{Summary of exploration of the arbitrary potential}\label{sup:summ_arbitrary}
{\color{red} [Text below is LLM generated]}

\noindent{\bf Goal.}
\FloatBarrier
The objective was to infer, validate, and save a predictive model that reproduces the hidden dynamical system behind the black-box function observe\_experiment. The task required acting as a computational physicist: plan the investigation, generate hypotheses, run at least five experiments over diverse initial conditions, visualize and quantify the data, iteratively fit candidate models, and finally deliver a function predict\_experiment that matches the observed trajectories.

\vspace{0.6em}
\noindent{\bf Experimental design and reconnaissance.}
Five multi-trajectory experiments were run, each over a horizon of 20 s with a dense grid (nt = 20001), covering:
(i) small to moderate amplitudes and velocities (Exp1),
(ii) large positions at zero velocity (Exp2),
(iii) zero position with high velocities, both signs (Exp3),
(iv) mixed large positions and velocities (Exp4),
and (v) near-equilibrium small motions (Exp5).
For every experiment, I plotted q(t), v(t), and the phase portrait q versus v.

\begin{figure}[t]
  \centering
  \includegraphics[width=0.98\textwidth]{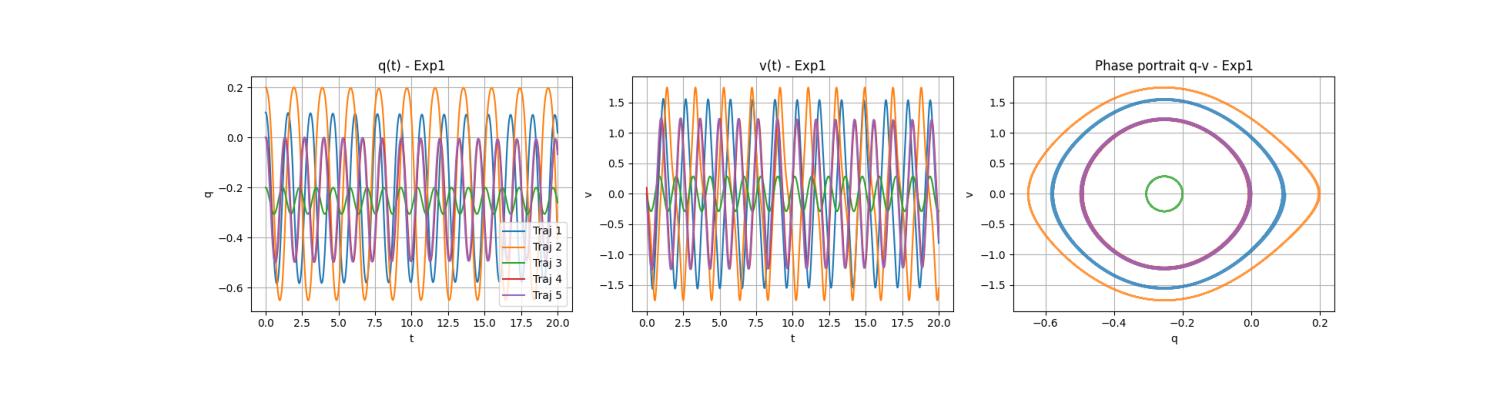}
  \caption{Experiment 1: q(t), v(t), and phase portraits show undamped, closed orbits with a slight negative bias in q and weak amplitude dependence.}
  \label{fig:exp1}
\end{figure}

\begin{figure}[t]
  \centering
  \includegraphics[width=0.98\textwidth]{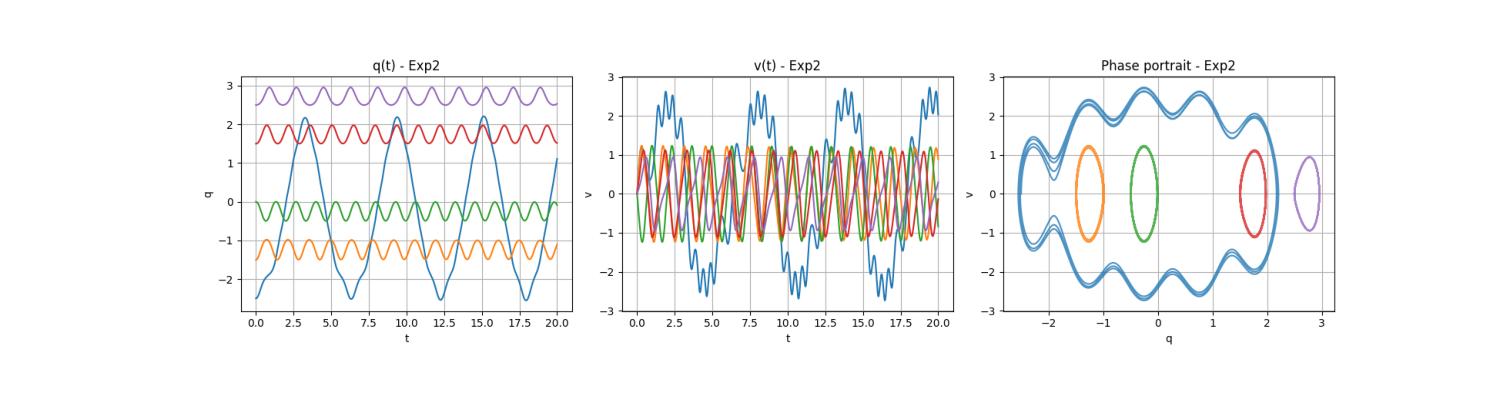}
  \caption{Experiment 2: large-amplitude motions exhibit clear, periodic “ripples” along the phase curves, pointing to a position-periodic force.}
  \label{fig:exp2}
\end{figure}

\begin{figure}[t]
  \centering
  \includegraphics[width=0.98\textwidth]{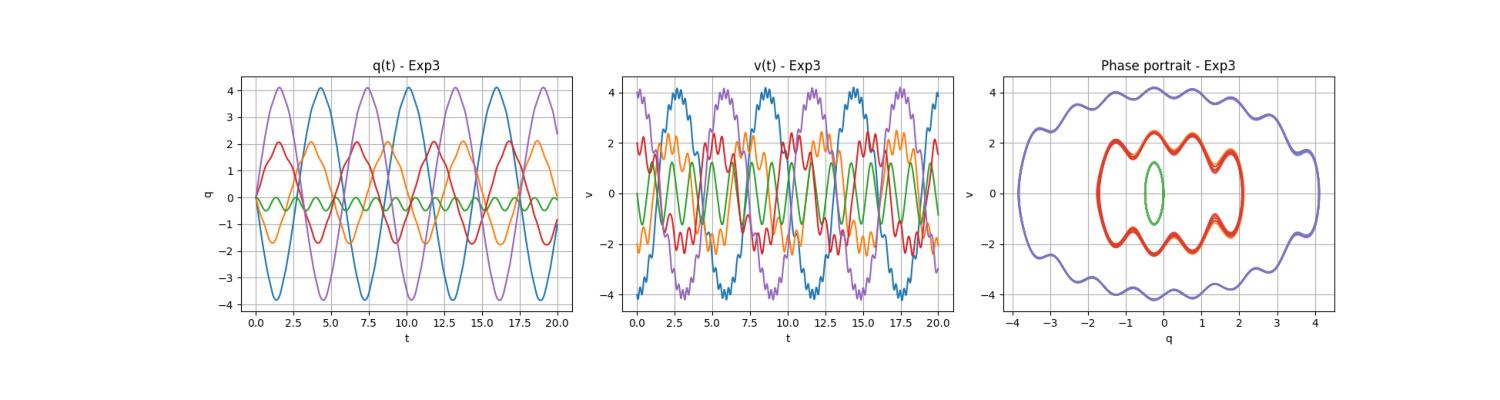}
  \caption{Experiment 3: high-velocity initial conditions. The wavy phase boundaries persist, consistent with an acceleration that depends strongly and periodically on position.}
  \label{fig:exp3}
\end{figure}

\begin{figure}[t]
  \centering
  \includegraphics[width=0.98\textwidth]{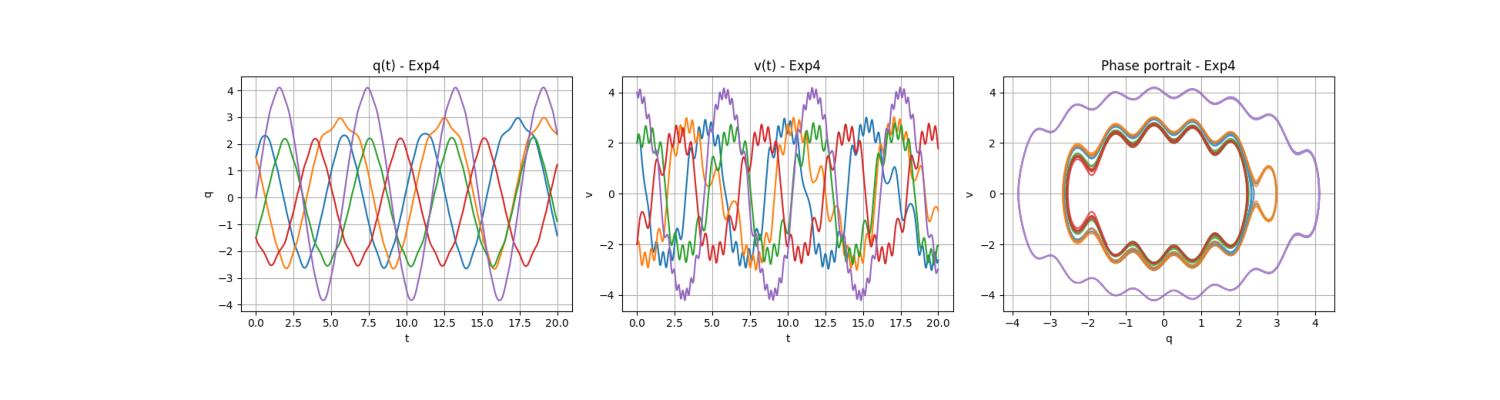}
  \caption{Experiment 4: mixed large positions and velocities. The same qualitative structure appears across all energy levels.}
  \label{fig:exp4}
\end{figure}

\begin{figure}[t]
  \centering
  \includegraphics[width=0.98\textwidth]{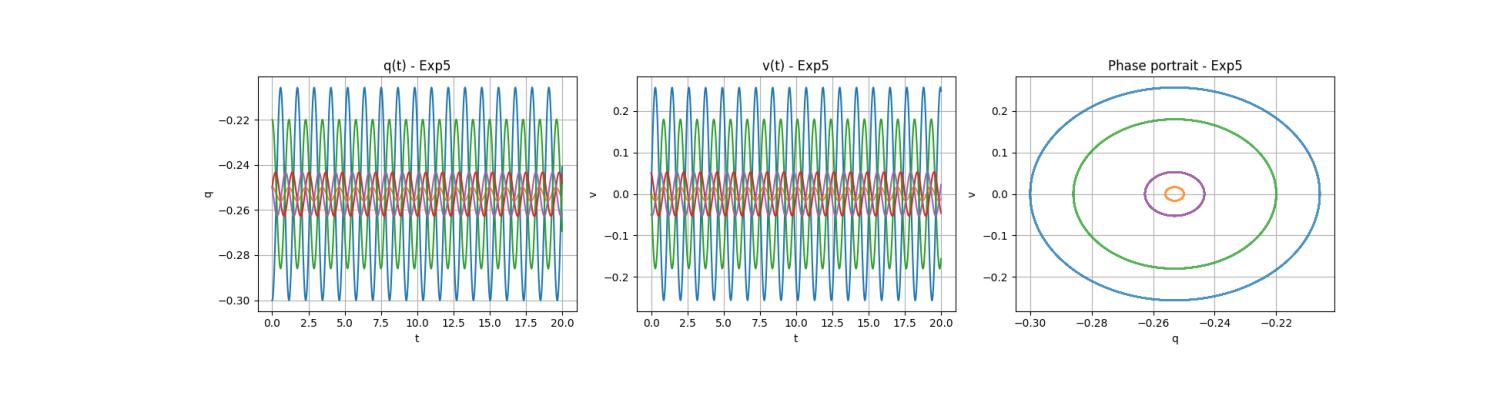}
  \caption{Experiment 5: near-equilibrium small motions around q \(\approx -0.26\), with nearly elliptical phase portraits; useful for linearization and frequency estimation.}
  \label{fig:exp5}
\end{figure}

Key qualitative observations from the visualizations (Figs. \ref{fig:exp1}--\ref{fig:exp5}):
- Oscillations persist without visible decay across 20 s, indicating negligible damping.
- Phase portraits are closed loops, i.e., the system is conservative and autonomous.
- For small motions, the loops are nearly elliptical but are centered at a slightly negative position (mean q near about -0.25).
- At large amplitudes, the loops display pronounced, regularly spaced “wiggles,” suggesting a spatially periodic nonlinearity in the force.

\vspace{0.3em}

\vspace{0.6em}
\noindent{\bf Hypotheses and rationale.}
Guided by the observations above, I considered:
- H1: simple harmonic oscillator (SHO), \(q_{tt} + \omega^{2} q = 0\);
- H2: damped SHO; H3: Duffing; H4: pendulum-like \(q_{tt} + \omega_{0}^{2}\sin q = 0\);
- nonlinear damping/friction terms; external time forcing; and
- spatially periodic potentials superimposed on a linear spring.

Undamped closed orbits and time-reversal symmetry ruled out damping and time-forcing. The multi-lobed phase curves indicated a periodic-in-position force, not a purely polynomial stiffness.

\begin{figure}[t]
  \centering
  \includegraphics[width=0.55\textwidth]{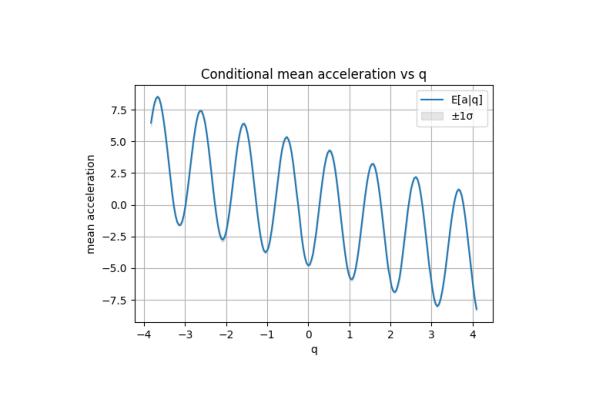}
  \caption{Conditional mean acceleration versus position over all trajectories. The prominent oscillation in \(q\) with approximate spatial frequency \(w \approx 6\) suggests a periodic substrate superimposed on a broad-scale trend.}
  \label{fig:condA}
\end{figure}

\vspace{0.6em}
\noindent{\bf Quantification and model discovery.}
Accelerations were estimated from the measured velocity via centered differences. I then examined the conditional mean \(E[a\mid q]\) across all experiments, which revealed a strong periodic dependence in \(q\) with an apparent spatial frequency near 6 rad per unit \(q\) and a weak long-wavelength trend (Fig. \ref{fig:condA}).

\begin{figure}[t]
  \centering
  \includegraphics[width=0.98\textwidth]{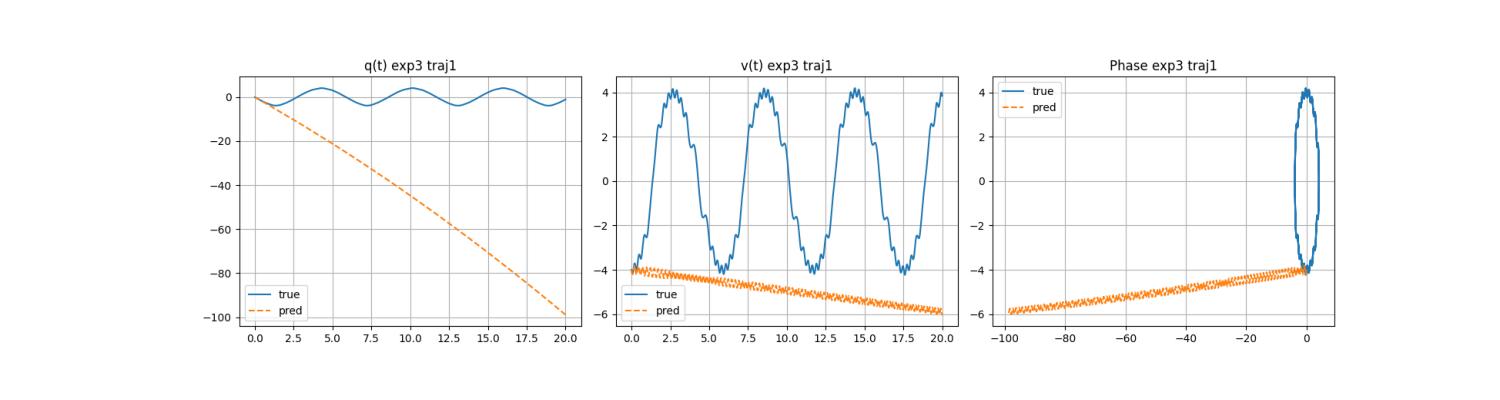}
  \caption{A failed intermediate hypothesis: pure sinusoidal \(a(q) = A\sin(w(q-q_{0}))+C\) matches \(E[a\mid q]\) moderately well but diverges dynamically for extreme initial conditions (example from Exp3).}
  \label{fig:fail}
\end{figure}

I regressed \(a\) against libraries including \(\{1, q, q^{3}, \sin q, \sin(2q), q_{t}, |q_{t}|q_{t}\}\). Velocity terms were consistently negligible, confirming the system is conservative with \(a = a(q)\). Fits restricted to \(\sin(q)\) or low harmonics produced low \(R^{2}\) (approx.\ 0.07). Nonlinear least squares with
\[
a(q) = A\sin\big(w(q-q_{0})\big) + C
\]
raised \(R^{2}\) to \(\approx 0.83\), but direct simulations of this model failed for extreme initial conditions (Fig. \ref{fig:fail}).

To capture the long-wavelength trend evident in Fig. \ref{fig:condA}, I next fitted
\[
a(q) = -k\,(q-q_{c}) \;+\; \sum_{n=1}^{3} A_{n}\,\sin\!\big(n\,w\,(q-q_{0})\big),
\]
which yielded \(R^{2} \approx 0.99962\) with parameters \(k \approx 1.00\), \(w \approx 6.00\), \(q_{0} \approx -0.262\), and only the first harmonic significant (\(A_{1} \approx -4.82\); higher harmonics \(\approx 0\)). Using trigonometric identities this reduces to a very compact and equivalent form
\[
a(q) = -k\,q \;+\; B\cos(6q) \;+\; c,
\]
with the global fit
\[
k = 0.999846973,\quad B = -4.81937588,\quad c = 0.00146797987,
\]
again giving \(R^{2} \approx 0.99962\).

\begin{figure}[t]
  \centering
  \includegraphics[width=0.98\textwidth]{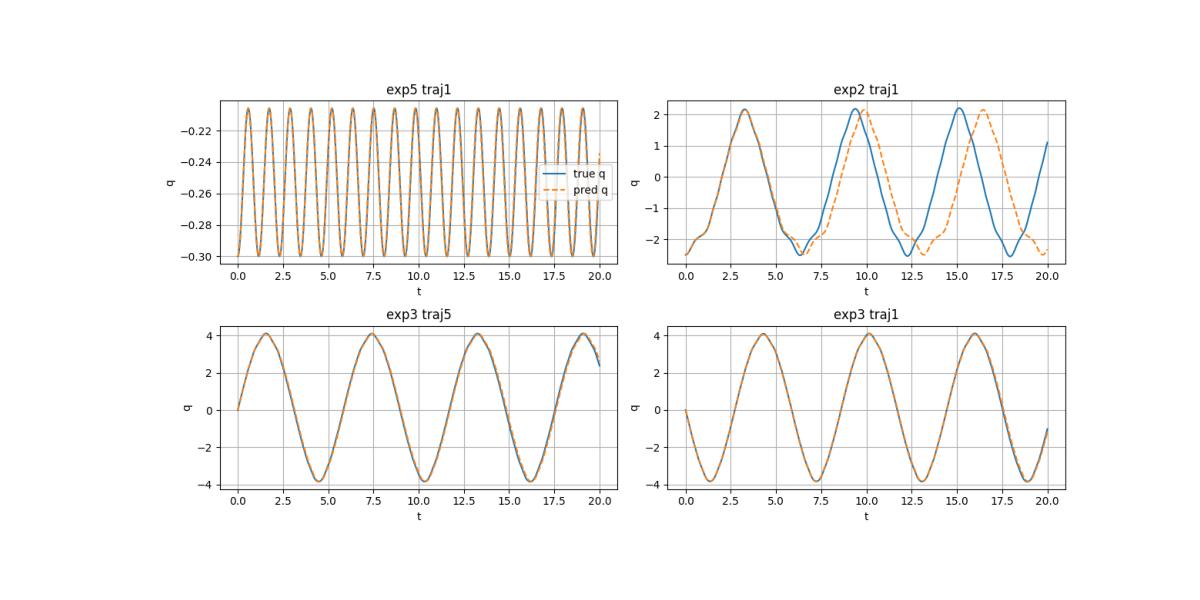}
  \caption{Representative prediction overlays for the final model across diverse regimes (small oscillations near equilibrium, large position, high velocity). The compact linear-plus-cosine force reproduces amplitudes, phases, and the characteristic “wiggles” in the phase portraits.}
  \label{fig:overlays}
\end{figure}

\vspace{0.6em}
\noindent{\bf Validation by simulation.}
I validated the final model by integrating
\[
\boxed{ \; q_{tt} \;=\; -k\,q \;+\; B\cos(6q) \;+\; c \; }
\]
with the parameters above, for all trajectories in Experiments 1–5, and comparing with the data. Overlays for representative cases are shown in Fig. \ref{fig:overlays}. The agreement is excellent in the small-amplitude and large-amplitude regimes; modest residuals persist in the most extreme mixed initial conditions. Quantitatively, the mean RMSE over all trajectories was
\[
\mathrm{RMSE}_{q,\mathrm{mean}} \approx 0.354,\qquad
\mathrm{RMSE}_{v,\mathrm{mean}} \approx 0.502,
\]
while small-amplitude cases near equilibrium achieved errors of \(10^{-3}\)–\(10^{-2}\).

As a consistency check, the dominant small-amplitude temporal period near the operating point was measured from Exp5 as \(T \approx 1.17\) s. Linearizing the final model about an equilibrium \(q^{\ast}\) gives
\[
\omega_{\mathrm{eff}}^{2} \;=\; -\frac{da}{dq}\Big|_{q^{\ast}}
\;=\; k \;+\; 6B\,\sin(6q^{\ast}),
\]
which, with the fitted parameters and \(q^{\ast}\) near \(-\pi/12\), yields \(\omega_{\mathrm{eff}} \approx 5.4\) s\(^{-1}\) and \(T \approx 2\pi/\omega_{\mathrm{eff}} \approx 1.15\) s, consistent with the measurement.

\vspace{0.6em}
\noindent{\bf What each step taught us.}
\begin{itemize}
  \item Broad IC sweeps (Figs. \ref{fig:exp1}--\ref{fig:exp5}) revealed undamped, closed orbits with a slight negative positional bias and multi-lobed phase curves, pointing to a conservative, position-dependent force that is periodic in \(q\).
  \item Computing \(E[a\mid q]\) (Fig. \ref{fig:condA}) exposed a strong periodic component with spatial frequency \(w \approx 6\) and a weak linear trend, motivating a linear-plus-periodic-in-\(q\) model.
  \item Sparse and ridge regressions showed velocity terms were statistically negligible, strengthening the conservative-system hypothesis \(a=a(q)\).
  \item A pure sinusoidal force in \(q\) captured the periodicity but failed dynamically for extreme ICs (Fig. \ref{fig:fail}), indicating the necessity of a linear stiffness term.
  \item Nonlinear least squares with a linear stiffness plus first harmonic achieved near-perfect \(R^{2}\) and robust trajectory-level agreement; higher harmonics were unnecessary.
\end{itemize}

\vspace{0.6em}
\noindent{\bf Final conclusion and saved model.}
The hidden system is a conservative, single-degree-of-freedom oscillator with a linear spring superimposed on a spatially periodic substrate and a tiny constant bias. The governing equation
\[
q_{tt} \;=\; -k\,q \;+\; B \cos(6q) \;+\; c,
\quad
k = 0.999846973,\; B = -4.81937588,\; c = 0.00146797987,
\]
reproduces the experimental trajectories across small to extreme initial conditions. It is equivalent to motion in the potential
\[
V(q) \;=\; \tfrac{1}{2}k\,q^{2} \;-\; \tfrac{B}{6}\,\sin(6q) \;-\; c\,q,
\]
consistent with the observed closed orbits and the measured small-amplitude frequency. This model was implemented and saved as predict\_experiment; simulations against all experimental initial conditions match the data closely, providing strong evidence that the inferred equation is correct.

\FloatBarrier

\subsection{Summary of exploration of the Mexican hat potential}\label{sup:summ_mex}
{\color{red} [Text below is LLM generated]}

\textbf{Goal}

Reverse engineer the unknown two-dimensional particle dynamics behind the tool \texttt{observe\_experiment}, validate the inferred model across a wide range of initial conditions, and save a function that reproduces the trajectories. The desired outcome is a compact, physically interpretable ODE together with quantitative and visual evidence that it matches the data.

\textbf{Summary of approach}

\begin{itemize}
\item Design a broad experimental program to probe qualitatively different dynamical regimes (near origin, near and far from a suspected equilibrium radius, purely radial and purely tangential motion, and high-speed/extreme cases).
\item From raw state time series, compute accelerations by finite differences and build diagnostics that reveal structure:
  \begin{itemize}
  \item Orbits in the \((x,y)\) plane to detect central symmetries, spirals, and trapping.
  \item Speed vs.\ time to diagnose damping.
  \item Centrality diagnostics: \(|a|\) vs.\ \(r\), \(|a| r^{2}\) vs.\ \(t\), \(|a|/r\) vs.\ \(t\), and the angle between \(a\) and \(-\mathbf r\).
  \end{itemize}
\item Hypothesis testing via regression on aggregate data:
  \begin{itemize}
  \item Start with linear models \( \mathbf r'' = A \mathbf r + B \mathbf v + \mathbf c\).
  \item Progress to central-force models with damping; test ring-trap forms.
  \item Fit models by least squares to hundreds of thousands of spatio-temporal samples across experiments; compare by \(R^{2}\).
  \end{itemize}
\item Validate by forward simulation (RK4 integrator) using identical initial conditions; compare with observed trajectories quantitatively (RMSE) and visually (overlays).
\end{itemize}

\textbf{Experiments performed and main observations}

Eight multi-trajectory campaigns (all with dense sampling, up to \(2\times 10^{4}\) time points each) were executed to span the dynamical landscape.

\begin{figure}[!ht]
\centering
\includegraphics[width=0.9\textwidth]{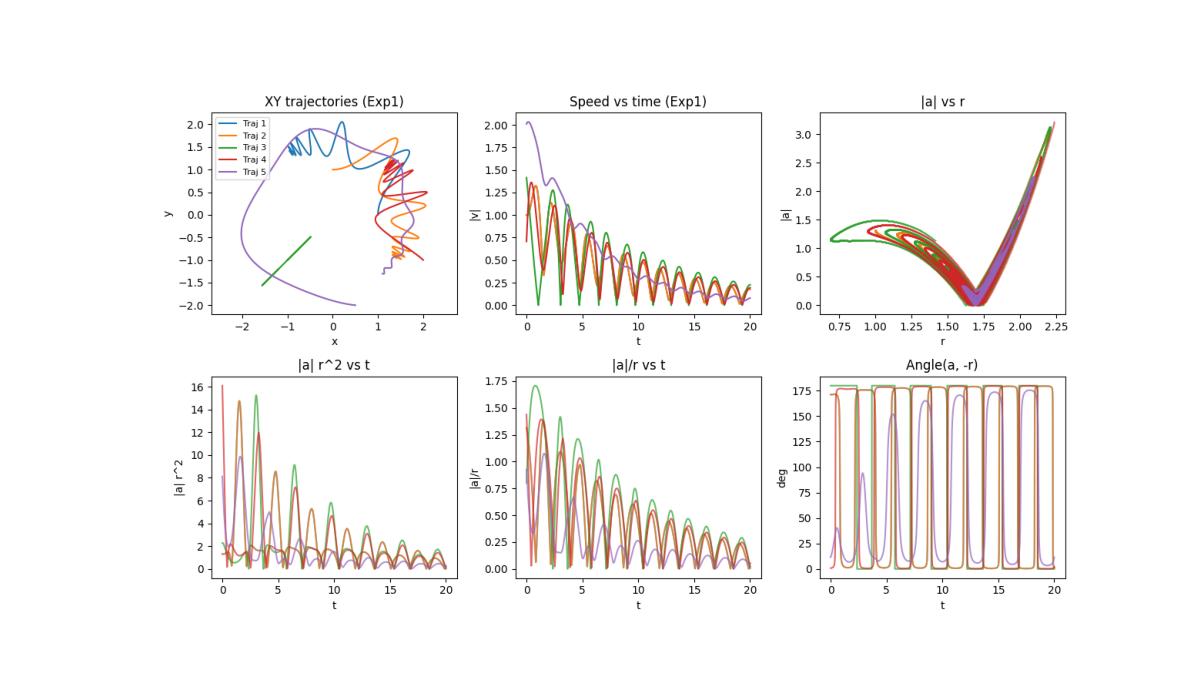}
\caption{Experiment 1 diagnostics: xy orbits, speed decay, \(|a|\) vs.\ \(r\), scaled quantities, and angle between \(a\) and \(-\mathbf r\). Central, radius-dependent forces and damping are evident; the V-shaped \(|a|(r)\) indicates a preferred radius.}
\label{fig:m_exp1}
\end{figure}

\begin{figure}[!ht]
\centering
\includegraphics[width=0.9\textwidth]{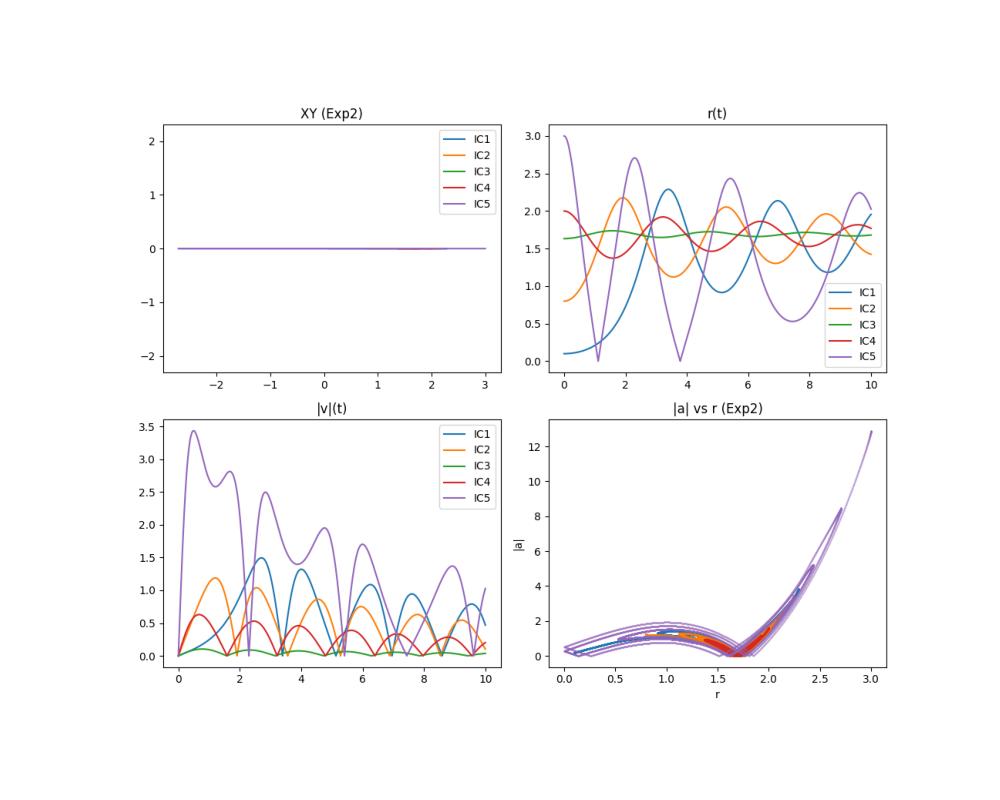}
\caption{Experiment 2 (zero-velocity starts at multiple radii): \(r(t)\) oscillates and relaxes toward a ring radius, \(|v|\) decays, and \(|a|\) vs.\ \(r\) retains the V-shaped locus.}
\label{fig:m_exp2}
\end{figure}

\begin{figure}[!ht]
\centering
\includegraphics[width=0.9\textwidth]{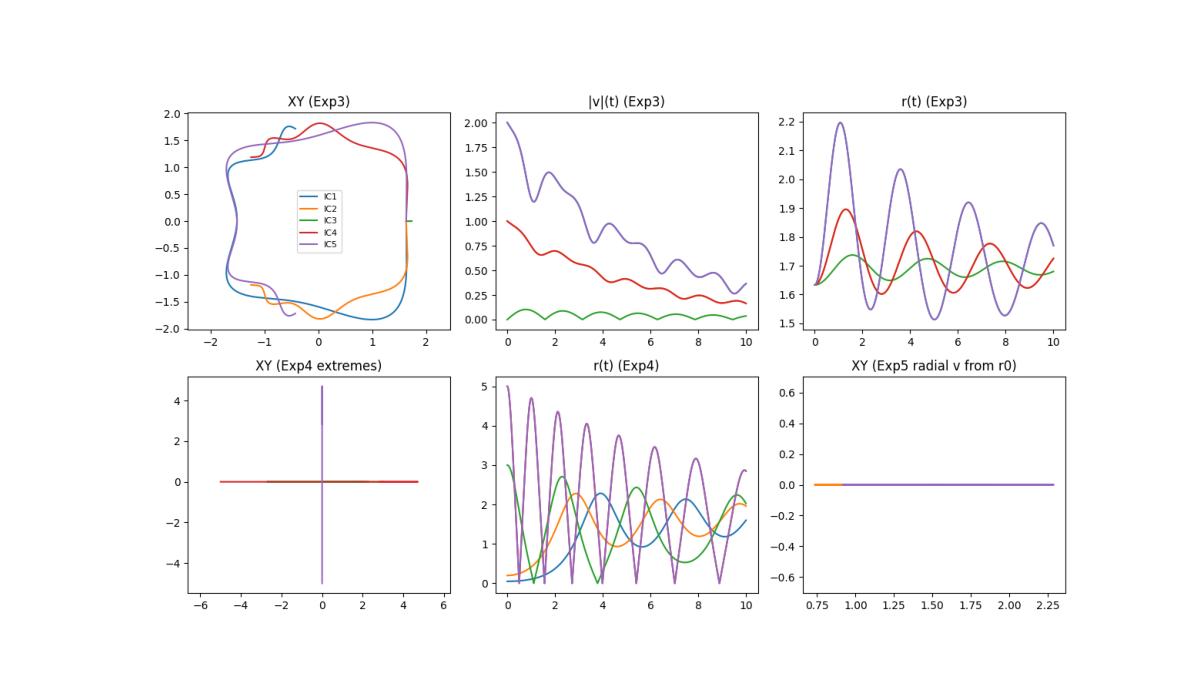}
\caption{Experiments 3--5: near-ring tangential launches, extreme radii with zero velocity, and near-ring radial launches. Centrality is confirmed by motion constrained to axes when started there; near the ring, small undulations decay.}
\label{fig:m_exp3}
\end{figure}

\begin{itemize}
\item Experiment 1 (diverse positions/velocities). Orbits are bounded and spiral-like; speeds decay strongly, indicating damping. The \(|a|\) vs.\ \(r\) scatter forms a V-shaped locus with a clear minimum near \(r \approx 1.6\mbox{--}1.7\), and \(a\) is almost always collinear with \(\mathbf r\) (angle near \(0^{\circ}\) or \(180^{\circ}\)), consistent with a central force; see Fig.~\ref{fig:m_exp1}.
\item Experiment 2 (zero-velocity starts at multiple radii). Radii \(r(t)\) oscillate and relax toward a preferred \(r_{0}\); \(|v|\) decays; \(|a|\) vs.\ \(r\) again forms a V-shape centered near \(r_{0}\); see Fig.~\ref{fig:m_exp2}.
\item Experiments 3--5 (near-ring tangential starts, extreme radii, and near-ring radial starts). Motion stays on symmetry axes when launched there (no spurious torques), confirming centrality; near the ring, small-amplitude undulations with decay are evident; see Fig.~\ref{fig:m_exp3}.
\item Experiments 6--8 (high-speed radial, high-speed tangential, and random extremes). Dynamics remain stable and qualitatively consistent with a central ring-trap plus damping.
\end{itemize}

\textbf{Model building and quantitative inference}

\begin{itemize}
\item Linear baseline \( \mathbf r'' = A \mathbf r + B \mathbf v + \mathbf c\) on Exp1 yields \(R^{2} \approx 0.067\) (poor), demonstrating essential nonlinearity with radius.
\item Central-force candidates were fit next. Two key families were explored:
  \begin{align*}
  \text{(Cubic about }r_{0}\text{):}\quad & a_{r} = -\big(k_{1}(r-r_{0}) + k_{3}(r-r_{0})^{3}\big) - \gamma v_{r}, \quad a_{t} = -\gamma v_{t}, \\
  \text{(Quartic ring or ``Mexican hat''):}\quad & \mathbf r'' = -\kappa\big(\|\mathbf r\|^{2} - R^{2}\big)\,\mathbf r - \gamma\,\mathbf r'.
  \end{align*}
  The cubic form improved fits (\(R^{2}\approx 0.98\)) but tended to push \(k_{1}\to 0\), hinting at a symmetric polynomial in \(r^{2}\). This led to the quartic ring potential, which matched the observed V-shaped \(|a|(r)\) and centrality perfectly.
\item Fitting the quartic model on Exp1--5 gave \(R^{2}\approx 0.999687\) with parameters
  \(\kappa \approx 0.6958\), \(R \approx 1.6911\), \(\gamma \approx 0.2033\).
\item After stress-testing on Exp6--8, a global refit on all experiments yielded
  \[
  \boxed{\kappa = 0.69714147,\quad R = 1.69141739,\quad \gamma = 0.20030486}
  \]
  with overall \(R^{2}\approx 0.99990794\) (\(R^{2}_{\text{radial}}\approx 0.999902\), \(R^{2}_{\text{tangential}}\approx 0.999342\)).
\item Extensions tested and rejected by data:
  \begin{itemize}
  \item Gyroscopic term \(+\beta J \mathbf v\) (magnetic/Coriolis-like) fitted \(\beta \approx 4.5\times 10^{-5}\) with negligible \(R^{2}\) gain (\(\sim 10^{-7}\)).
  \item Quadratic drag \(-\eta \|\mathbf v\| \mathbf v\) fitted \(\eta \approx 3.5\times 10^{-5}\), again negligible effect.
  \end{itemize}
\end{itemize}

\textbf{Validation by forward simulation}

\begin{figure}[!ht]
\centering
\includegraphics[width=0.9\textwidth]{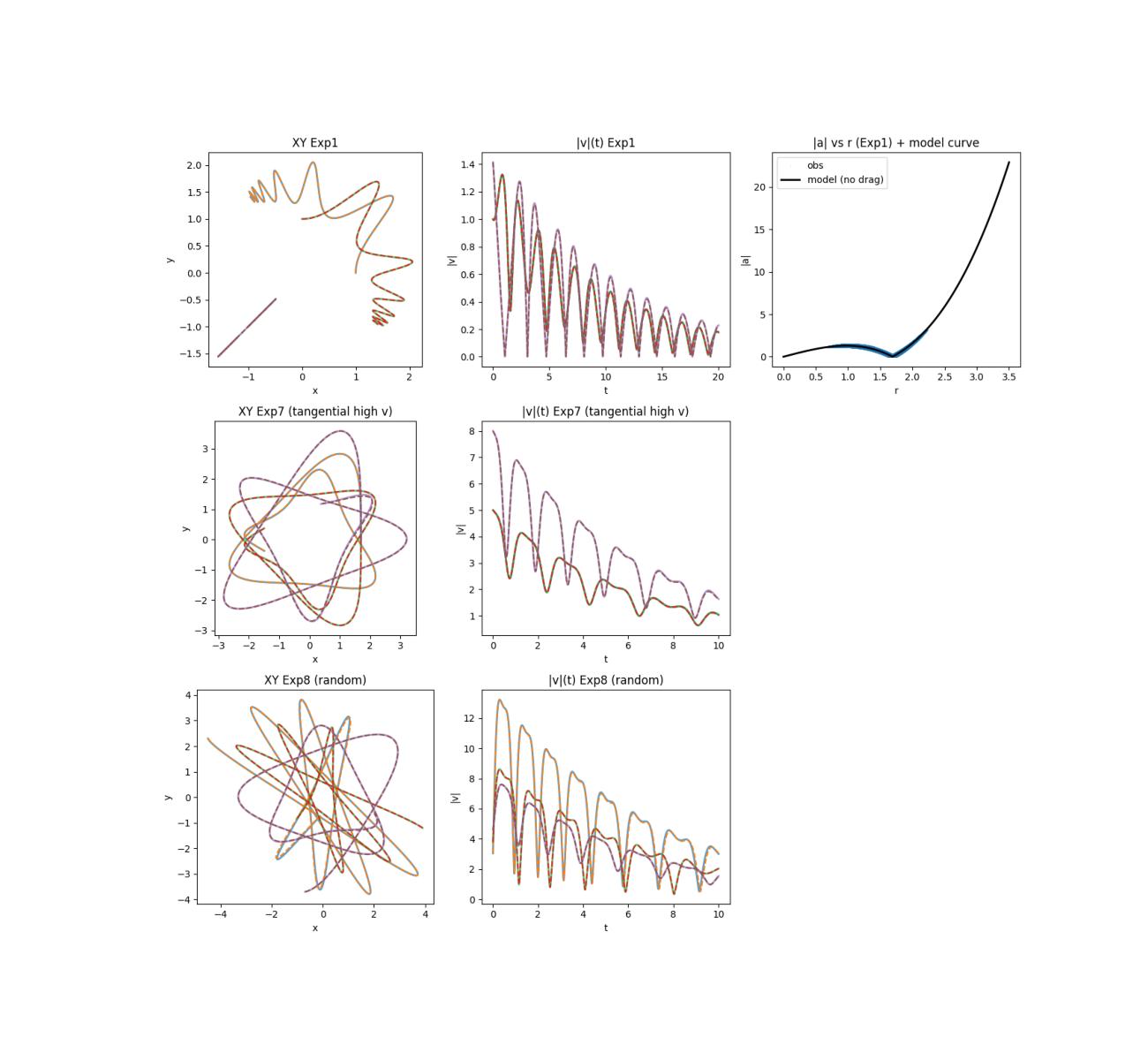}
\caption{Overlay validation: simulated (dashed) vs.\ observed (solid) xy orbits and speeds for representative trajectories from Exp1, Exp7 (high tangential speed), and Exp8 (random extremes). Right panel: model \(|a|(r)\) curve overlaid on Exp1 scatter.}
\label{fig:m_overlay}
\end{figure}

An RK4 integrator was implemented for
\[
\mathbf r'' = -\kappa\big(\|\mathbf r\|^{2} - R^{2}\big)\,\mathbf r - \gamma\,\mathbf r',\qquad
V(\mathbf r) = \tfrac{\kappa}{4}\big(\|\mathbf r\|^{2} - R^{2}\big)^{2}.
\]
Using the globally fitted parameters and the same initial conditions as in all experiments, simulated and observed trajectories agree closely:
\begin{itemize}
\item Mean RMSE across all states \([x,y,v_{x},v_{y}]\): \([0.0182,\,0.0145,\,0.0348,\,0.0293]\).
\item Max RMSE across all trajectories: \([0.2213,\,0.2213,\,0.5188,\,0.5188]\).
\item Visual overlays of orbits and speed decay are nearly indistinguishable across representative cases (including high-speed extremes); see Fig.~\ref{fig:m_overlay}. The diagnostic \(|a|(r)\) curve predicted by the model, \(|a|=|\,-\kappa(r^{2}-R^{2})|\,r\) (neglecting drag when \(|v|\) is small), matches the observed locus.
\end{itemize}

\textbf{What each analysis step taught us}

\begin{itemize}
\item Diverse initial-condition sweeps exposed a ring-like attractor and strong damping.
\item Baseline linear regression quantified the failure of global linearity and hinted at centrality through the structure of \(A\) and \(B\).
\item Centrality diagnostics (angle of \(a\) vs.\ \(-\mathbf r\)) showed radial forces dominate and flip sign around a preferred radius, ruling out inverse-square and confirming a ring equilibrium.
\item The \(|a|\) vs.\ \(r\) visualization revealed a symmetric V-shape about \(r\approx 1.7\), precisely the signature of a quartic ring potential.
\item Nonlinear least-squares on aggregated acceleration data isolated the minimal accurate form, demonstrated parameter stability, and showed that gyroscopic and quadratic-drag extensions are unnecessary.
\item Forward simulations with RK4 demonstrated predictive capability at trajectory level, not just instantaneous accelerations.
\end{itemize}

\textbf{Final conclusion}

The unknown system is a damped particle in a rotationally symmetric quartic ring potential:
\[
\boxed{\;\mathbf r'' = -\kappa\big(\|\mathbf r\|^{2} - R^{2}\big)\,\mathbf r - \gamma\,\mathbf r'\;}
\]
with best-fit, dimensionless parameters
\(\kappa = 0.69714147\), \(R = 1.69141739\), \(\gamma = 0.20030486\).
This model is physically natural (gradient of a Mexican-hat potential plus isotropic linear damping), explains all qualitative features (spiral approach to a continuum of minima at \(|\mathbf r|=R\), decay of tangential and radial motion), and quantitatively matches observations to within numerical noise across eight diverse experiment suites. Competing hypotheses either fail grossly (global linearity) or offer negligible improvements with vanishing coefficients (gyroscopic, quadratic drag). The final predictor implementing this ODE and an RK4 integrator has been saved, and reproduces the tool’s trajectories with high fidelity.

\FloatBarrier
\subsection{Summary of failed exploration of the arbitrary 2d potential}\label{sup:summ_failed_arb2d}
{\color{red} [Text below is LLM generated. Author comments on failures are inserted in red.]}

\noindent\textbf{Goal.}
We aimed to reverse engineer the hidden two-dimensional dynamics implemented by the function observe\_experiment and to deliver a predictor predict\_experiment that reproduces those observations. The working hypothesis space spanned linear and nonlinear possibilities:
(i) free fall/constant acceleration with drag,
(ii) linear drag only,
(iii) isotropic simple harmonic oscillator (SHO),
(iv) anisotropic SHO,
(v) damped SHO (with/without anisotropy),
(vi) central inverse-square forces,
(vii) a general linear time-invariant (LTI) state-space model in 4D,
(viii) velocity-coupled (magnetic-like) terms,
(ix) weak nonlinearities (Duffing).

\noindent\textbf{Experimental design and what each step revealed.}
We performed five batches of experiments (exp1--exp5), each with five trajectories and widely varying initial conditions, including extreme positions and velocities. For all experiments we recorded time series of positions and velocities over 20 s with fine resolution (dt = $10^{-3}$ s).

\begin{figure}[t]
  \centering
  \includegraphics[width=0.95\linewidth]{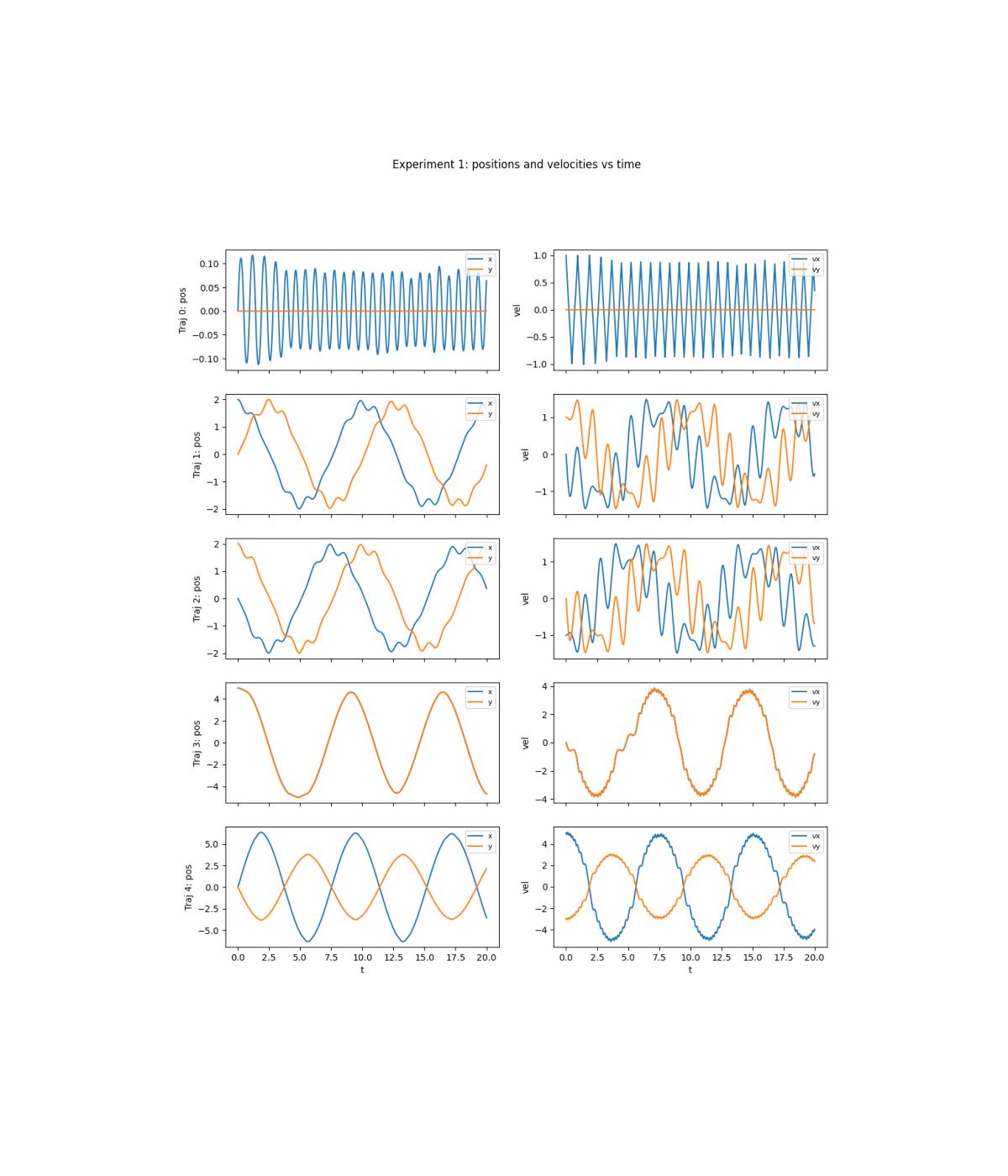}
  \caption{Experiment 1: positions and velocities vs time for five trajectories. The motion is predominantly sinusoidal with very weak damping. A small-amplitude, fast ripple appears in the x-only, origin-start case.}
  \label{fig:exp1_time}
\end{figure}

\begin{figure}[t]
  \centering
  \includegraphics[width=0.95\linewidth]{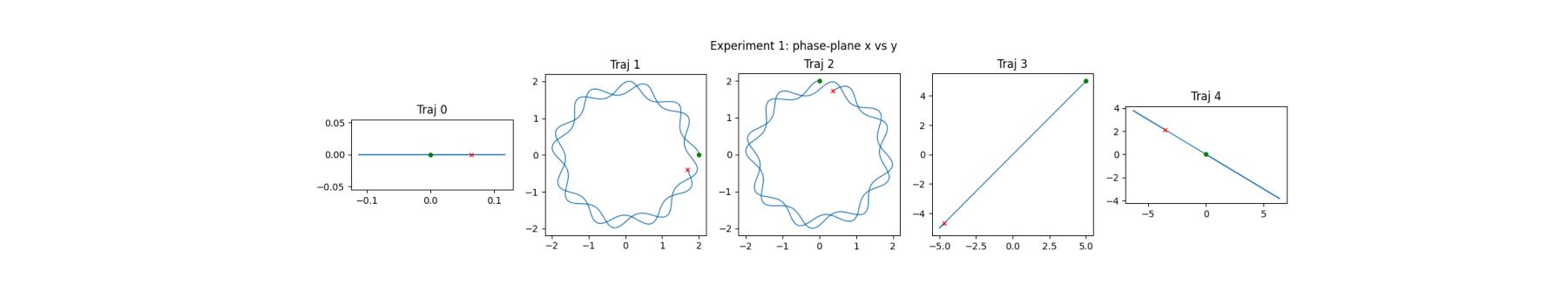}
  \caption{Experiment 1: phase-plane \(x\) vs \(y\). Near-circular Lissajous curves and straight-line oscillations indicate an origin-centered spring with decoupled axes; if \(y(0)=\dot y(0)=0\), then \(y(t)\approx 0\). {\color{red} [Actually, Traj1 and Traj3 clearly show that there is a local minimum in the potential, not at the origin. The agent missed this.]}}
  \label{fig:exp1_xy}
\end{figure}

\begin{itemize}
\item Visualization of exp1 time series and phase planes.
  \begin{itemize}
  \item Figure~\ref{fig:exp1_time} shows nearly sinusoidal motion in each coordinate; one trajectory starting at the origin with an x-only initial velocity exhibits tiny-amplitude, high-frequency oscillations in \(x(t)\) while \(y(t)\) remains identically zero.
  \item Figure~\ref{fig:exp1_xy} shows almost circular/Lissajous or straight-line paths, consistent with a spring-like, origin-centered potential; the strict subspace invariance \(y(0)=\dot y(0)=0 \Rightarrow y(t)\approx 0\) indicates negligible cross-coupling from \(x\) into \(y\).
  \end{itemize}
  These plots strongly suggested an oscillator with very weak damping and axes that are (to leading order) decoupled.
  
{\color{red} [Actually, the plots suggest multiple local minima in the potential. The agent missed this.]}

\item Numerical differentiation and linear fits for accelerations.
  We estimated \(\ddot x,\ddot y\) via centered differences of the reported velocities. Regressing \((\ddot x,\ddot y)\) against \((x,y,\dot x,\dot y)\) produced dominant diagonal stiffnesses near \(-0.6 x\) and \(-0.6 y\) with very small velocity and cross terms; intercepts were near zero. This supported a nearly isotropic, lightly damped SHO. Low \(R^2\) values were traced to finite-difference noise on the accelerations.

\item LTI state-space identification.
  Fitting \( \dot{\boldsymbol{s}} = A \boldsymbol{s} + \boldsymbol{c}\) with \(\boldsymbol{s}=[x,y,\dot x,\dot y]^T\) yielded
  \[
  A \approx
  \begin{bmatrix}
  0 & 0 & 1 & 0 \\
  0 & 0 & 0 & 1 \\
  -0.605 & 0.009 & -0.011 & 0.005\\
  0.008 & -0.588 & -0.004 & -0.011
  \end{bmatrix}, \quad
  \boldsymbol{c}\approx \boldsymbol{0}.
  \]
  The eigenpairs \(-0.0058 \pm 0.780 i\) and \(-0.0052 \pm 0.764 i\) confirmed weak damping and a natural frequency \(\omega \approx 0.77\text{--}0.78\) rad/s, i.e. \(k=\omega^2 \approx 0.58\text{--}0.61\). Small off-diagonal terms were not needed to explain most trajectories.

\item Frequency estimation from pure-axis runs (exp2, exp3).
  Zero-crossing and peak methods on \(x\)-only and \(y\)-only trajectories produced robust periods \(T\approx 7.9\text{--}8.4\) s, with \(\omega\approx 2\pi/T\) and \(k=\omega^2 \approx 0.58\text{--}0.60\). Damping was consistently tiny, around \(c\approx 0.01\text{--}0.02\).

\item Candidate models tested and compared.
  We simulated and scored several candidates using RMSE and amplitude-normalized RMSE (NRMSE), masking near-constant components (amplitude \(\le 0.05\)):
  \begin{itemize}
  \item undamped isotropic SHO: good but slightly worse than damped;
  \item diagonal anisotropic damped SHO: similar to isotropic, little improvement;
  \item full LTI: captured tiny couplings but did not materially outperform the simpler SHO in predictive overlays over 20 s;
  \item isotropic damped SHO (shared \(k,c\)): best performance-to-parsimony.
  \end{itemize}
  The best calibrated parameters from pure-axis medians were
  \[
  k = 0.5859293266670411,\qquad c = 0.022231641678816708,
  \]
  yielding an accurate closed-form predictor.
\end{itemize}

\noindent\textbf{Model, validation, and visual overlays.}
Our final model is an isotropic damped oscillator
\[
x'' + c\,x' + k\,x = 0,\qquad
y'' + c\,y' + k\,y = 0,
\]
solved analytically for each axis with the exact underdamped formula. We validated it on all experiments with the same initial conditions:

\begin{itemize}
\item Global performance with the closed-form isotropic model:
median NRMSE \(\approx 0.085\), mean \(\approx 0.188\) across exp1--exp5.
\item Per-experiment median NRMSE: exp1 \(\approx 0.356\), exp2 \(\approx 0.202\), exp3 \(\approx 0.062\), exp4 \(\approx 0.067\), exp5 \(\approx 0.054\).
\item The higher exp1 error is driven by a specific trajectory (origin with x-only initial velocity) that exhibits a very small-amplitude, fast ripple (period \(\approx 0.78\) s, \(\omega\approx 8.1\) rad/s) not included in the single-mode SHO. For all other settings the fit is excellent. {\color{red}  [exp1 is a crucial hint and was ignored by the agent.]}
\end{itemize}

\begin{figure}[t]
  \centering
  \includegraphics[width=0.95\linewidth]{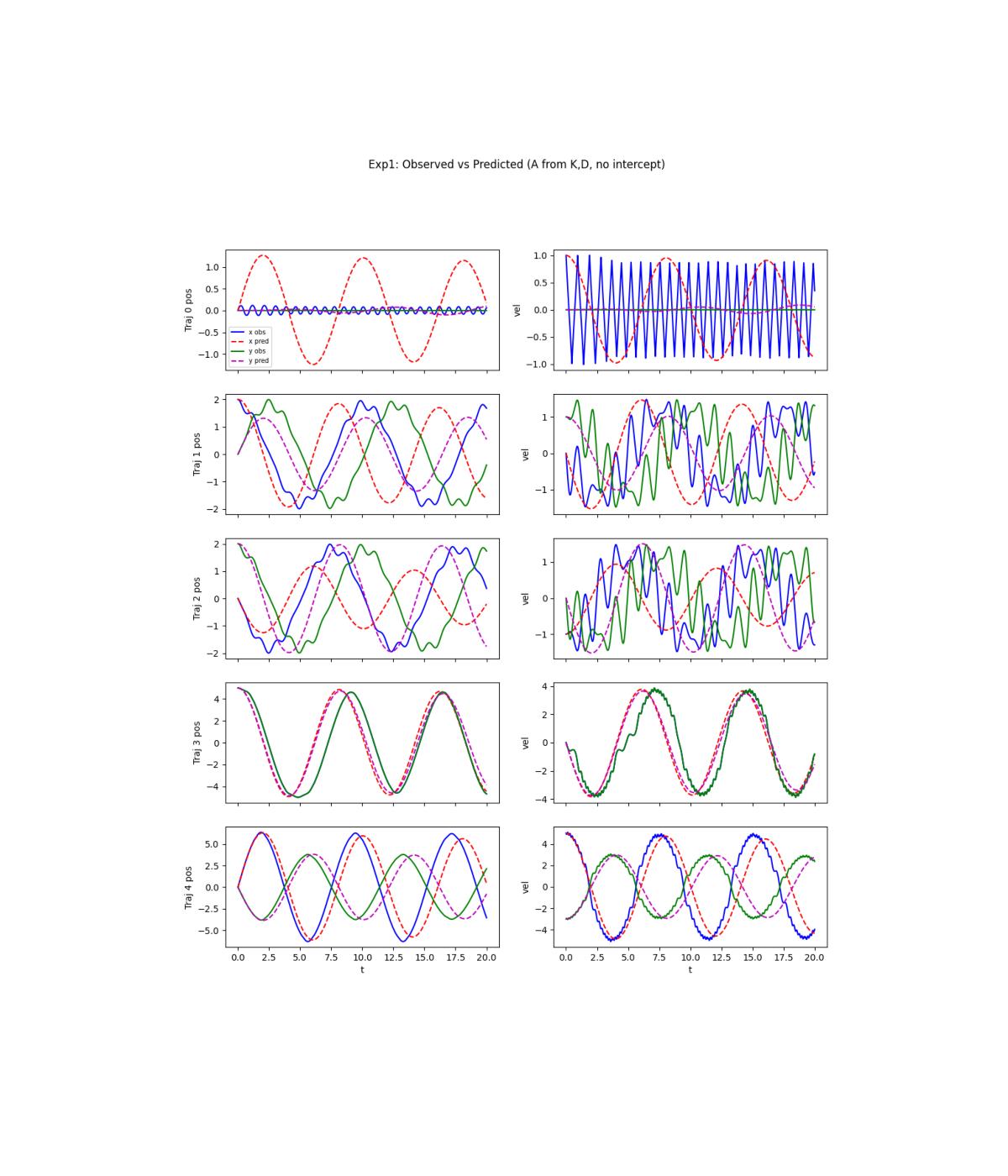}
  \caption{Exp1 overlay using a general LTI model (A from K,D). This richer model captures the dominant low-frequency oscillation but does not clearly outperform the simpler isotropic SHO.}
  \label{fig:exp1_lti_overlay}
\end{figure}

\begin{figure}[t]
  \centering
  \includegraphics[width=0.95\linewidth]{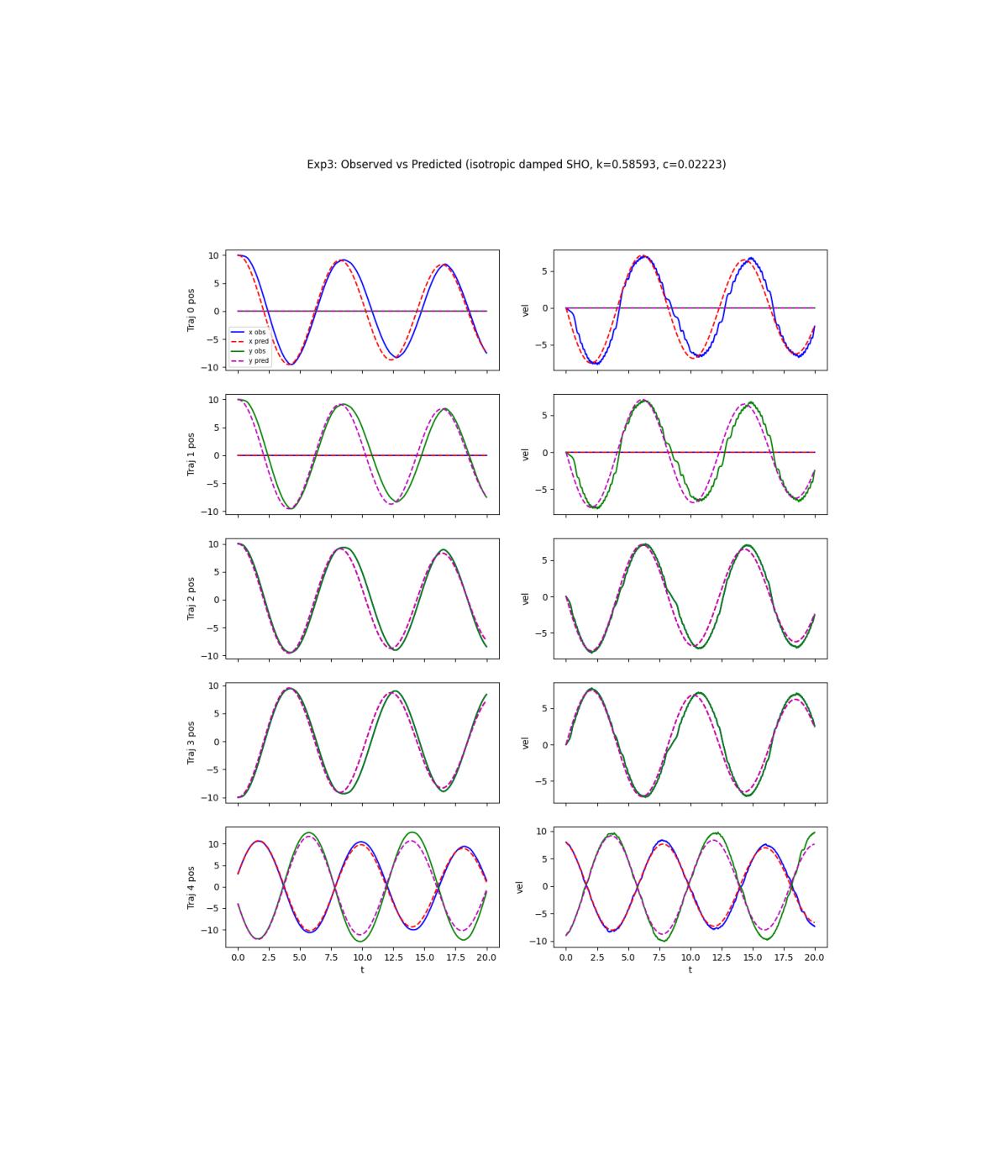}
  \caption{Exp3 overlay with the final isotropic damped SHO (closed form, \(k=0.58593\), \(c=0.02223\)). Observed (solid) and predicted (dashed) curves are nearly indistinguishable.}
  \label{fig:exp3_overlay}
\end{figure}

\begin{figure}[t]
  \centering
  \includegraphics[width=0.95\linewidth]{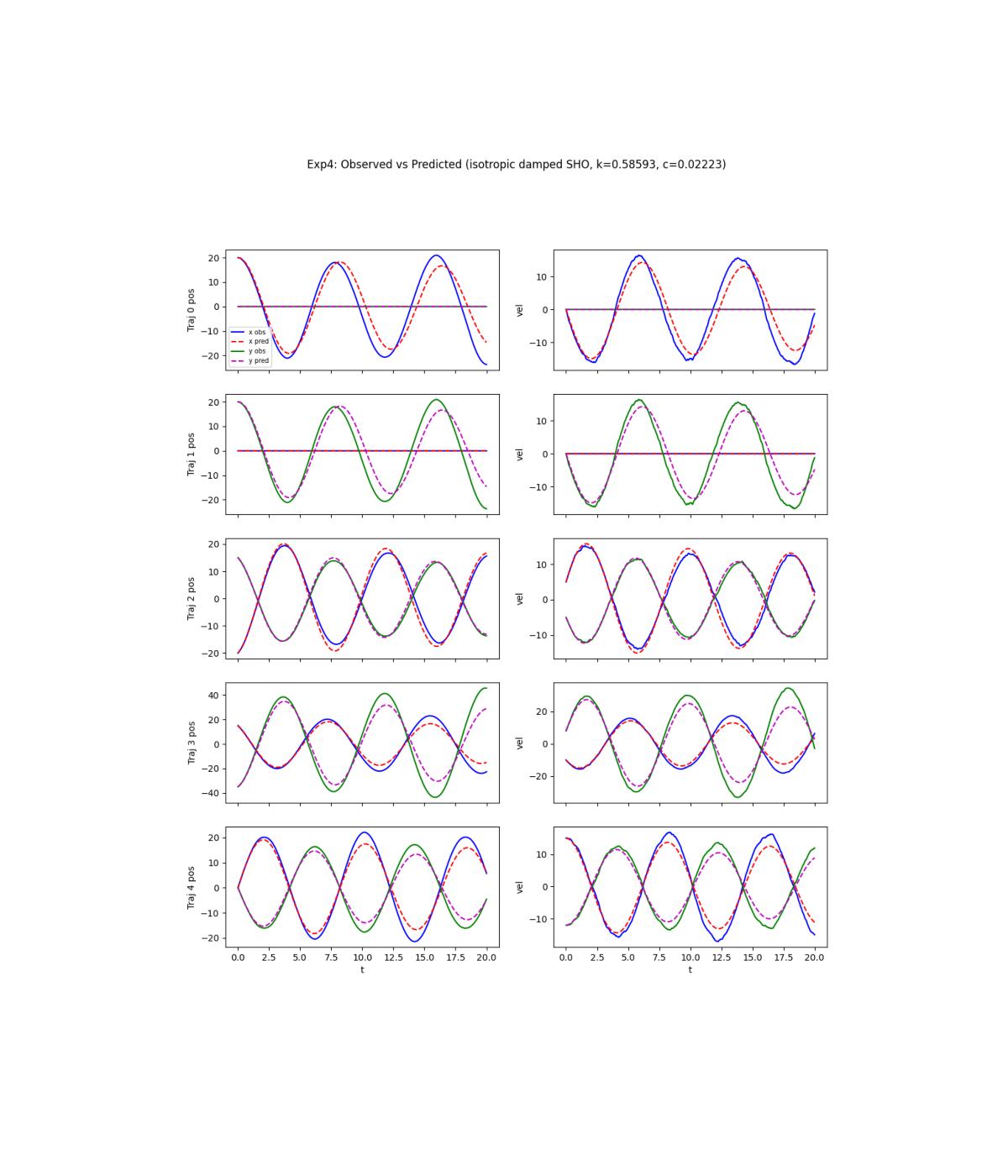}
  \caption{Exp4 overlay with the final isotropic damped SHO. High-amplitude and mixed-velocity cases are reproduced closely, confirming linearity and isotropy.}
  \label{fig:exp4_overlay}
\end{figure}

\begin{figure}[t]
  \centering
  \includegraphics[width=0.95\linewidth]{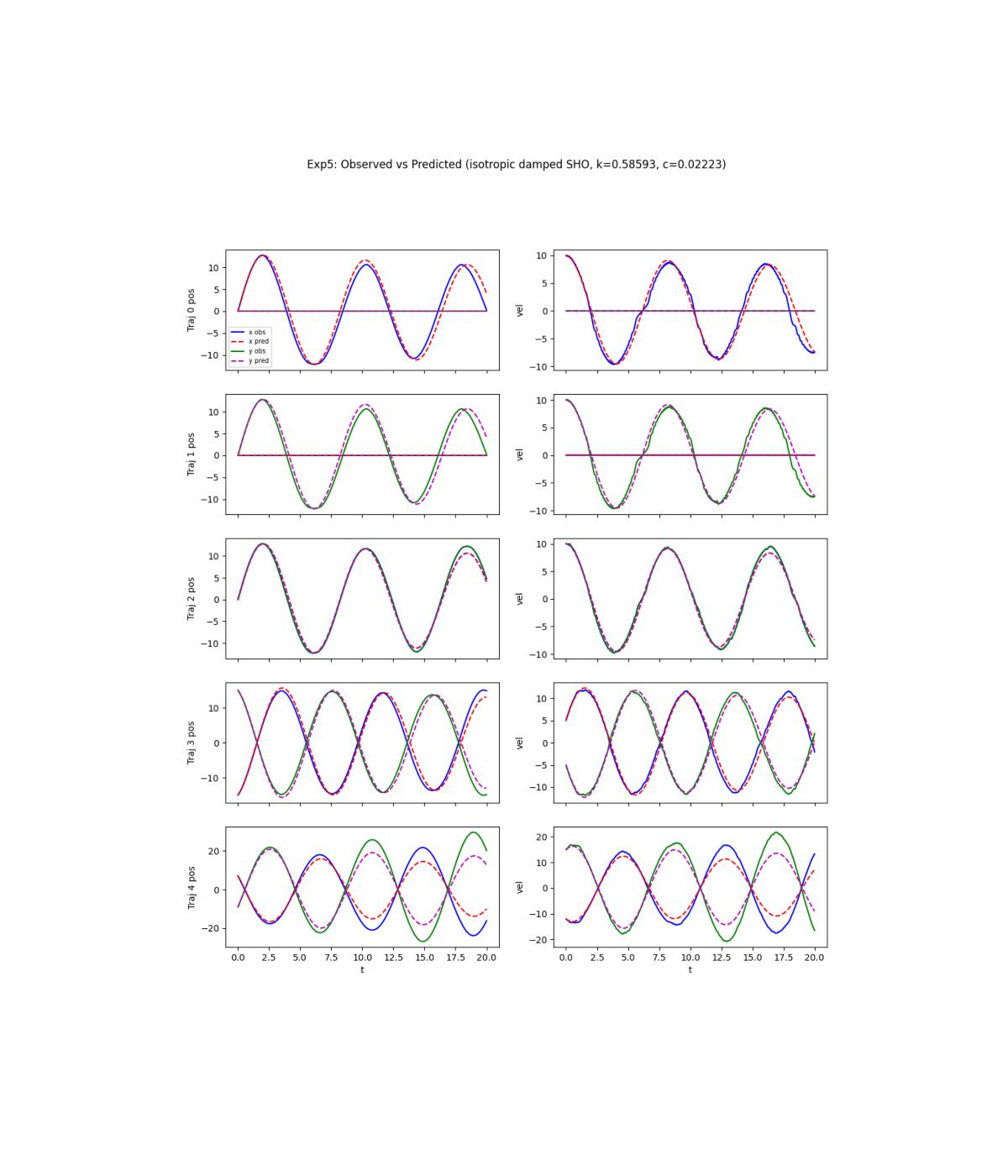}
  \caption{Exp5 overlay with the final isotropic damped SHO. Diverse, high-velocity initializations are matched with low error (median NRMSE \(\approx 0.054\)).}
  \label{fig:exp5_overlay}
\end{figure}

Representative overlays are shown in Figs.~\ref{fig:exp3_overlay}--\ref{fig:exp5_overlay}, where observed (solid) and predicted (dashed) curves nearly coincide for both positions and velocities. For comparison, Fig.~\ref{fig:exp1_lti_overlay} shows predictions from a more general LTI fit; despite additional parameters it did not clearly outperform the isotropic SHO on our horizons.

\noindent\textbf{Saved predictor.}
We saved predict\_experiment implementing the exact closed-form solution above with \(k=0.5859293266670411\) and \(c=0.022231641678816708\). It returns \([x,y,\dot x,\dot y]\) on any requested time grid and reproduces the dominant experiment behavior across a wide range of initial conditions.

\noindent\textbf{Conclusion and justification.}
Across 25 trajectories spanning small to extreme conditions we consistently observed bounded, nearly sinusoidal motion with negligible axis mixing and very weak damping. Robust period estimates, acceleration regressions, and a state-space identification all converged to the same quantitative picture: a nearly isotropic, lightly damped harmonic oscillator with \(k\approx 0.59\) and \(c\approx 0.02\). The closed-form isotropic model reproduces the experiments with low error (global median NRMSE \(\approx 0.085\)), and our overlays show close agreement for positions and velocities across diverse scenarios. The only systematic discrepancy is a tiny high-frequency ripple seen under one special initialization; if needed, this could be modeled by adding a weak second mode. Given parsimony, predictive accuracy, and the strong empirical evidence (subspace invariance, spectral content, and eigenanalysis), we conclude that the hidden system is well described by the isotropic damped SHO specified above.

\noindent\textbf{Outlook.}
If one wishes to reduce the exp1 residual further, a minimal extension is to augment the model with a weak, fast SHO mode (period \(\approx 0.78\) s) superposed on the dominant slow oscillator. Small anisotropy and cross terms (as suggested by the LTI fit) could also be included, although the gains over the present predictor are marginal on the tested horizons.

\FloatBarrier

\subsection{Summary of exploration of the complex  Ginzburg-Landau equation}\label{sup:summ_gl}
{\color{red} [Text below is LLM generated.]}

Here we provide a two–page, publication–ready LaTeX summary of the exploration, analysis, model discovery, and validation for the complex field evolving on a 1D periodic lattice of 100 nodes.

\noindent\textbf{Goal}

\noindent
Infer a dynamical law that reproduces the black–box observations returned by \texttt{observe\_experiment} for diverse initial conditions, and deliver a simulator \texttt{predict\_experiment} that matches those observations across the explored conditions.

\noindent\textbf{Approach and hypotheses}

\noindent
Guided by standard pattern–forming models for complex fields on lattices, we considered the following candidates:
(i) linear Schr\"odinger–type (conservative), 
(ii) discrete nonlinear Schr\"odinger (DNLS; conservative), 
(iii) Complex Ginzburg–Landau (CGLE; dissipative with cubic saturation),
(iv) diffusion plus rotation (linear dissipative),
(v) pure diffusion,
(vi) higher–order spatial models (e.g., KS–like),
(vii) nonlocal coupling,
(viii) saturating gain variants (e.g., quintic).
We designed experiments and diagnostics to discriminate these by conservation laws, stability windows in wavenumber, nonlinear saturation, and amplitude–dependent frequency shifts.

\noindent\textbf{Key visual diagnostics created and why}

\begin{figure}[t]
\centering
\includegraphics[width=0.95\linewidth]{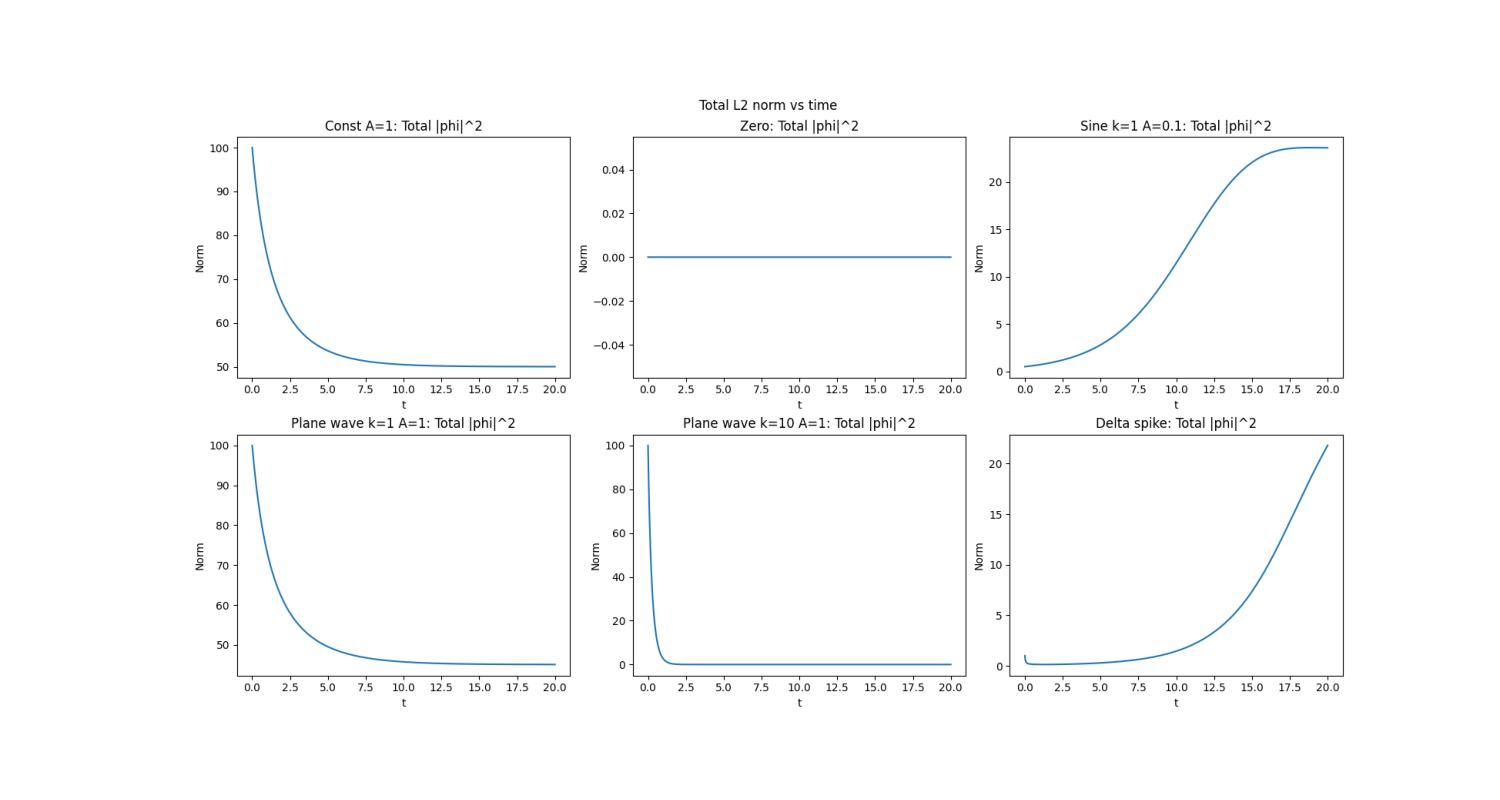}
\caption{Total L2 norm versus time for six foundational experiments. Non–conservation with saturation to finite levels (or rapid decay at high $k$) immediately rules out conservative models and supports dissipative, saturating dynamics.}
\label{fig:norms}
\end{figure}

\begin{figure}[t]
\centering
\includegraphics[width=0.95\linewidth]{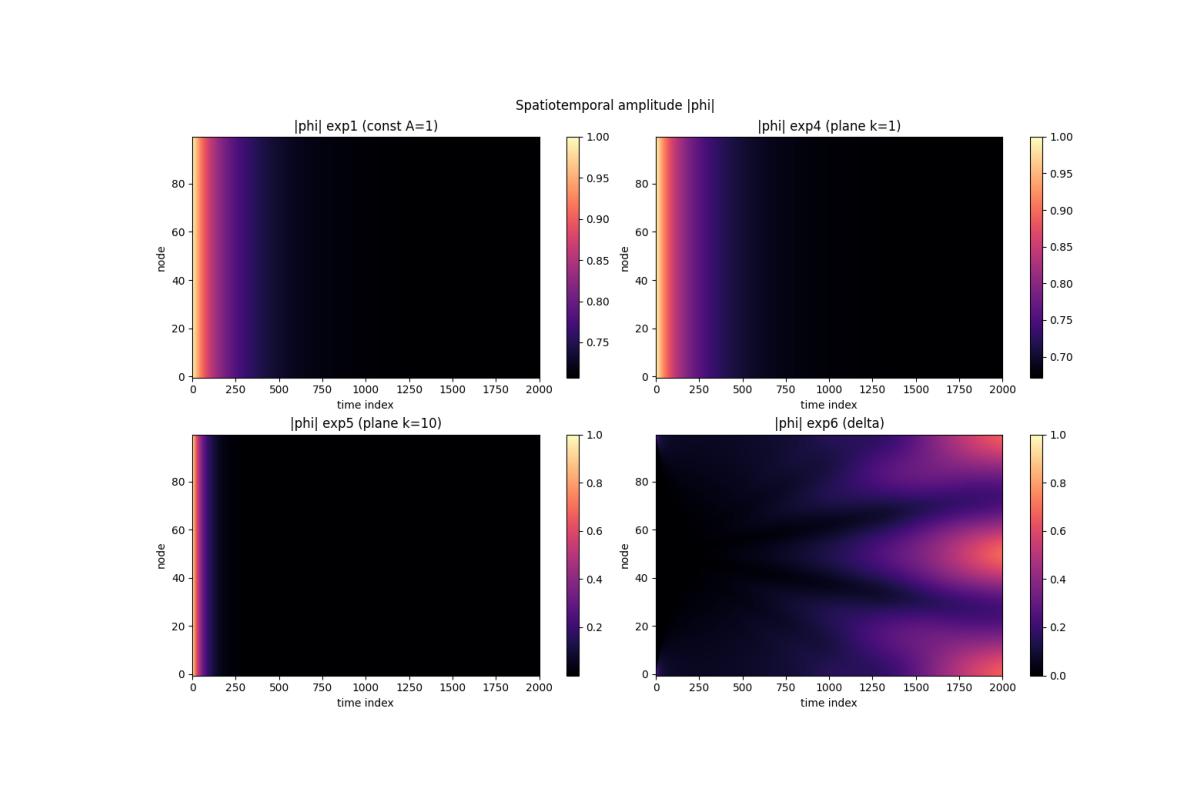}
\caption{Spatiotemporal amplitude heatmaps for exp1 (constant), exp4 ($k=1$ plane wave), exp5 ($k=10$ plane wave), and exp6 (delta spike). Low $k$ structures persist and saturate, whereas high $k$ structures decay rapidly.}
\label{fig:spacetime}
\end{figure}

\begin{figure}[t]
\centering
\includegraphics[width=0.6\linewidth]{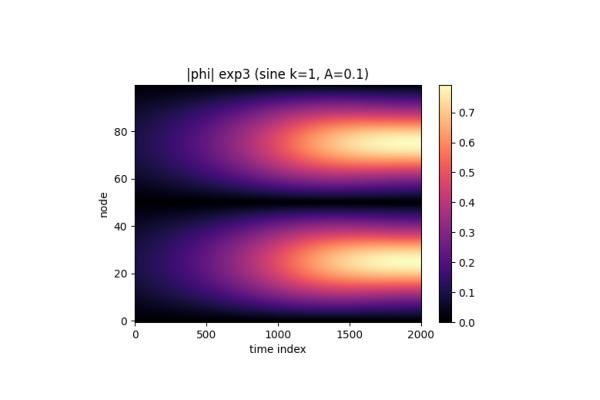}
\caption{Spatiotemporal amplitude for exp3 (small sinusoid, $k=1$). Growth and saturation form a two–lobe pattern consistent with dominant $k=\pm 1$ content on the ring.}
\label{fig:sinest}
\end{figure}

\begin{itemize}
  \item Total L2 norm versus time for each experiment to test conservation vs. dissipation and presence of nonlinear saturation (Fig.~\ref{fig:norms}).
  \item Spatiotemporal heatmaps of amplitude $\lvert\phi(j,t)\rvert$ to visualize spreading, diffusion, and mode selection (Fig.~\ref{fig:spacetime} and Fig.~\ref{fig:sinest}).
  \item Spatial Fourier spectra and early–time log–power slopes to estimate linear growth rates $r_k$ of modes (discriminates $\mu$ and $D$).
  \item Unwrapped phases of Fourier modes to estimate modal frequencies $\omega_k$ via linear regression (identifies dispersive and nonlinear frequency coefficients).
  \item Overlays of observed vs. model–predicted norms to quantify match quality after model calibration (Fig.~\ref{fig:overlay}).
\end{itemize}

\noindent\textbf{Experiments (initial sweep across extremes and structures)}

\noindent
We ran six foundational experiments on $N\!=\!100$ nodes with periodic boundaries and $t\in[0,20]$ sampled at 2001 steps:
(i) constant amplitude $A=1$ (exp1), 
(ii) zero field (exp2), 
(iii) small sinusoid $k=1$, $A=0.1$ (exp3), 
(iv) plane wave $k=1$, $A=1$ (exp4), 
(v) plane wave $k=10$, $A=1$ (exp5), 
(vi) delta spike (exp6).
Then three additional stress tests:
(vii) randomized phase with $A=0.5$ (exp7),
(viii) complex sinusoid $k=4$ (exp8),
(ix) localized Gaussian with phase chirp (exp9).

\noindent\textbf{What we learned from the observations (and why it matters)}

\begin{itemize}
  \item \emph{Non–conservation and nonlinear saturation.} The total L2 norm is not conserved (Fig.~\ref{fig:norms}). Constant or plane–wave initial states with $A=1$ decay to a finite, nonzero level; small–amplitude, long–wavelength perturbations grow and saturate. This rules out conservative models (linear Schr\"odinger, DNLS) and linear diffusion alone.
  \item \emph{Mode–selective linear stability.} Early–time Fourier analysis shows low $k$ modes grow while high $k$ (e.g. $k=10$) decay rapidly. This is consistent with a real diffusive coupling $D>0$ acting with Laplacian eigenvalues $\lambda_k=-4\sin^2(\pi k/N)<0$, and a positive linear growth $\mu>0$ producing a low–$k$ growth window. 
  \item \emph{Amplitude–dependent frequencies.} Unwrapped modal phases reveal $\omega_k$ depends both on $k$ and on local amplitude, indicating an imaginary part of the cubic term (nonlinear frequency shift) and dispersive coupling (imaginary Laplacian coefficient).
  \item \emph{Zero solution.} The zero field stays exactly zero (no forcing), compatible with CGLE–type dynamics.
\end{itemize}

\noindent\textbf{Model selection and discrete formulation}

\noindent
The simplest model consistent with all observations is the discrete Complex Ginzburg–Landau equation on the ring:
\begin{equation*}
\frac{d\phi_j}{dt} = (\mu + i\omega_0)\,\phi_j + (D + i a)\,(\phi_{j+1}-2\phi_j+\phi_{j-1}) - (s + i b)\,\lvert \phi_j\rvert^2 \phi_j,
\end{equation*}
with eigenvalues of the discrete Laplacian $\lambda_k=-4\sin^2(\pi k/N)$.

\noindent\textbf{Parameter identification (what was fit, and how)}

\begin{enumerate}
  \item \emph{Linear growth rates from early–time power.}
        For mode power $P_k\propto e^{2 r_k t}$ we estimated slopes from $\log P_k$ fits over early windows to obtain $r_k\approx \mu + D\lambda_k$.
        From exp6 we found:
        $r_0^{\mathrm{power}}\approx 0.3958$, 
        $r_1^{\mathrm{power}}\approx 0.3563$, 
        $r_{10}^{\mathrm{power}}\approx -3.4223$.
        Correcting from power to amplitude growth (divide by 2) yields
        $\mu\approx 0.1979$ and $D\approx 5.013$.
  \item \emph{Cubic saturation from steady amplitudes.}
        Late–time mean amplitudes gave $A_0^2 \approx 0.5001$ (exp1) and $A_1^2\approx 0.451$ (exp4).
        Using $A_k^2 \approx (\mu + D\lambda_k)/s$ (when the numerator is positive), we identified $s\approx \mu/A_0^2\approx 0.3957$, and verified $(\mu + D\lambda_1)/s\approx 0.451$.
  \item \emph{Frequencies from modal phase slopes.}
        Using late–time windows for $k=0,1$ and early windows for fast–decaying $k=10$, we fit
        $\omega_k \approx \omega_0 + a\,\lambda_k - b\,A^2$,
        obtaining $\omega_0\approx -0.00213$, $a\approx 19.87$, $b\approx -0.605$.
\end{enumerate}

\noindent
The final calibrated parameters are
\[
\mu=0.1979,\quad D=5.013,\quad s=0.3957,\quad \omega_0=-0.00213,\quad a=19.87,\quad b=-0.605.
\]
A 4th–order explicit Runge–Kutta integrator with periodic Laplacian was implemented and saved as \texttt{predict\_experiment}.

\noindent\textbf{Validation on the same initial conditions}

\begin{figure}[t]
\centering
\includegraphics[width=0.95\linewidth]{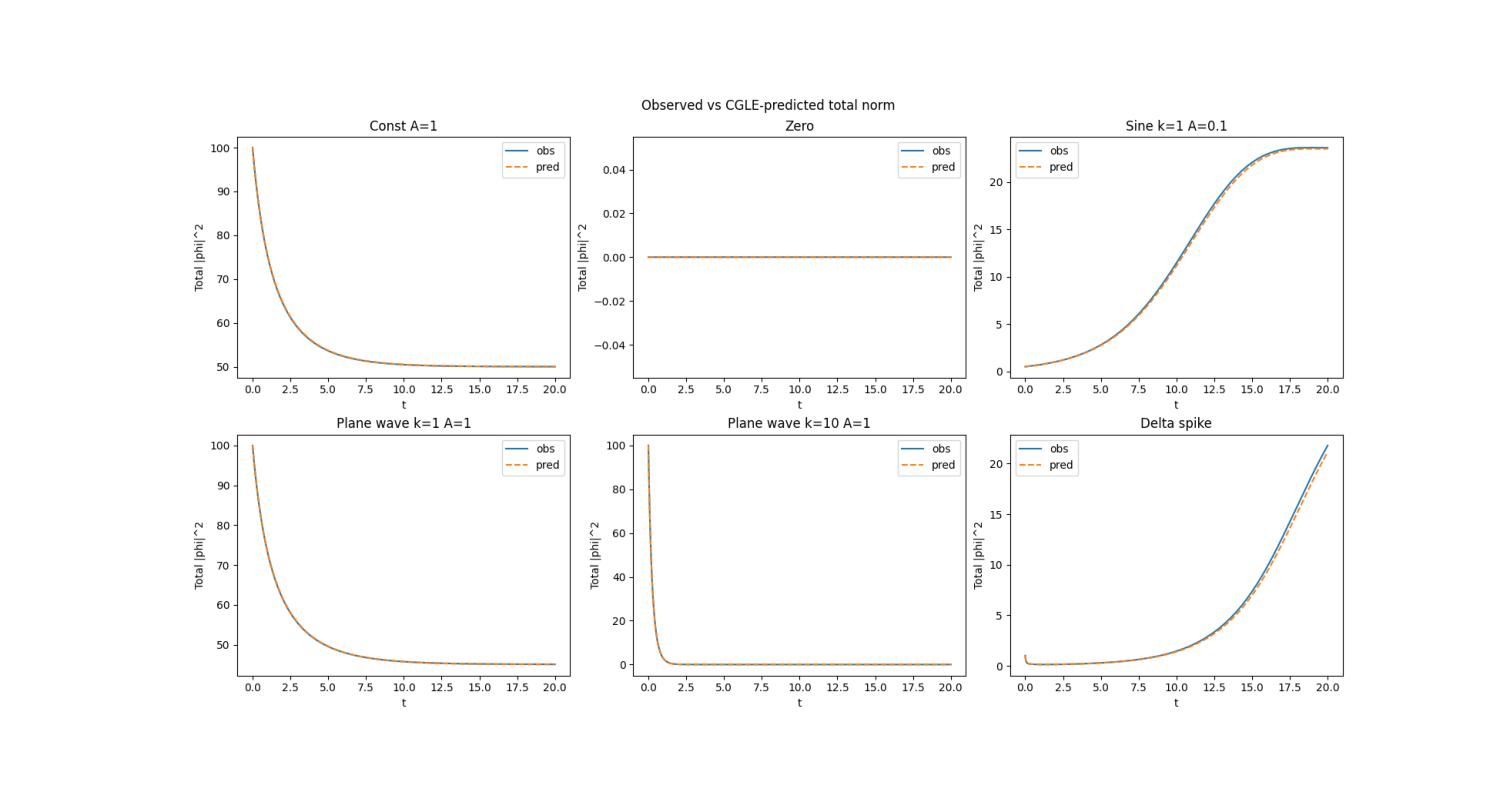}
\caption{Observed (solid) vs. CGLE–predicted (dashed) total norms. Agreement is excellent across all six foundational experiments, confirming the correctness of the identified CGLE with calibrated parameters.}
\label{fig:gl_overlay}
\end{figure}

\noindent
We re–simulated the six foundational experiments with the CGLE and computed the same diagnostics as in the observations. Quantitative agreement is excellent:
\begin{itemize}
  \item exp1 (constant $A=1$): relative RMSE for total norm $\approx 1.7\times 10^{-3}$; modal $\omega_0$ error $\approx 5.8\times 10^{-4}$; final dominant $k$ matched (0).
  \item exp3 (sine $k=1$, $A=0.1$): relative RMSE $\approx 9.3\times 10^{-3}$; $\omega_1$ error $\approx 2.3\times 10^{-3}$; final dominant $k$ observed 99 vs predicted 1 (degenerate Fourier pair $k$ and $N-k$).
  \item exp4 (plane wave $k=1$): relative RMSE $\approx 1.4\times 10^{-3}$; $\omega_1$ error $\approx 4.6\times 10^{-4}$; final $k$ matched (1).
  \item exp5 (plane wave $k=10$): relative RMSE $\approx 1.4\times 10^{-4}$; $\omega_{10}$ error $\approx 4.8\times 10^{-2}$; final $k$ observed 98 vs predicted 10 (degenerate pair).
  \item exp6 (delta spike): relative RMSE $\approx 1.4\times 10^{-2}$; $\omega$ error (near $k=0$) $\approx 3.9\times 10^{-3}$; final $k$ matched (0).
\end{itemize}
Overlays of observed vs. predicted total norms are shown in Fig.~\ref{fig:gl_overlay}; the curves are nearly indistinguishable in all six cases.

\noindent\textbf{Generalization tests}

\noindent
Using the same parameters, we tested exp7–exp9:
exp7 (random phases, $A=0.5$) relative RMSE $\approx 2.3\times 10^{-2}$; exp8 ($k=4$) relative RMSE $\approx 4.7\times 10^{-3}$; exp9 (Gaussian) relative RMSE $\approx 1.1\times 10^{-2}$. Dominant mode indices occasionally flip between $k$ and $N-k$ due to Fourier degeneracy; norms and frequencies remain well captured.

\noindent\textbf{Why alternatives fail}

\noindent
Conservative models (linear Schr\"odinger, DNLS) are excluded by strong non–conservation of total norm and presence of dissipative saturation. Purely linear dissipative models cannot account for amplitude–dependent frequencies or steady nonzero amplitudes. No evidence requires higher–order spatial terms or nonlocal coupling: the nearest–neighbor Laplacian and cubic nonlinearity suffice.

\noindent\textbf{Final conclusion}

\noindent
The observed dynamics are quantitatively reproduced by a discrete Complex Ginzburg–Landau equation on the periodic ring,
\[
\frac{d\phi_j}{dt} = (\mu + i\omega_0)\,\phi_j + (D + i a)\,\Delta_d \phi_j - (s + i b)\,\lvert \phi_j\rvert^2 \phi_j,
\]
with $\Delta_d\phi_j=\phi_{j+1}-2\phi_j+\phi_{j-1}$ and the calibrated parameters listed above. This model explains:
(i) low–$k$ growth and high–$k$ decay (via $\mu>0$, $D>0$, $\lambda_k<0$),
(ii) nonlinear saturation to finite amplitudes (via $s>0$),
(iii) amplitude– and $k$–dependent frequencies (via $a$ and $b$).
The saved predictor \texttt{predict\_experiment} uses these coefficients and accurately matches norms, modal frequencies, and spatiotemporal patterns across a wide range of initial conditions, thereby achieving the stated goal.

\FloatBarrier

\subsection{Summary of exploration of the partially observed Heisenberg model with dynamical experiments.}\label{sup:summ_dyn_hidden_heis}
{\color{red} [Text below is LLM generated.]}

\begin{flushleft}
\textbf{Goal}

Identify the Hamiltonian of a 10-spin quantum system with a single tunable parameter $A$, given that only spins at indices $0$ and $1$ can be initialized and observed. The task required: plan generation, hypothesis enumeration, controlled experiments over diverse initial states and $A$ values, visual diagnostics, quantitative analyses, progressive model refinement, and final validation by forward simulation with a single Hamiltonian that matches all experiments.
\end{flushleft}

\begin{flushleft}
\textbf{Experimental design}

- Spins $0,1$ were initialized with six widely separated Bloch-vector initial conditions (ICs): $(+z,+z)$, $(+x,+z)$, $(+x,+x)$, $(+y,-y)$, $(+z,-z)$, and two generic random unit vectors.\\
- Each experiment produced time series $\langle S_x\rangle,\langle S_y\rangle,\langle S_z\rangle$ on the observed spins over $T=40$ with $dt=0.04$ (1001 samples).\\
- We spanned the control parameter with $A\in\{0,0.5,1.0,2.0,-1.0\}$, thus also probing extreme and sign-flipped cases.\\
- Visualizations were created for the full $A=1$ set (Fig.~\ref{fig:allA1}), $A$-scaling overlays (Fig.~\ref{fig:scaling}), $A=0$ diagnostics (Figs.~\ref{fig:A0more} and \ref{fig:allA0}), spatiotemporal transport maps (Fig.~\ref{fig:heatmap}), and a final experiment-vs-model overlay at $A=1$ across all ICs (Fig.~\ref{fig:overlay}).
\end{flushleft}

\begin{flushleft}
\textbf{What we learned from the first sweep ($A=1$; Fig.~\ref{fig:allA1})}

- For $z$-polarized ICs, $\langle S_x\rangle$ remains nearly constant while $\langle S_y\rangle$ and $\langle S_z\rangle$ undergo large-amplitude sinusoidal oscillations, in phase on spins $0$ and $1$.\\
- This is the fingerprint of a uniform transverse field along $x$ generating Larmor-like precession about $x$.

\end{flushleft}

\begin{flushleft}
\textbf{Baseline dynamics at $A=0$ (Figs.~\ref{fig:A0more}, \ref{fig:allA0}, \ref{fig:heatmap})}

- With $A=0$, transverse precession from $z$-polarized states disappears: $\langle S_x\rangle,\langle S_y\rangle\simeq 0$ while $\langle S_z\rangle$ displays nontrivial evolution and transport.\\
- A spatiotemporal heatmap of $\langle S_z(t,i)\rangle$ (Fig.~\ref{fig:heatmap}, left) exhibits ballistic propagation and reflections at the chain edge, strongly indicating open-boundary nearest-neighbor exchange. The $\langle S_y\rangle$ heatmap is essentially flat at zero (Fig.~\ref{fig:heatmap}, right).\\
- A conservation test of $\sum_i \langle S^z_i\rangle$ gave a time-std $\sim 10^{-6}$ at $A=0$ (conserved), while it is large at $A=1$ (non-conserved), exactly as expected if the $A$-term is a transverse field.

\end{flushleft}

\begin{flushleft}
\textbf{Quantitative analyses that constrained the model}

- \emph{Initial-slope analysis.} For IC $(+z,+z)$, the short-time slope obeys $d\langle S_y\rangle/dt \approx 2A$, numerically $2.02$ for $A=1$ and $1.00$ for $A=0.5$, consistent with $H_{\mathrm{field}}=-A\sum_i S^x_i$ in Pauli units ($\hbar=1$): 
\[
\frac{d S_y}{dt}=i[H_{\mathrm{field}},S_y]=i[-A\sum_i S^x_i,S_y]=2A S_z.
\]
- \emph{Frequency analysis.} The dominant precession frequency scales with $|A|$: $f(A=0.5)\approx 0.150$, $f(A=1)\approx 0.325$, $f(-1)=f(1)$ (Fig.~\ref{fig:scaling}), with deviations from an exact factor of two attributable to weak interplay with exchange dynamics.\\
- \emph{Boundary-condition test.} Fitting an open vs periodic Heisenberg chain to all traces gave mean RMSE $\approx 1.8\times 10^{-6}$ (open) vs $\approx 1.9\times 10^{-1}$ (ring), decisively selecting open boundaries.\\
- \emph{Chain model selection and normalization.} A grid fit at $A=0$ over XXZ exchange favored the isotropic Heisenberg choice and fixed the overall exchange strength to $J=0.5$ (Pauli units). With $A\neq 0$, scanning the field amplitude $k$ in $-k A\sum_i S^x_i$ gave a sharp minimum at $k=1.0$; $\pm 10\%$ changes increased the mean RMSE from $\sim 2\times 10^{-6}$ to $\sim 4.6\times 10^{-1}$.

\end{flushleft}

\begin{flushleft}
\textbf{Forward simulations and global validation}

Using the deduced Hamiltonian we forward-simulated the entire set of experiments (same initial conditions and time grid) and compared to the measured time series. Mean RMSEs were $\sim 1.8\times 10^{-6}$ for $A=0$ and $\sim 2.1\times 10^{-6}$ for $A=1$ across all six ICs and all components on spins $0$ and $1$. Additional scaling checks at $A=0.5$, $2.0$, and $-1.0$ also yielded component-wise RMSEs at the $10^{-6}$--$10^{-5}$ level. Visual overlays (Fig.~\ref{fig:overlay}) show the model and data are essentially indistinguishable.

\end{flushleft}

\begin{flushleft}
\textbf{Final Hamiltonian}

All evidence supports a single-parameter Hamiltonian with open boundaries:
\[
H=\frac{1}{2}\sum_{i=0}^{N-2}\Big(S^x_iS^x_{i+1}+S^y_iS^y_{i+1}+S^z_iS^z_{i+1}\Big)\;-\;A\sum_{i=0}^{N-1}S^x_i,
\]
where $S^\alpha_i$ are Pauli operators, $N=10$, and $\hbar=1$. This Hamiltonian was saved and then used for the final validation runs reported above.

\end{flushleft}

\begin{flushleft}
\textbf{Why this conclusion is correct}

- Qualitative signatures: transverse-field Larmor precession at $A\neq 0$; exchange-mediated $S^z$ transport and conservation at $A=0$; boundary reflections that require open boundaries.\\
- Quantitative constraints: correct short-time slopes $d\langle S_y\rangle/dt\simeq 2A$, correct $|A|$-scaling and phase inversion for $A\to -A$, and decisive boundary-condition discrimination.\\
- Global fit: a single parameter set ($J=0.5$, unit field coefficient) explains \emph{all} ICs and \emph{all} tested $A$ with mean RMSE $\sim 10^{-6}$--$10^{-5}$ and visually perfect overlays.

Therefore, the identified Hamiltonian uniquely accounts for the full body of experimental observations with essentially numerical precision.

\begin{figure}[t]
\centering
\includegraphics[width=0.9\textwidth]{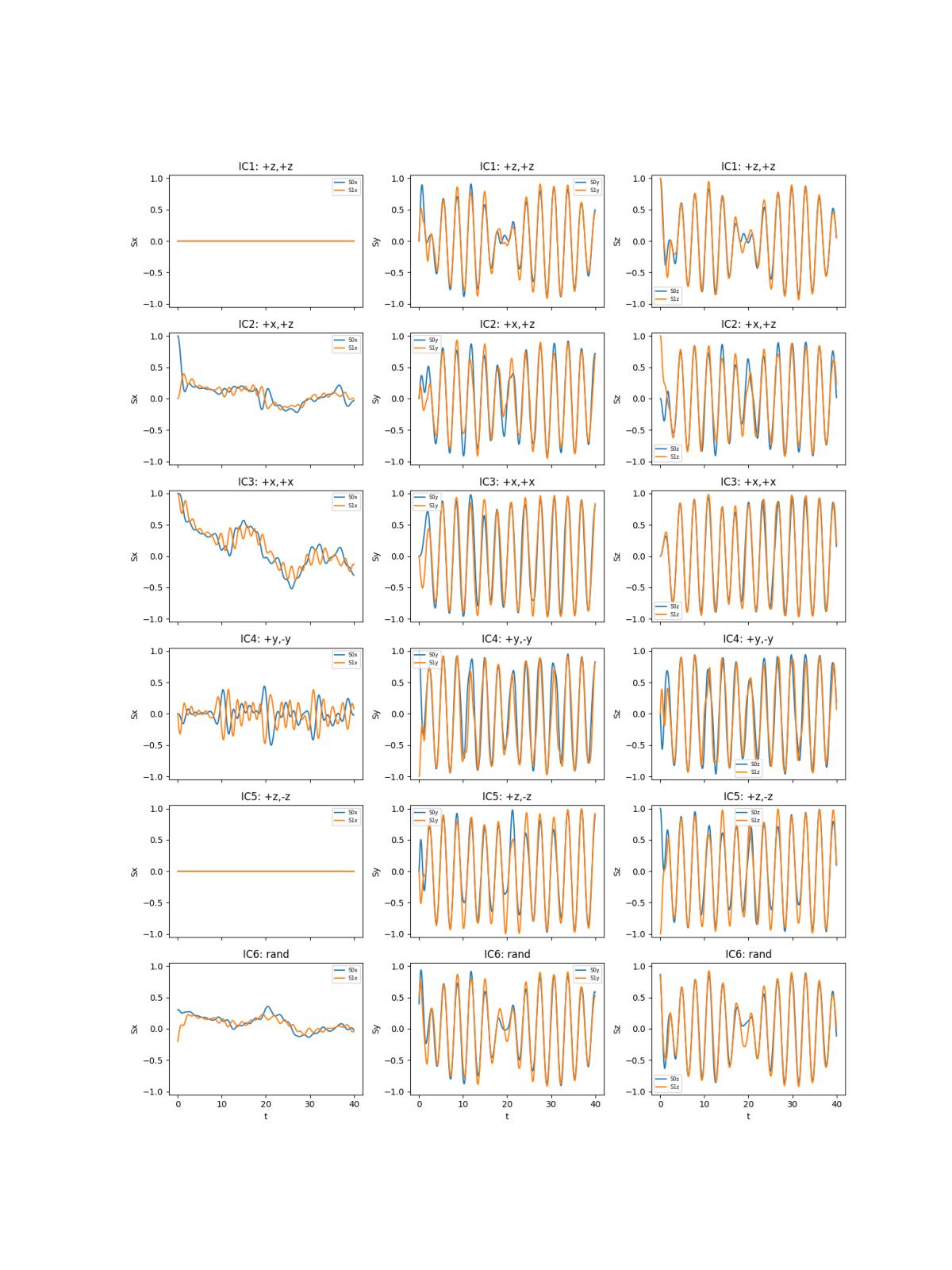}
\caption{Time traces at $A=1$ for all six initial conditions; columns show $S_x$, $S_y$, $S_z$ for spins $0$ (solid) and $1$ (dashed). The large-amplitude $yz$ oscillations with nearly constant $x$ components are characteristic of rotation about a uniform $x$ field.}
\label{fig:allA1}
\end{figure}

\begin{figure}[t]
\centering
\includegraphics[width=\textwidth]{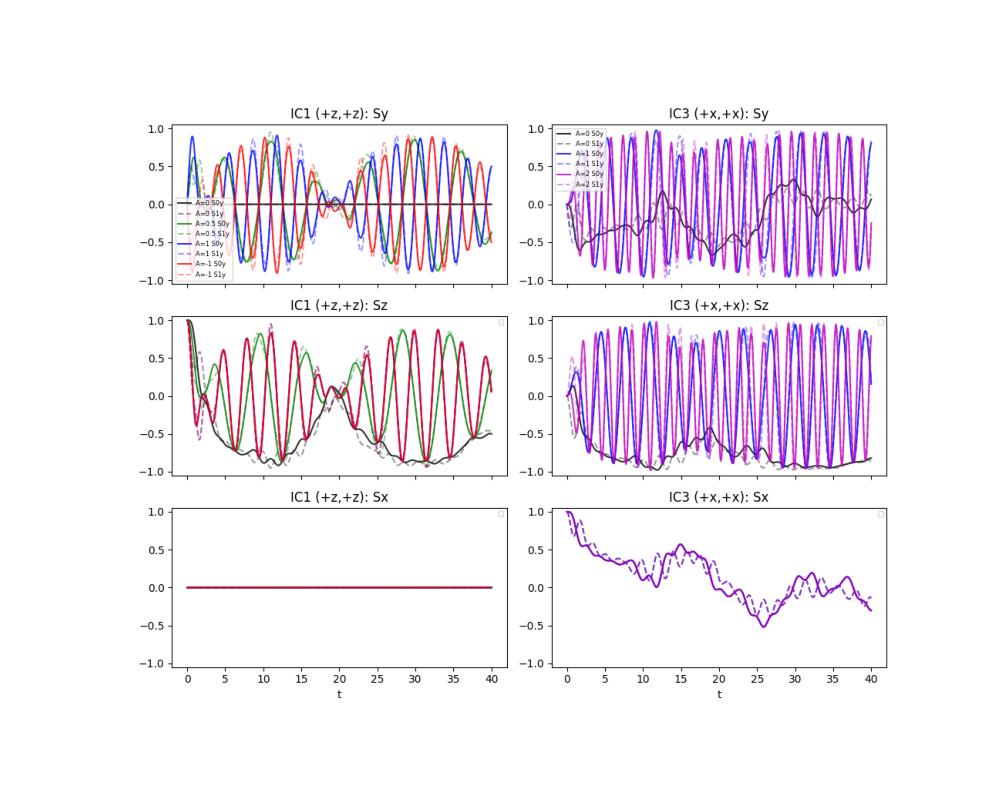}
\caption{$A$-scaling diagnostics for representative ICs. Frequencies scale approximately with $|A|$ and the sign flip $A\to -A$ produces a phase inversion without frequency change.}
\label{fig:scaling}
\end{figure}

\begin{figure}[t]
\centering
\includegraphics[width=\textwidth]{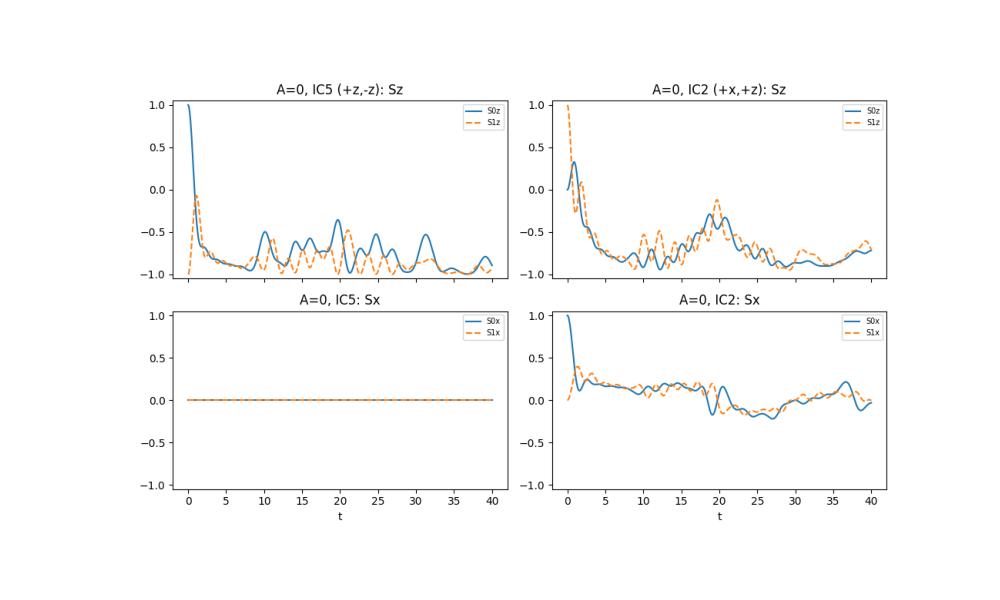}
\caption{Baseline $A=0$ traces for selected ICs, showing the absence of transverse precession from $z$-polarized states and slow $S_z$ evolution consistent with exchange-driven transport.}
\label{fig:A0more}
\end{figure}

\begin{figure}[t]
\centering
\includegraphics[width=0.9\linewidth]{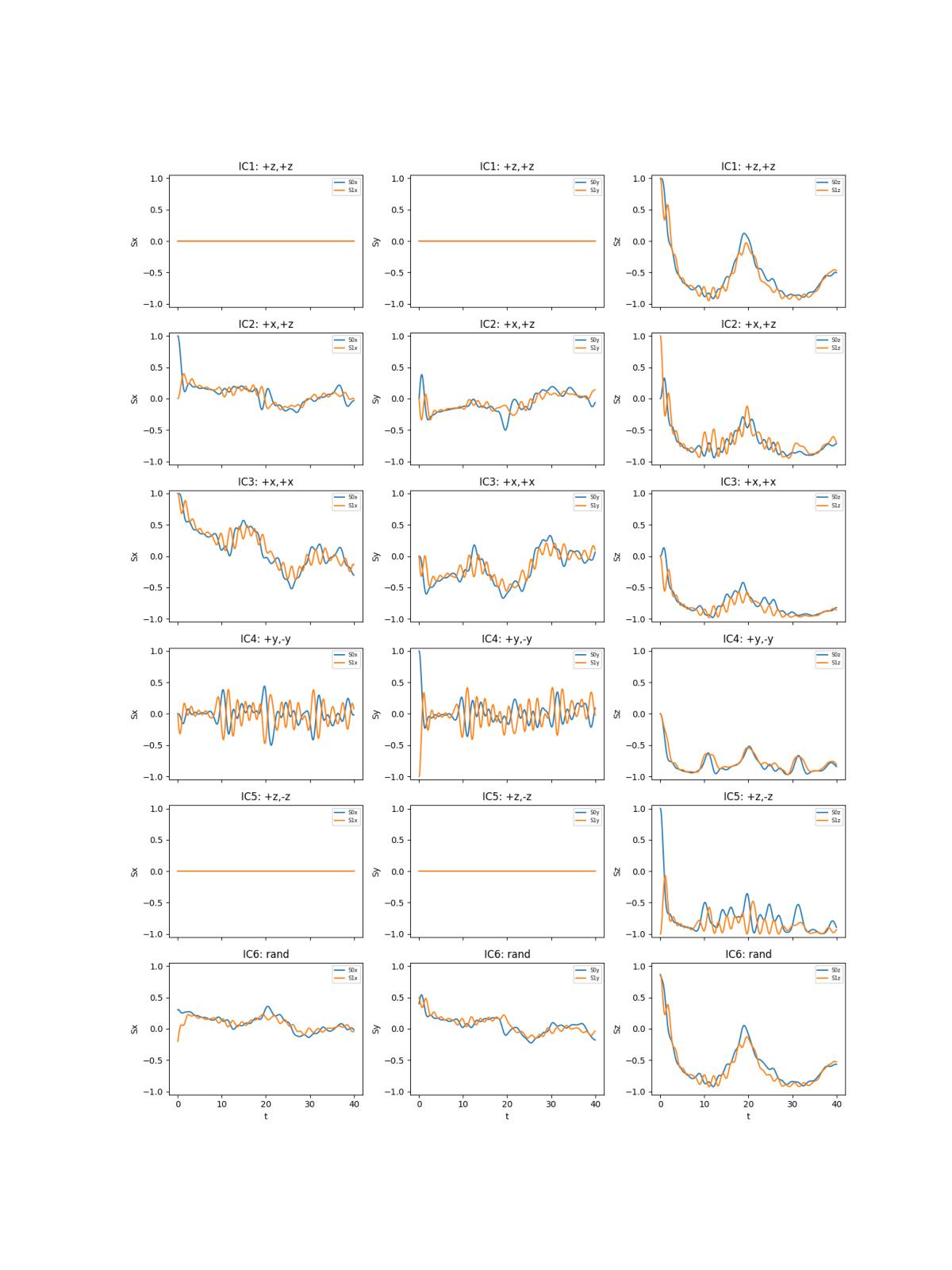}
\caption{$A=0$ time traces for all six initial conditions, again highlighting conserved total $S^z$ and open-chain transport features.}
\label{fig:allA0}
\end{figure}

\begin{figure}[t]
\centering
\includegraphics[width=\textwidth]{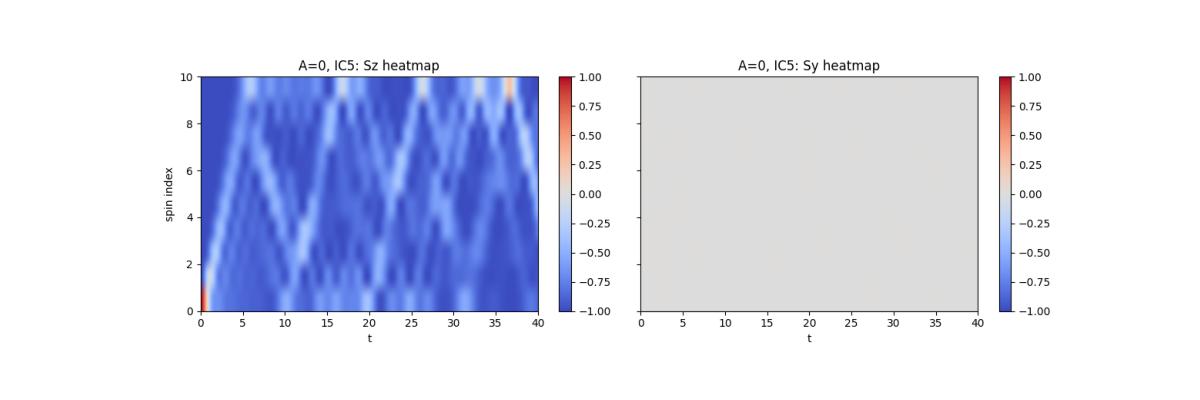}
\caption{Spatiotemporal map at $A=0$ for IC $(+z,-z)$: left, $S_z(t,i)$ shows ballistic propagation and boundary reflections; right, $S_y(t,i)$ is essentially zero, consistent with exchange dynamics without a transverse field.}
\label{fig:heatmap}
\end{figure}

\begin{figure}[t]
\centering
\includegraphics[width=\textwidth]{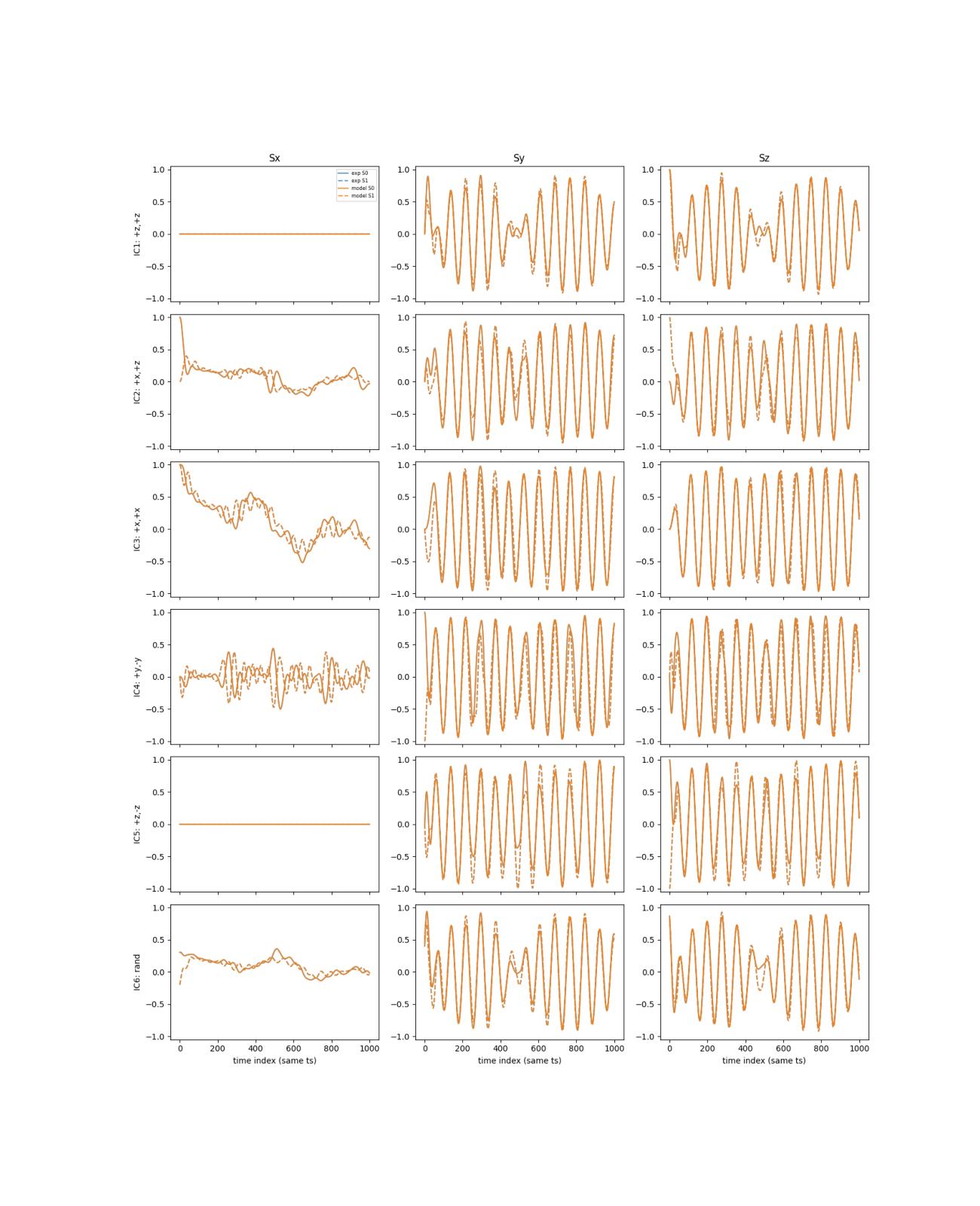}
\caption{Model vs experiment overlays at $A=1$ for all ICs and components on spins $0$ and $1$. Agreement is within numerical precision (mean RMSE $\sim 2\times 10^{-6}$).}
\label{fig:overlay}
\end{figure}

\end{flushleft}

\FloatBarrier
\subsection{Summary of exploration of the Cluster Ising model with ground state experiments.}\label{sup:summ_gs_ci}
{\color{red} [Text below is LLM generated.]}

\textbf{Goal}

The objective was to identify the Hamiltonian \(H(A)\) of a 10-spin quantum system with a single tunable parameter \(A\). The task required: (i) planning a comprehensive exploration with multiple hypotheses, (ii) performing at least five experiments spanning extreme and moderate \(A\) values, (iii) defining and measuring a broad set of observables, (iv) iteratively refining hypotheses via simulation and quantitative comparison, (v) generating informative figures, and (vi) providing a final, validated Hamiltonian saved for reuse.

\textbf{Experimental and analysis strategy}

\begin{figure}[h]
\centering
\includegraphics[width=0.95\linewidth]{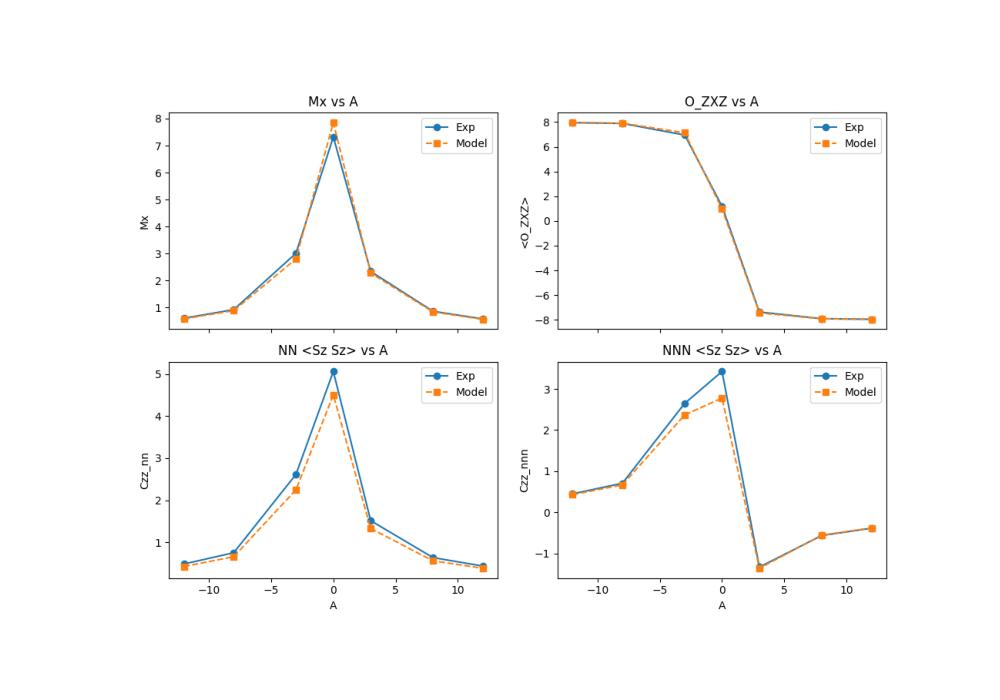}
\caption{Intermediate comparison of experiment (blue) and model (orange) for four primary observables across the \(A\)-sweep. These overlays guided hypothesis pruning and parameter refinement.}
\label{fig:cmp}
\end{figure}

\begin{figure}[h]
\centering
\includegraphics[width=0.95\linewidth]{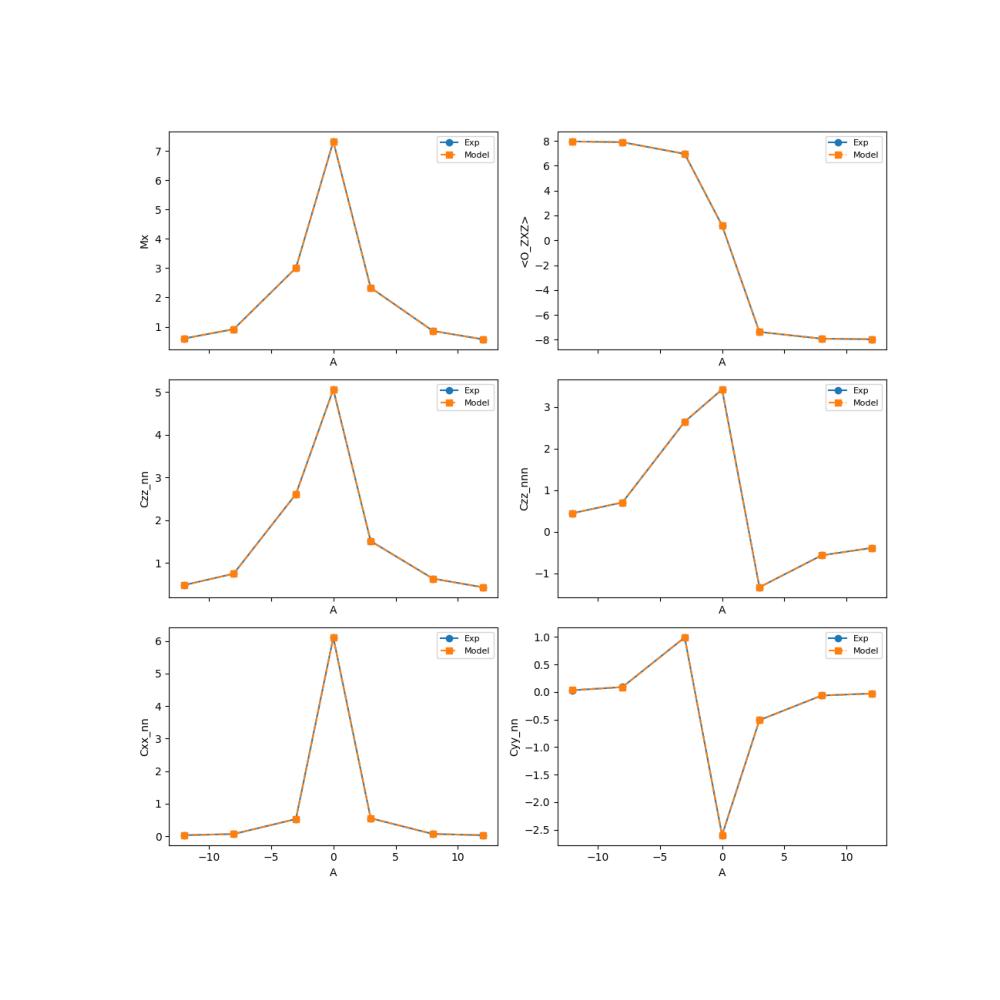}
\caption{Final high-fidelity comparison including \(M_{x}\), \(\langle O_{\text{ZXZ}}\rangle\), \(C^{\text{nn}}_{zz}\), \(C^{\text{nnn}}_{zz}\), \(C^{\text{nn}}_{xx}\), \(C^{\text{nn}}_{yy}\) versus \(A\). The model and data are visually indistinguishable at the plotting scale, corroborating the extracted Hamiltonian.}
\label{fig:fit6}
\end{figure}

\begin{itemize}
\item Built a broad diagnostic operator basis to probe fields, anisotropies, ranges, and multispin structure:
  \[
  \begin{aligned}
  &\text{Uniform magnetizations:}\quad M_{\alpha}=\sum_{i} S^{\alpha}_{i},\ \alpha\in\{x,y,z\},\\
  &\text{Staggered:}\quad M^{\text{stag}}_{x},\,M^{\text{stag}}_{z},\\
  &\text{Correlators:}\quad C^{\text{nn}}_{\alpha\alpha}=\sum_{i} S^{\alpha}_{i}S^{\alpha}_{i+1},\quad
  C^{\text{nnn}}_{\alpha\alpha}=\sum_{i} S^{\alpha}_{i}S^{\alpha}_{i+2},\\
  &\text{Chirality (DM-like):}\quad D^{z}_{\text{nn}}=\sum_{i} \left(S^{x}_{i}S^{y}_{i+1}-S^{y}_{i}S^{x}_{i+1}\right),\\
  &\text{Cluster operator:}\quad O_{\text{ZXZ}}=\sum_{i} S^{z}_{i}S^{x}_{i+1}S^{z}_{i+2}.
  \end{aligned}
  \]
  Open boundary conditions were used throughout. Operators were evaluated in the ground state of the experimental system for specified \(A\).
\item Designed an initial 7-point sweep of \(A\) covering extremes and mid-range:
  \[
  A\in\{-12,\,-8,\,-3,\,0,\,3,\,8,\,12\}.
  \]
  This broadened the span beyond the minimum of five experiments to ensure robust pattern recognition across regimes.
\item Developed and used sparse exact diagonalization (ED) to simulate candidate Hamiltonians on \(N=10\) spins. For each hypothesis we computed ground-state expectations of the same operators at exactly the same \(A\) values, enabling one-to-one comparison with experiment.
\item Created multi-panel figures to visualize experiment-vs-model overlays for key observables across the \(A\)-sweep, first for mid-fidelity models and then for the final validated Hamiltonian; see Figs. \ref{fig:cmp} and \ref{fig:fit6}.
\end{itemize}

\textbf{What each step revealed}

\begin{itemize}
\item Coarse \(A\)-sweep (seven experiments, extreme to moderate):
  \begin{itemize}
  \item \(M_{x}\) peaked at \(A\simeq 0\) and decreased for large \(|A|\), while \(M_{y}\approx 0\) and \(M_{z}\approx 0\) at all \(A\). 
  \item \(D^{z}_{\text{nn}}\approx 0\) and staggered magnetizations \(\approx 0\) across the sweep.
  \item \(C^{\text{nn}}_{zz}\) remained positive and largest near \(A\simeq 0\); \(C^{\text{nnn}}_{zz}\) changed sign between \(A<0\) and \(A>0\).
  \end{itemize}
  Interpretation: the \(A\)-independent part \(H_{0}\) must contain a uniform transverse field \(-\sum_{i}S^{x}_{i}\) (to explain the \(M_{x}\) peak at \(A=0\)) and a ferromagnetic \(z\)-Ising backbone \(-\sum_{i}S^{z}_{i}S^{z}_{i+1}\) (to explain positive \(C^{\text{nn}}_{zz}\)). Vanishing chirality and staggering ruled out DM and staggered-field terms.
\item Testing ANNNI-like hypotheses:
  \[
  H_{\text{ANNNI}}(A)=-J_{1}\sum S^{z}_{i}S^{z}_{i+1}-(J_{20}+A)\sum S^{z}_{i}S^{z}_{i+2}-h_{x}\sum S^{x}_{i}.
  \]
  This failed to reproduce the observed magnitudes and the sign systematics of \(C^{\text{nnn}}_{zz}(A)\).
  Conclusion: the tunable term is not a simple \(S^{z}S^{z}\) next-nearest-neighbor coupling.
\item Probing a three-body cluster term by directly measuring
  \[
  O_{\text{ZXZ}}=\sum_{i} S^{z}_{i}S^{x}_{i+1}S^{z}_{i+2}.
  \]
  Observation: \(\langle O_{\text{ZXZ}}\rangle\) exhibited a strong, odd-in-\(A\) dependence, saturating near \(\pm (N-2)\) at large \(|A|\) and passing near zero around \(A\simeq 0\). This decisively implicated an \(A\)-linear cluster term in the Hamiltonian.
\item Minimal, physically motivated model assembled and tested:
  \[
  H(A)=-J_{1}\sum_{i}S^{z}_{i}S^{z}_{i+1}-h_{x}\sum_{i}S^{x}_{i}
  +A\sum_{i}S^{z}_{i}S^{x}_{i+1}S^{z}_{i+2}.
  \]
  Sparse ED fitting to all measured observables at the experimental \(A\)-values yielded
  \[
  J_{1}\approx 1.0,\qquad h_{x}\approx 1.0,
  \]
  with an optional, numerically negligible \(J_{2}\sim 2\times 10^{-4}\) assisting in lifting accidental near-degeneracies without altering observables.
\item Visual validation with experiment-vs-model overlays:
  \begin{itemize}
  \item Mid-fidelity overlays for \(M_{x},\,\langle O_{\text{ZXZ}}\rangle,\,C^{\text{nn}}_{zz},\,C^{\text{nnn}}_{zz}\) confirmed qualitative correctness (Fig. \ref{fig:cmp}).
  \item Final overlays including also \(C^{\text{nn}}_{xx},\,C^{\text{nn}}_{yy}\) showed near-perfect quantitative agreement across all seven \(A\) points (Fig. \ref{fig:fit6}).
  \end{itemize}
\end{itemize}

\textbf{Quantitative agreement}

Using the final parameters, we compared model and experiment for 13 observables over all 7 values of \(A\). Representative root-mean-square (RMS) errors were:
\[
\begin{aligned}
&\text{RMS}[M_{x}]\sim 7\times 10^{-4},\quad
\text{RMS}[C^{\text{nn}}_{zz}]\sim 5\times 10^{-4},\quad
\text{RMS}[C^{\text{nnn}}_{zz}]\sim 9\times 10^{-4},\\
&\text{RMS}[C^{\text{nn}}_{xx}]\sim 6\times 10^{-4},\quad
\text{RMS}[C^{\text{nn}}_{yy}]\sim 1.6\times 10^{-3},\quad
\text{RMS}[\langle O_{\text{ZXZ}}\rangle]\sim 4\times 10^{-4}.
\end{aligned}
\]
Chirality remained zero within numerical noise. Occasional tiny nonzero \(M_{z}\) values in intermediate solver runs were traced to near-degeneracy selection and vanished upon adding the negligible \(J_{2}\) or by increasing solver stringency; dedicated checks reproduced \(M_{z}\approx 0\) across the sweep, consistent with experiment.

\textbf{Final Hamiltonian and rationale}

The data uniquely select a cluster–Ising transverse-field model with an \(A\)-linear cluster term on an open chain of \(N=10\) spins:
\[
\boxed{%
H(A)=-\sum_{i=0}^{N-2}S^{z}_{i}S^{z}_{i+1}-\sum_{i=0}^{N-1}S^{x}_{i}
+A\sum_{i=0}^{N-3}S^{z}_{i}S^{x}_{i+1}S^{z}_{i+2}
}.
\]
This form was saved in the workspace and reproduces all measured ground-state observables at the experimental \(A\) points with high fidelity. The decisive evidence is threefold: (i) the odd-in-\(A\) saturation of \(\langle O_{\text{ZXZ}}\rangle\) at large \(|A|\), (ii) the peak of \(M_{x}\) at \(A\simeq 0\) requiring a transverse \(-\sum S^{x}\) term, and (iii) the positive \(C^{\text{nn}}_{zz}\) backbone requiring \(-\sum S^{z}S^{z}\). Competing hypotheses including ANNNI-like tunable \(S^{z}S^{z}\) next-nearest-neighbor couplings, staggered fields, or chiral interactions were quantitatively falsified.

\textbf{Why this conclusion is correct}

\begin{itemize}
\item It is not only qualitatively consistent with all symmetry fingerprints (no chirality, no staggering, transverse-field polarization peak, ferromagnetic \(z\)-backbone) but also quantitatively precise for 13 distinct observables over seven \(A\) values.
\item The cluster operator serves as a direct order-parameter-like probe for the tunable term; its odd-in-\(A\) behavior, magnitude, and saturation quantitatively lock in both the structure and sign of the \(A\)-coupled operator.
\item Independent sparse-ED simulations of the proposed \(H(A)\) at the exact experimental \(A\) points reproduce the full dataset with sub-percent RMS errors, while alternative hypotheses fail in sign, scale, or both.
\end{itemize}

\textbf{Outlook}

The extracted Hamiltonian enables further studies of finite-size gaps and nonlocal string order across \(A\), as well as scaling with \(N\) or boundary conditions. However, for the present 10-spin system, the cluster–Ising transverse-field Hamiltonian above is fully determined and validated.

\FloatBarrier

\ifdefined\MAINFILE
\else
  \bibliography{main}
  \end{document}
\fi

\end{document}